\pdfoutput=1

\documentclass{article}
\usepackage{ijcai26}

\usepackage{times}
\usepackage{soul}
\usepackage{url}
\usepackage[hidelinks]{hyperref}
\usepackage[utf8]{inputenc}
\usepackage[small]{caption}
\usepackage{graphicx}
\usepackage{amsmath}
\usepackage{amsthm}
\usepackage{booktabs}

\usepackage{amssymb}
\usepackage{mathtools}
\usepackage{dsfont}
\usepackage{textcomp} %
\usepackage{IEEEtrantools}
\usepackage{tabularx}
\usepackage{array}
\usepackage{threeparttable}
\usepackage[table]{xcolor}
\usepackage{adjustbox}
\usepackage{placeins}
\usepackage{subfig}
\usepackage[ruled,vlined]{algorithm2e}
\usepackage{tikz}
\usetikzlibrary{arrows.meta,positioning,shapes.geometric}
\usepackage{pifont}
\usepackage[most]{tcolorbox}
\usepackage{enumitem}
\usepackage{microtype}
\graphicspath{{Figure/}}

\newcommand*\emptycirc[1][1ex]{\tikz\draw[thick] (0,0) circle (3pt);}
\newcommand*\halfcircc[1][1ex]{\tikz{
    \draw[thick] (0,0) circle (4pt);
    \fill[black] (0,0) -- ++(0, 4pt) arc[start angle=90, end angle=-90, radius=4pt] -- cycle;
}}
\newcommand*\halfcirc[1][1ex]{\tikz{
    \draw[thick] (0,0) circle (5pt);
    \fill[black] (0,0) -- ++(0, 5pt) arc[start angle=90, end angle=-90, radius=5pt] -- cycle;
}}
\newcommand{\sysname}{\textsf{ReShap}}

\newtcolorbox{promptbox}[1]{
    colback=gray!5,
    colframe=black!75,
    fonttitle=\bfseries,
    title=#1,
    arc=2mm,
    enhanced,
    breakable,
    boxsep=1mm, left=2mm, right=2mm, top=1mm, bottom=1mm,
    attach boxed title to top left={xshift=3mm, yshift=-2mm},
    boxed title style={colback=black!75}
}

\title{ReShap: Interpretable Reinforcement Learning for High-Dimensional Multi-Action Settings}

\author{
Yiyu Qian$^1$\And
Pengfei Guo$^2$\And
Liyuan Zhao$^3$\\
\affiliations
$^1$RMIT University\\
$^2$Damiao Tech\\
$^3$University of California, Irvine\\
\emails
s4218854@student.rmit.edu.au,
pengfei.guo2001@gmail.com,
liyuanz6@uci.edu
}

\begin{document}
\maketitle

\begin{abstract}
Deep Reinforcement Learning (RL) plays a crucial role in tackling complex tasks, yet its reliance on deep neural networks makes the learned policies opaque. To enhance the trustworthiness of RL policies, existing work (e.g., the {SILVER} framework~\cite{li2025explainability}) has focused on Shapley-based interpretability, but is typically restricted to low-dimensional state spaces, binary-action settings, and a single class of linear surrogate. We propose $\sysname$, a policy-distillation framework that extends Shapley-based interpretability to high-dimensional, multi-action environments. $\sysname$ extracts compact feature representations from raw image observations, performs SHAP-based feature attribution to characterize how each feature contributes to the policy's decisions, and distills the deep RL policy into an interpretable surrogate trained on the full set of (state, policy-action) pairs along the policy's visited distribution. We instantiate the surrogate with three interpretable model families (decision trees, linear regression, and logistic regression) and compare their fidelity and task return. We evaluate $\sysname$ on Atari and Box2D environments using three deep RL algorithms and conduct a human-subject study to assess the clarity and trustworthiness of the resulting policies. Results show that $\sysname$ preserves competitive task performance, that decision trees generally outperform linear and logistic regression, and that linear regression is weakest in most settings.
\end{abstract}

\section{Introduction}
Deep Reinforcement Learning (RL) has emerged as a powerful paradigm for solving diverse and challenging tasks, ranging from strategic gameplay, such as AlphaGo Zero~\cite{shaheen2025reinforcement}, and Atari games~\cite{jang2024analyzing}, to robotic control~\cite{tang2025deep} and fine-tuning of Large Language Models (LLMs)~\cite{zhai2024fine}. These advances are largely attributable to the representational power of Deep Neural Networks (DNNs), which excel at feature extraction and processing high-dimensional data. However, the opaque nature of DNNs poses significant challenges for understanding and interpreting the decision-making processes of RL agents~\cite{murad2024unraveling,fan2023rethinking}. This lack of transparency is particularly concerning in high-stakes applications, such as autonomous driving or healthcare~\cite{wang2024trustmis}, where opaque decisions can directly lead to accidents or harm. %

To address this issue, Explainable RL (XRL) has emerged to help humans understand RL agents’ decisions, improving transparency and trust in high-stakes applications~\cite{shu2026trustworthy}. Within XRL, two related but distinct concepts are often distinguished: interpretability and explainability. Interpretability refers to the intrinsic transparency of a model's structure and operation~\cite{li2025explainability}, often achieved through simpler models such as decision trees~\cite{costa2024evolving} or logistic regression~\cite{tahirovic2023interpretability}, which render the policy self-explanatory. Approaches based on this concept are typically categorized as model-level explanation methods~\cite{cheng2025survey}. Explainability, by contrast, involves post-hoc techniques that provide insights into the behavior of trained models, aiming to clarify or justify their decisions~\cite{li2025explainability}. These approaches include dataset-level explanation methods, which identify influential data in RL~\cite{cheng2025survey}. 
A representative approach is Shapley Additive Explanations (SHAP)~\cite{lundberg2017unified}, which assigns feature importance values based on Shapley theory to provide local explanations~\cite{song2024comprehensive}. Recent XRL studies~\cite{beechey2023explaining,beechey2025theoretical} have applied SHAP to quantify the contribution of individual state features to RL value function outputs, demonstrating its effectiveness in providing meaningful explanations. However, these studies focus on a single aspect, either interpretability or explainability.

To combine model-level transparency with post-hoc analysis for a more comprehensive understanding of deep RL policies, \citeauthor{li2025explainability}~\shortcite{li2025explainability} propose a Shapley value-based Interpretable poLicy Via Explanation Regression ({SILVER}) framework.
Although {SILVER} performs well in relatively simple environments such as CartPole and MountainCar, it relies on two key assumptions: first, that the observation space consists of discrete, well-defined features, which simplifies the computation of Shapley values; and second, that the policy operates in a binary action space, where action labeling at boundary points relies solely on geometric proximity. In complex environments with high-dimensional state spaces and large action spaces, both assumptions no longer hold. The lack of explicitly defined features makes direct SHAP analysis difficult, while multi-action scenarios render action modeling using geometric proximity infeasible. Other XRL studies have attempted to address these challenges in high-dimensional settings through decision-relevance visualization techniques (e.g., saliency maps)~\cite{greydanus2018visualizing}, causal modeling~\cite{madumal2020explainable}, or by employing modified SHAP methods adapted for continuous environments~\cite{beechey2023explaining} or specific applications such as counter-UAV systems~\cite{ccetin2024analyzing}. However, these approaches either fail to produce globally interpretable policies or lack systematic evaluation of the quality of the explanations. Therefore, there remains a pressing need for an XRL method capable of providing both interpretability and explainability in high-dimensional, multi-action environments.

To address these limitations, we propose \textsf{ReShap}, a framework that extends Shapley-based interpretability to multi-action and high-dimensional environments. Rather than fitting a surrogate on a small set of geometrically selected points, \textsf{ReShap} distills the RL policy directly from the full set of state-action pairs the policy produces along its visited distribution, so that the interpretable surrogate reflects the agent's actual behavior. This design yields interpretable policies that remain consistent with the RL policy, even in visually complex domains.

This paper makes three primary contributions:

\begin{itemize}[leftmargin=*,topsep=0pt]
\item \textbf{Scalable Interpretability in High-Dimensional Environments}: We extend {SILVER} to operate on visually rich Atari domains, where both the state and action spaces are significantly more complex than in low-dimensional control tasks. This extension enables SHAP analysis on compact feature representations learned from raw RGB observations, enhancing the framework’s scalability and robustness.

\item \textbf{Interpretable Policy Distillation with Multiple Surrogate Families}: We distill the deep RL policy into interpretable surrogates trained on the full set of (state, policy-action) pairs drawn from the policy's visited distribution, and we systematically compare three surrogate families (decision trees, linear regression, and logistic regression) in terms of fidelity to the RL policy and task return.

\item \textbf{Human-Centered Evaluation of Comprehensibility and Trust}: We conduct a human-subject study to evaluate how the derived interpretable policy supports user understanding and trust. Through structured feedback, we assess the extent to which these models enhance the perceived transparency and reliability of RL agents’ decision-making processes.
\end{itemize}
Our approach follows the indirect paradigm, using policy distillation to transform deep RL policies into compact surrogate models for high-dimensional, multi-action environments, ensuring transparency without sacrificing performance.  Our code is available at \url{https://anonymous.4open.science/r/ReShap-E14E/}.

\section{Related Work}
This section reviews related work on XRL methods in interpretability and explainability.

\paragraph{Interpretability.}
Interpretability-oriented XRL improves transparency by expressing policies in human-understandable forms~\cite{glanois2024survey}. Direct methods design interpretable controllers such as decision trees~\cite{mahbooba2021explainable} or analytical formulas~\cite{hein2018interpretable}, with recent advances in differentiable trees via distillation~\cite{gokhale2024distill2explain}. Indirect methods first train RL policies and then convert them into interpretable surrogates, e.g., VIPER~\cite{bastani2018verifiable} and PIRL~\cite{verma2018programmatically}. Despite progress in interpretable policy representations, existing interpretability-oriented XRL approaches often struggle to scale to high-dimensional state spaces and multi-action settings without sacrificing performance.

\paragraph{Explainability.}
Explainability-focused XRL methods analyze how training data influence policy learning. The influence function~\cite{koh2017understanding} estimates the effect of reweighting a sample on model parameters and can be extended to RL for assessing replay transitions.
Shapley values~\cite{shapley1953value} provide a principled foundation for attribution.
Extensions such as Data Shapley~\cite{Ghorbani2019} and FreeShap~\cite{Wang2024} improve scalability and quantify each sample’s contribution to policy formation.
Recent studies move beyond static attribution toward behavioral and counterfactual reasoning: process-mining methods~\cite{qian2025exploring} expose sequential decision logic; SVERL~\cite{beechey2025theoretical} unifies behavioral and predictive perspectives; and counterfactual explainers~\cite{dong2025counterfactual} illustrate how alternative actions change outcomes. However, most explainability-oriented XRL methods focus on post-hoc attribution or behavioral analysis, with limited ability to translate explanations into compact and reusable policy representations.

\section{Preliminaries}
\subsection{Foundation of Reinforcement Learning}

Reinforcement learning (RL) focuses on learning sequential decision-making policies through interaction with an environment~\cite{sutton2018reinforcement}. The environment is commonly modeled as a Markov Decision Process (MDP), defined by the tuple
\begin{IEEEeqnarray}{rCL} 
\mathcal{M} = (S, A, P, R, \gamma), \nonumber
\end{IEEEeqnarray}
where \(S\) denotes the state space, \(A\) the action space, \(P(s' \mid s, a)\) the state transition probability, \(R(s, a)\) the reward function, and \(\gamma \in (0,1)\) the discount factor.

At each time step \(t\), the agent observes a state \(s_t \in S\) and selects an action \(a_t \in A\) according to a policy \(\pi\). The policy is defined as a mapping from states to actions or action distributions:
\begin{IEEEeqnarray}{rCL} 
\pi(a \mid s) \;\triangleq\; {\Pr}^{\pi}(a_t = a \mid s_t = s). \nonumber
\end{IEEEeqnarray}
A policy is deterministic if it outputs a single action for each state, and stochastic if it produces a probability distribution over actions.
The objective of reinforcement learning is to learn a policy \(\pi\) that maximizes the expected cumulative discounted return,
\begin{IEEEeqnarray}{rCL} 
J(\pi) = \mathbb{E}_{\pi}\!\left[ \sum_{t=0}^{\infty} \gamma^{t} R(s_t, a_t) \right], \nonumber
\end{IEEEeqnarray}
where the expectation is taken over trajectories induced by the policy \(\pi\) and the environment dynamics \(P\).

In practical implementations, a state \(s \in S\) is typically represented as a feature vector $s = \{s_1, s_2, \dots, s_n\}$. 
The policy output therefore depends on the joint contribution of multiple state features rather than any individual feature in isolation. This feature-based formulation naturally motivates the use of attribution methods to quantify the contribution of each state feature to the policy’s decision, which forms the foundation of the analysis presented in this work.

\subsection{Shapley Values in Reinforcement Learning}

Shapley values~\cite{das2022values} originate from cooperative game theory and provide a principled framework for attributing the output of a system to its individual components. Given a set of players $N = \{1, \dots, n\}$ and a characteristic function $v: 2^N \rightarrow \mathbb{R}$ that assigns a value to each coalition, the Shapley value of player $i \in N$ is defined as
\begin{equation}
\label{eq:shapley-value}
\phi_i(v) = \sum_{C \subseteq N \setminus \{i\}}
\frac{|C|!(n - |C| - 1)!}{n!}
\bigl[v(C \cup \{i\}) - v(C)\bigr],
\end{equation}
where $!$ denotes the factorial function used to account for the combinatorial permutations of players, and $\phi_i(v)$ represents the expected marginal contribution of player $i$ over all possible coalition orderings.

In RL, a state $s \in S$ is typically represented as a feature vector
$s = \{s_1, s_2, \dots, s_n\}$.
These state features can be naturally interpreted as players in a cooperative game, while the policy output reflects the collective value generated by their joint contribution. Consequently, attributing a policy’s decision to individual state features can be formulated as computing their Shapley values.
To apply Shapley values in this setting, it is necessary to define a characteristic function $v(C)$ that captures the policy output when only a subset of state features $s_C \subseteq s$ is observed. However, a trained policy $\pi(a \mid s)$ is generally undefined for partial inputs, and naive feature masking may result in out-of-distribution states and unreliable attributions.

To address this issue, we adapt an on-manifold characteristic function formulation~\cite{frye2020shapley,beechey2023explaining,beechey2025theoretical}, which preserves the data distribution by marginalizing over the unobserved features conditioned on the observed subset. Specifically, for a deterministic policy \( \pi: S \rightarrow A \), which outputs actions, the characteristic function is given by:

\begin{equation}
\label{eq:char-deterministic}
v^{\pi}(C) := \pi_C(s) = \sum_{s' \in S} {\Pr}^{\pi}(s'|s_C) \pi(s')
\end{equation}
where \( s' = s_C \cup s_C' \), and \( {\Pr}^{\pi}(s'|s_C) \) is the probability of being in state \( s' \) given the partial state features \( s_C \) observed under policy \( \pi \). For a stochastic policy \( \pi: S \times A \rightarrow [0, 1] \), which outputs action probabilities, the characteristic function is:
\begin{equation}
\label{eq:char-stochastic}
v^{\pi}(C)
:= \pi_C(a \mid s)
= \sum_{s' \in S} {\Pr}^{\pi}(s' \mid s_C)\,\pi(a \mid s').
\end{equation}

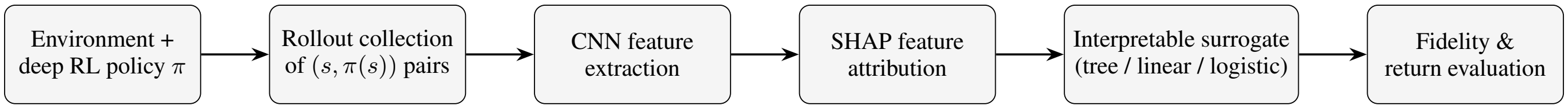
\begin{figure*}[t]
    \centering
    \resizebox{\textwidth}{!}{%
    \begin{tikzpicture}[
        box/.style={draw, rounded corners, align=center, minimum height=13mm,
                     minimum width=26mm, fill=gray!8, font=\small},
        node distance=6mm and 9mm
    ]
        \node[box] (env)  {Environment +\\ deep RL policy $\pi$};
        \node[box, right=of env]  (roll) {Rollout collection\\ of $(s,\pi(s))$ pairs};
        \node[box, right=of roll] (cnn)  {CNN feature\\ extraction};
        \node[box, right=of cnn]  (shap) {SHAP feature\\ attribution};
        \node[box, right=of shap] (surr) {Interpretable surrogate\\ (tree / linear / logistic)};
        \node[box, right=of surr] (eval) {Fidelity \&\\ return evaluation};
        \draw[-{Stealth}, thick] (env)  -- (roll);
        \draw[-{Stealth}, thick] (roll) -- (cnn);
        \draw[-{Stealth}, thick] (cnn)  -- (shap);
        \draw[-{Stealth}, thick] (shap) -- (surr);
        \draw[-{Stealth}, thick] (surr) -- (eval);
    \end{tikzpicture}%
    }
    \caption{\textsf{ReShap} overview: the deep RL policy $\pi$ is rolled out to collect the states it visits together with the actions it takes; each state is encoded by a CNN feature extractor; SHAP attributes the contribution of every feature; and an interpretable surrogate (decision tree, linear regression, or logistic regression) is distilled on the full set of collected (state, action) pairs and evaluated by fidelity and episode return.}
    \label{fig:silver_method_overview}
\end{figure*}

\begin{algorithm}[tb]
\caption{\textsf{ReShap}: Interpretable Policy Distillation}
\label{alg:silver_rl_Guided}
\KwIn{Trained RL policy $\pi$, environment, number of rollout episodes $N$}
\KwData{Interpretable model family $g$ (decision tree / linear / logistic regression)}
\KwOut{Interpretable surrogate policy $\hat{\pi}$}

\textbf{Step 1: Rollout Data Collection}\\
Initialize dataset $D \leftarrow \{\}$\\
\For{$e = 1$ \KwTo $N$}{
    \ForEach{visited state $s_t$ in episode $e$}{
        $a_t \leftarrow \pi(s_t)$ \tcp*[r]{Query RL policy for its action}
        $x_t \leftarrow \text{CNN-Feature}(s_t)$ \tcp*[r]{Compact feature vector}
        $D \leftarrow D \cup \{(x_t, a_t)\}$ \tcp*[r]{Store the labeled state}
    }
}

\textbf{Step 2: SHAP Feature Attribution}\\
\ForEach{$(x_t, a_t) \in D$}{
    $\Phi_{s_t} \leftarrow \text{SHAP}(\pi, x_t)$ \tcp*[r]{Per-feature contributions}
}

\textbf{Step 3: Interpretable Surrogate Distillation}\\
$\hat{\pi} \leftarrow \text{Fit}(g, D)$ \tcp*[r]{Train surrogate on all (state, action) pairs}

\Return{$\hat{\pi}$ as the interpretable surrogate policy}\;

\end{algorithm}

\section{Methodology}
Figure~\ref{fig:silver_method_overview} illustrates the complete pipeline of $\sysname$. We train deep RL models using Stable-Baselines3~\cite{raffin2021stable}, a widely adopted RL framework, and employ a CNN-based feature extractor to obtain compact representations from raw RGB observations. Using the trained deep RL models, we roll out 1,000 episodes per environment to collect the states the policy visits together with the actions it takes, and we conduct SHAP analysis on these states to generate Shapley vectors that capture feature contributions to the agent’s decision-making. We then distill the policy into an interpretable surrogate trained on the collected state-action pairs, and subsequently conduct a human-subject study to evaluate the comprehensibility and trustworthiness of the derived policy.  The overall workflow is summarized in Algorithm~\ref{alg:silver_rl_Guided}.

\subsection{Feature Extraction}
Figure~\ref{fig: Features Extractor} shows our CNN-based feature extractor, which converts stacked RGB Atari frames $(4,84,84)$ into a compact state vector of 16 features (visualized in Appendix~\ref{Appendix: Feature Maps}). Such low-dimensional representations are critical for achieving stable and efficient policy learning, as they distill high-dimensional visual inputs into task-relevant features while minimizing redundancy~\cite{echchahed2025survey}. This design is fully compatible with Stable-Baselines3, using the NatureCNN encoder within the CnnPolicy architecture.

\subsection{SHAP Analysis}
For each state $s \in S$, we compute the Shapley value of each feature $s_i$ under the policy $\pi$, denoted as $\phi_i(v^\pi)$. This value represents the marginal contribution of feature $i$ to the policy’s output in state $s$, and is computed by substituting the characteristic function $v^\pi(C)$ from Eqs.~\eqref{eq:char-deterministic} or~\eqref{eq:char-stochastic} into the Shapley value definition in Eq.~\eqref{eq:shapley-value}.  

The resulting attributions $\phi_i(v^\pi)$ quantify how each feature influences the policy’s action selection. For instance, in a two-action scenario ($A=\{a_1=-1, a_2=1 \}$), a positive $\phi_i(v^\pi)$ indicates that feature $s_i$ drives the policy toward action $a_2$, whereas a negative value favors action $a_1$. Features with identical marginal contributions receive the same Shapley value, revealing symmetries in the policy’s reasoning.

Finally, we aggregate the feature attributions into a per-state vector $\Phi_s = (\phi_1(v^\pi), \dots, \phi_n(v^\pi))$, which summarizes how each feature contributes to the policy’s action selection and provides the attribution basis used to explain the interpretable surrogates constructed next.

\subsection{Interpretable Policy Construction}
We construct the interpretable policy by distilling the deep RL policy $\pi$ into a compact surrogate. The complete procedure is presented in Algorithm~\ref{alg:silver_rl_Guided}. Rather than fitting the surrogate on a small set of specially selected states, we train it on the \emph{full} set of state-action pairs that the policy produces along its visited distribution, which keeps the surrogate faithful to the behavior the agent actually exhibits. The construction proceeds in three stages:
\begin{itemize}[leftmargin=*, topsep=0pt, itemsep=1pt, parsep=0pt]
    \item \textbf{Rollout Data Collection}:
    Roll out the trained policy and record every visited state together with the action the policy takes.

    \item \textbf{SHAP Feature Attribution}:
    Compute a Shapley vector for each collected state to explain how features drive the policy's decisions.

    \item \textbf{Interpretable Models Fitting}:
    Fit decision trees, linear regression, and logistic regression on the full collected dataset to approximate the policy.
\end{itemize}

\subsubsection{Rollout Data Collection}
We roll out the trained policy $\pi$ for $N$ episodes and record every visited state together with the action chosen by the policy, $(s_t, \pi(s_t))$. Each state is passed through the CNN feature extractor to obtain its compact feature vector $x_t$, yielding a labeled dataset
\[
D = \{(x_t, a_t)\}, \qquad a_t = \pi(s_t).
\]
Because the labels are taken directly from the RL policy on the states the policy itself visits, $D$ is behaviorally consistent with $\pi$ by construction and covers the state region the agent actually encounters.

\subsubsection{SHAP Feature Attribution}
For each collected state we compute its Shapley vector $\Phi_s \in \mathbb{R}^n$, as described in the SHAP Analysis section, where each entry quantifies the marginal contribution of a feature to the policy's decision. These attributions are used to analyze and explain the policy's feature dependence and to make the resulting surrogates human-readable; they are not used to sub-select the training states.

\subsubsection{Interpretable Models Fitting}
We fit an interpretable surrogate $\hat{\pi}$ on the full labeled dataset $D$. Unlike \citeauthor{li2025explainability}~\shortcite{li2025explainability}, who employ linear regression exclusively, we explore three interpretable model families to capture different aspects of policy behavior: a decision tree exposes hierarchical if--then reasoning patterns, linear regression yields direct coefficient-based explanations, and logistic regression represents softmax probabilities among actions. Fitting on the full visited state-action distribution, rather than on a sparse set of selected points, allows the surrogate to approximate the policy across the states the agent actually encounters.

\subsection{Human Study}
To assess the interpretability and trustworthiness of our framework, we conduct a human-subject study, following recent XRL evaluation standards~\cite{kohler2025evaluating,alabdulkarim2025experiential}. The study consists of two tasks designed to measure participants’ ability to comprehend and trust interpretable policy.

\subsubsection{Policy Comprehension Task} 
Participants are presented with selected states from Atari environments, together with the corresponding interpretable policy. They are asked to manually infer which action the agent would take based on interpretable policy, following the policy interpretability protocol proposed by \citeauthor{kohler2025evaluating}~\shortcite{kohler2025evaluating}. This task evaluates how effectively humans can understand and predict the agent’s actions based on interpretable policy.

\subsubsection{Trust Calibration Task} 
Participants are presented with paired gameplay clips of agents acting under either the original policy or a fixed interpretable policy. They are asked to rate how much they trust each agent’s decisions using a 5-point Likert scale, adapted from the interpretability evaluation survey implemented by \citeauthor{sarch2025grounded}~\shortcite{sarch2025grounded}. This task assesses how reliably and consistently humans perceive the surrogate policy compared to the original policy.

\begin{figure*}[t]
    \centering
    \subfloat[MsPacman DQN -- SHAP value]{
        \includegraphics[width=0.32\textwidth]{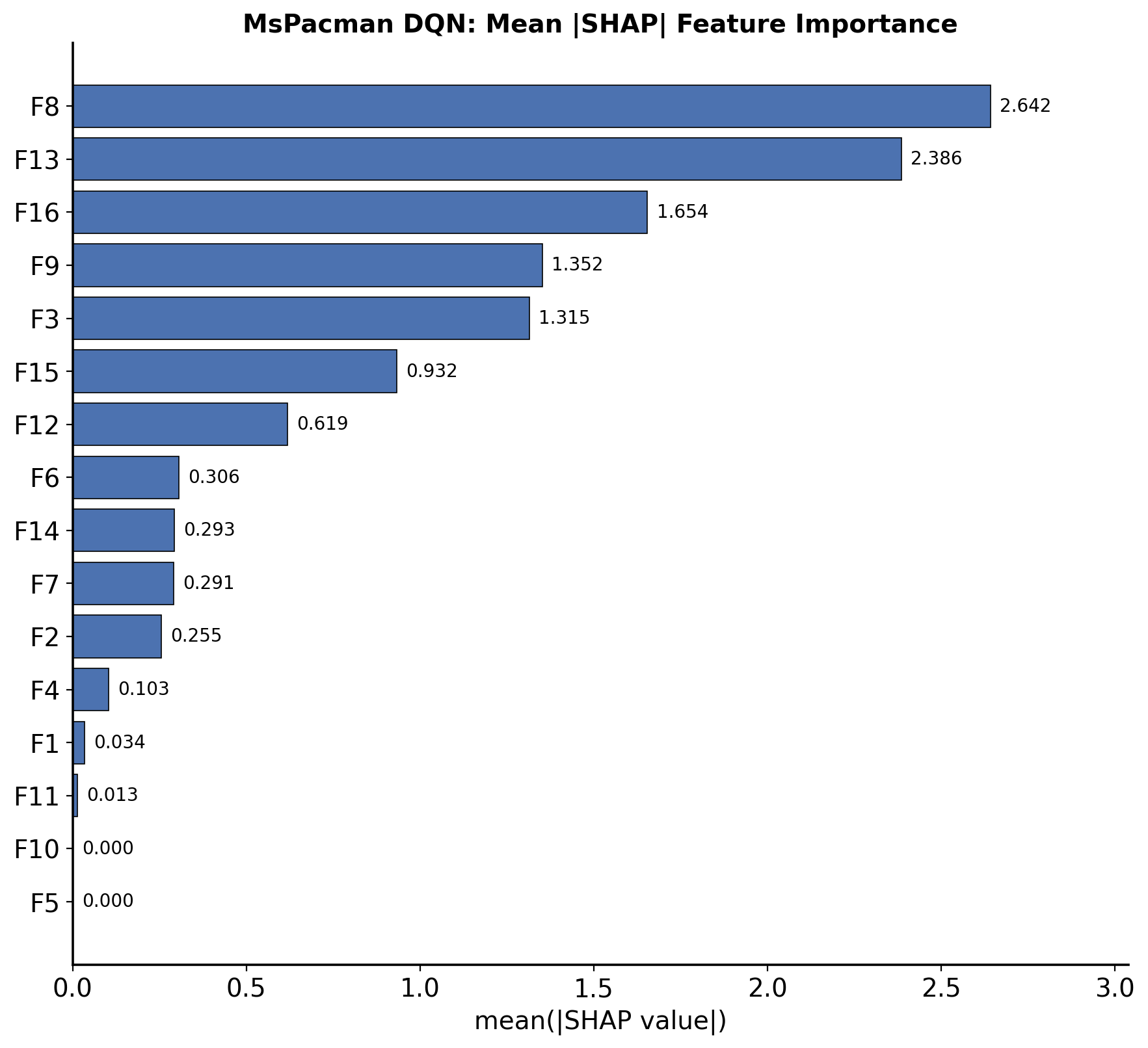}
    }
    \hfill
    \subfloat[MsPacman A2C -- SHAP value\label{fig:MsPacman_A2C_SHAP}]{
        \includegraphics[width=0.32\textwidth]{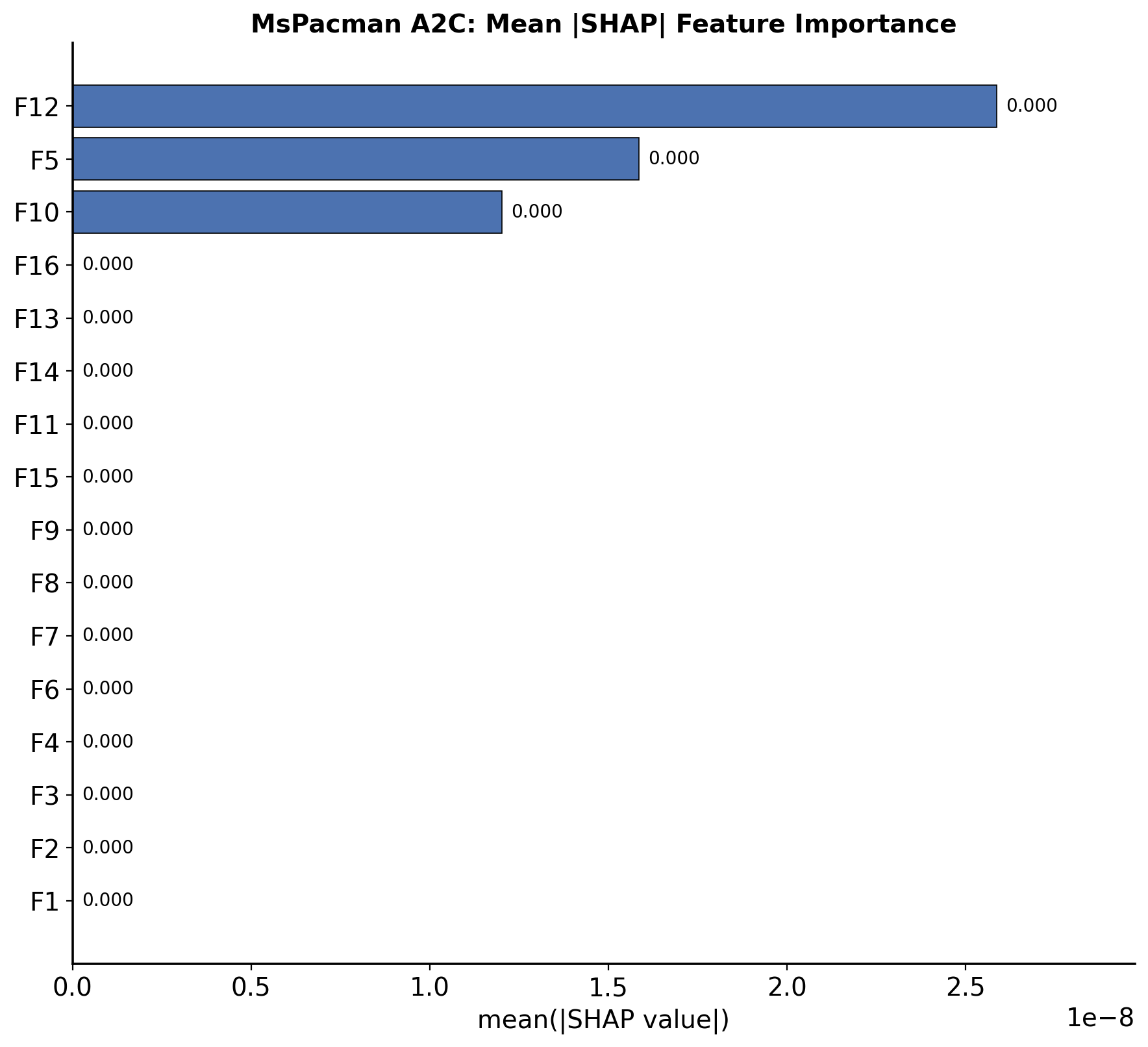}
    }
    \hfill
    \subfloat[MsPacman PPO -- SHAP value\label{fig:MsPacman_PPO_SHAP}]{
        \includegraphics[width=0.32\textwidth]{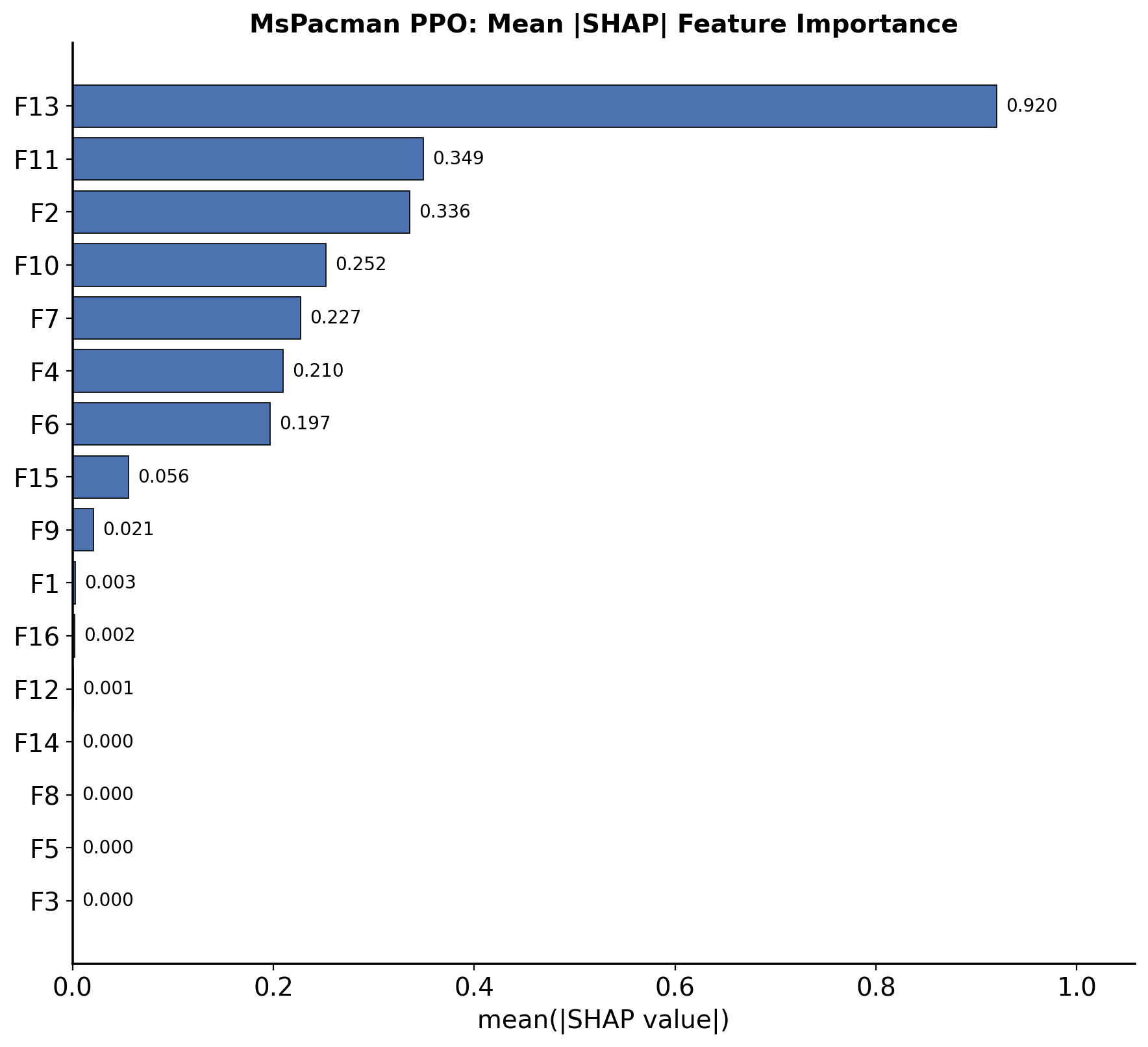}
    }
    \caption{Mean absolute SHAP value of each feature for MsPacman under different algorithms (DQN, A2C, PPO). The A2C rollout is single-action, so its SHAP attributions are negligible (note the axis scale).}
    \label{fig:MsPacman_SHAP_comparison}
\end{figure*} 

\section{Experiments}
To evaluate the effectiveness of \textsf{ReShap}, we perform experiments. Our experiments consist of four steps:
\begin{enumerate}[leftmargin=*, topsep=0pt, itemsep=1pt, parsep=0pt]
    \item Training deep RL agents and collecting trajectories in two Atari environments and one Box2D environment.
    \item Applying SHAP analysis to quantify feature contributions to agents' decisions and to construct interpretable policies.
    \item Comparing the performance of the interpretable policies against their corresponding RL policies.
    \item Evaluating the clarity and trustworthiness of the interpretable policies through human-subject study.
\end{enumerate}
All Atari environments are wrapped using the \texttt{Atari Wrappers} from Stable-Baselines3, which clip rewards to $[-1, 0, 1]$ by its sign for training stability.

\subsection{Baseline}
Baseline experiment is conducted by applying \textsf{ReShap} to the CartPole environment; detailed results are provided in Appendix~\ref{Appendix: Baseline Experiment}.

\subsection{Visualization}
For visualization purposes, we utilize a mutimodal LLM (e.g., Gemini 2.5 Pro) to generate natural language explanations for our interpretable policies. The corresponding prompts for decision tree, linear regression, and logistic regression are provided in Appendices~\ref{Appendix: Decision Tree' Prompt}, ~\ref{Appendix: Linear Regression' Prompt}, ~\ref{Appendix: Logistic Regression' Prompt}.

\begin{figure*}[t]
    \centering
    \subfloat[MsPacman\label{fig:MsPacman_Mean_95_CI}]{
        \includegraphics[width=0.32\textwidth]{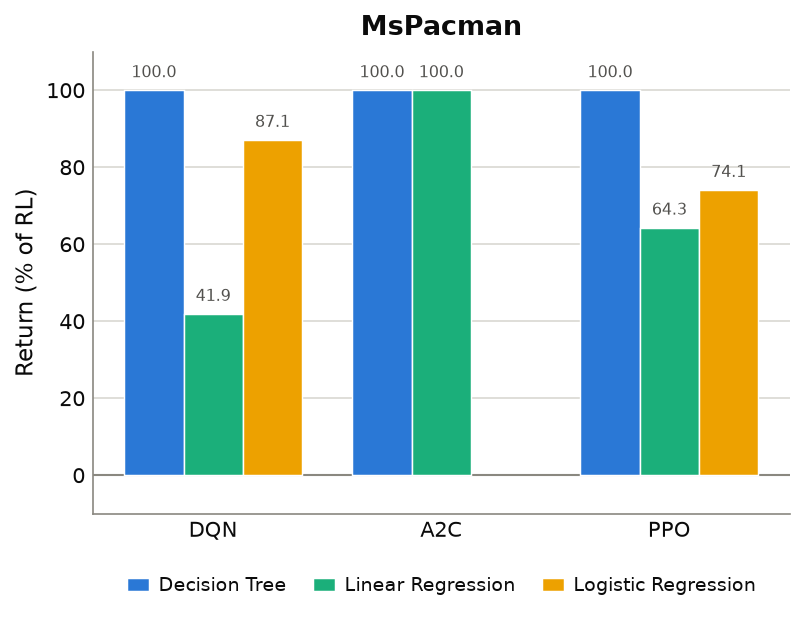}
    }
    \hfill
    \subfloat[RoadRunner\label{fig:RoadRunner_Mean_95_CI}]{
        \includegraphics[width=0.32\textwidth]{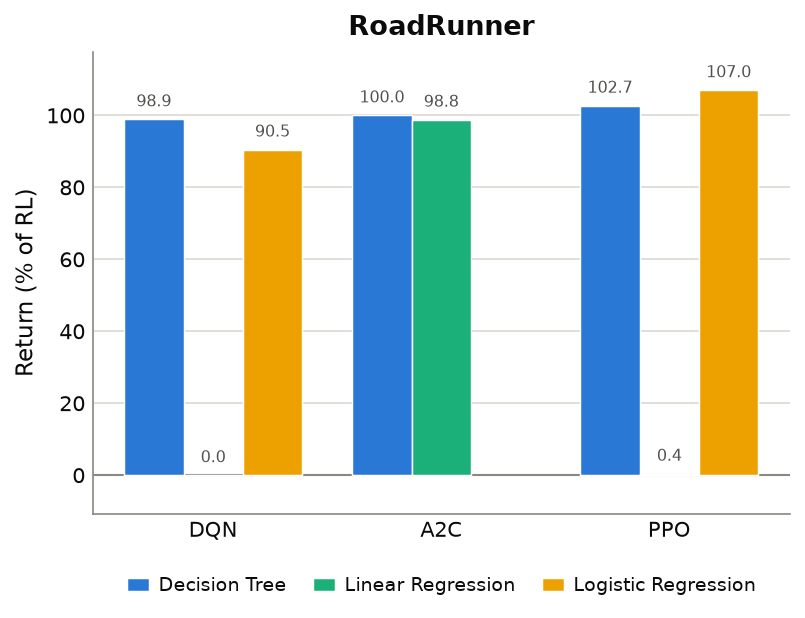}
    }
    \hfill
    \subfloat[LunarLander\label{fig:LunarLander_Mean_95_CI}]{
        \includegraphics[width=0.32\textwidth]{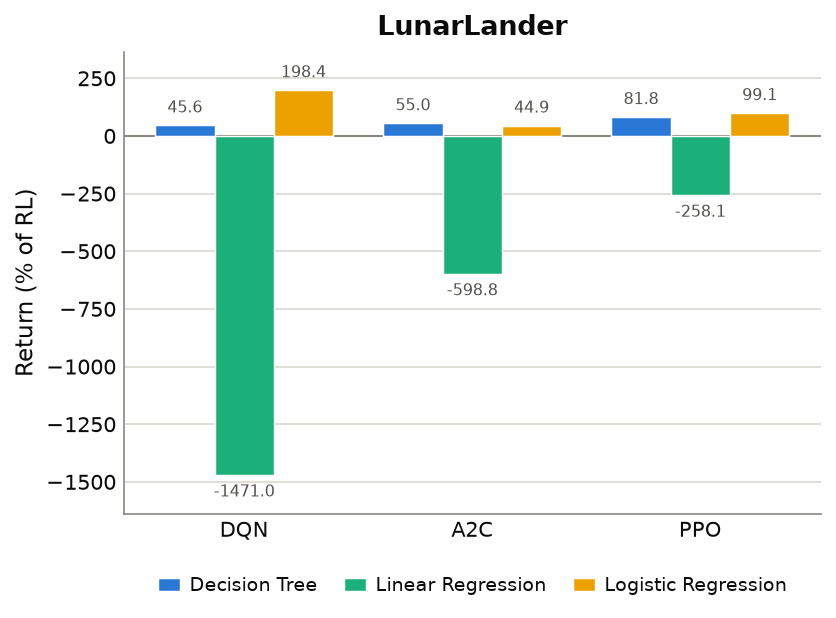}
    }
    \caption{Episode return of the interpretable surrogates as a percentage of the original RL policy's return, under full-distribution distillation, for MsPacman, RoadRunner, and LunarLander. Bars are grouped by RL algorithm (DQN, A2C, PPO), one bar per surrogate; the A2C logistic-regression bar is omitted where the rollout contains fewer than two distinct actions.}
    \label{fig:Performance_Comparison}
\end{figure*}

\subsection{Atari \& Box2D}

\subsubsection{MsPacman}
The MsPacman environment is a maze-based Atari game where the agent navigates to collect rewards and avoid ghosts. The observation space consists of high-dimensional RGB frames showing MsPacman, pellets, fruits, and ghosts. The action space includes 9 discrete movements with or without the fire button. The agent gains rewards for consuming pellets and fruits, and receives a penalty when caught by a ghost. The goal is to maximize the cumulative score through strategic navigation.

We sample state distributions from 1000 episodes collected by each deep RL algorithm. For every sampled state, we compute the Shapley values of its features using Eq.~\eqref{eq:shapley-value}, thereby obtaining a Shapley vector that quantifies the contribution of individual features to the policy’s decision. Figure~\ref{fig:MsPacman_SHAP_comparison} presents the mean absolute Shapley value of each of the 16 features for Deep Q-Network (DQN), Proximal Policy Optimization (PPO), and Advantage Actor-Critic (A2C), respectively.

We then roll out each RL policy and record all visited states together with the policy's chosen actions, forming a labeled dataset of $(\text{state}, \pi(\text{state}))$ pairs. Finally, we fit decision trees, linear regression, and logistic regression on this full dataset to obtain interpretable policies for DQN, PPO, and A2C.

\paragraph{Decision Tree.} 
Figures~\ref{fig:MsPacman_ppo_decision_tree}, \ref{fig:MsPacman_a2c_decision_tree}, and \ref{fig:MsPacman_dqn_decision_tree} illustrate decision trees for PPO, A2C, and DQN, respectively. 

Taking PPO as an example, consider an example state $S_1$ in Figure~\ref{fig:MsPacman_PPO_feature_map}. Starting from the root node of Figure~\ref{fig:MsPacman_ppo_decision_tree}, the decision process proceeds as follows, with the traced path highlighted in red:

\begin{enumerate}[leftmargin=*, topsep=0pt, itemsep=1pt, parsep=0pt]
  \item At the root node, check $F_{13} \leq 0.81$. \\
        Since $6.81 > 0.81$, move to the \textbf{right child}.
  \item Next, check $F_{7} \leq 1.97$. \\
        Since $0.00 \leq 1.97$, move to the \textbf{left child}.
  \item Then, check $F_{11} \leq 20.08$. \\
        Since $11.71 \leq 20.08$, move to the \textbf{left child} and reach a leaf:
        $\boxed{\text{class} = a_7}$, which corresponds to  $\boxed{\text{action} = 7}$.
\end{enumerate}

Appendix~\ref{Appendix: Natural Language Explanation Synthesis for Decision Tree} shows decision tree's natural language explanation in MsPacman PPO. 

\paragraph{Linear Regression.}
Table~\ref{tab:MsPacman Linear Regression} illustrates the interpretable policy boundaries derived from linear regression for PPO, A2C, and DQN. Here, $x_{1}$ through $x_{16}$ represent the 16 state features used in the analysis.

Taking PPO as an example, consider an example state $S_2$ in Figure~\ref{fig:MsPacman_PPO_feature_map} whose nonzero features are $x_{6}=8.85$, $x_{9}=4.40$, $x_{11}=8.25$, and $x_{13}=12.54$. Substituting into the decision boundary in Table~\ref{tab:MsPacman Linear Regression}, we obtain:

\[
\begin{aligned}
f(x) &= 5.68 - (4.19 \times 10^{-1}) \times 8.85 \\
&\quad + (1.90 \times 10^{-1}) \times 4.40 + (6.51 \times 10^{-2}) \times 8.25 \\
&\quad + (2.72 \times 10^{-1}) \times 12.54 \approx 6.76.
\end{aligned}
\]

Rounding to the nearest integer and clipping to the valid action range yields $\boxed{f(x) = 7}$, which corresponds to $\boxed{\text{action} = 7}$.

Although actions are discrete, linear regression serves as a continuous decision boundary function. Its outputs are discretized (by rounding to the nearest valid action index) to recover action indices, yielding a concise and interpretable mapping from state features to actions.

For linear regression's natural language explanation in MsPacman PPO, please refer to Appendix~\ref{Appendix: Natural Language Explanation Synthesis for Linear Regression}.

\paragraph{Logistic Regression.}
Table~\ref{tab:MsPacman_PPO_LogReg}, Table~\ref{tab:MsPacman_A2C_LogReg}, and Table~\ref{tab:MsPacman_DQN_LogReg} illustrate the interpretable policy boundaries derived from logistic regression for PPO, A2C, and DQN.

Taking PPO as an example, consider an example state $S_3$ in Figure~\ref{fig:MsPacman_PPO_feature_map}. Substituting its feature values into the corresponding decision boundaries in Table~\ref{tab:MsPacman_PPO_LogReg}, we obtain the following logits:
\[
\begin{aligned}
f_{0}(x) &\approx 3.84, \quad f_{6}(x) \approx 4.89, \\
f_{7}(x) &\approx -12.02, \quad f_{8}(x) \approx 3.29.
\end{aligned}
\]

Applying the softmax function to these logits converts them into normalized probabilities:
\[
P(a=i \mid x) = \frac{e^{f_i(x)}}{\sum_{j} e^{f_j(x)}},
\]
which here gives $P(a{=}0)\approx0.23$, $P(a{=}6)\approx0.65$, $P(a{=}7)\approx0.00$, and $P(a{=}8)\approx0.13$. The predicted action is the one with the highest probability:
\[
\hat{a} = \arg\max_{i} P(a=i \mid x).
\]
For this state, the highest probability corresponds to $\boxed{f_{6}(x)}$, indicating that the model predicts $\boxed{\text{action} = 6}$.

For logistic regression's natural language explanation in MsPacman PPO, please refer to Appendix~\ref{Appendix: Natural Language Explanation Synthesis for Logistic Regression}.

\subsubsection{RoadRunner}
We also evaluated our framework on RoadRunner. Details of the experiment are provided in Appendix~\ref{Appendix: Experiment on RoadRunner}.

\subsubsection{LunarLander}
Beyond the Atari domains, we additionally evaluate \textsf{ReShap} on the Box2D LunarLander environment. Details of the experiment are provided are provided in Appendix~\ref{Appendix: Experiment on LunarLander}.

\subsection{Fidelity Score}
Fidelity measures how well the surrogate policy covers the original policy~\cite{altmann2025surrogate}. The fidelity function proposed by \citeauthor{li2025explainability}~\shortcite{li2025explainability} further quantifies the behavioral difference between the interpretable policy and the original policy:

\begin{equation}
F(\pi_{\text{interp}}, \pi_{\text{orig}}) =
\frac{1}{|S|} \sum_{s \in S} \mathds{1}\{\pi_{\text{interp}}(s) = \pi_{\text{orig}}(s)\},
\label{eq:fidelity}
\end{equation}
where the original policy is treated as the ground truth.

Building on this notion, we define our fidelity score explicitly as the percentage of states where the interpretable and original policies output the same action. 

A detailed breakdown of fidelity scores by surrogate model and RL algorithm across all three environments is provided in Appendix~\ref{Appendix: Fidelity}.

\subsection{Human Study}
We conduct a human-subject study with 50 participants. To simplify evaluation and minimize cognitive burden, the study is restricted to the MsPacman environment and a single interpretable model. Since the decision tree outperforms the PPO policy and provides the clearest if--then style rules, we select it for evaluation.

\paragraph{Design.}
Participants are presented with representative MsPacman states, together with the decision tree. Their task is to infer the action chosen by the decision tree. While doing so, we collect their action inferences, response times, and subjective trust ratings.

\paragraph{Metrics.}
We measure three aspects:
\begin{itemize}[leftmargin=*, topsep=0pt, itemsep=1pt, parsep=0pt]
    \item \textbf{Response Time}: The average time taken to make an action inference, reflecting the efficiency of interpretability.
    \item \textbf{Accuracy}: The proportion of participant-inferred actions that matches the actual decision-tree outputs, reflecting the correctness of human interpretation.
    \item \textbf{Trust Ratings}: After completing all 4 trials, participants rate their overall trust in the decision-tree surrogate on a 5-point Likert scale, where:
    \begin{itemize}
        \item 1 = Very Low Trust (I would not rely on this model’s decisions at all)
        \item 2 = Low Trust (I have limited confidence; decisions often appear unreliable)
        \item 3 = Neutral (I am undecided; the model’s decisions sometimes make sense, sometimes not)
        \item 4 = High Trust (I generally have confidence in the model’s decisions, though not absolute)
        \item 5 = Very High Trust (I would fully rely on this model’s decisions with strong confidence)
    \end{itemize}
\end{itemize}

Appendix~\ref{Appendix: Results of Human Study} shows results of human-subject study. 

\section{Discussion}
Figure~\ref{fig:Performance_Comparison} reports the episode return of each interpretable surrogate as a percentage of its RL policy's return, grouped by algorithm. We go through the three environments in turn.

\paragraph{MsPacman.}
The decision tree matches the PPO and DQN return almost exactly. Logistic regression retains $74.1\%$ under PPO and $87.1\%$ under DQN, while linear regression trails at $64.3\%$ and $41.9\%$. The A2C policy takes a single action throughout its rollout, so every surrogate matches it trivially and this group carries no information.

\paragraph{RoadRunner.}
The pattern repeats. Decision trees keep $98.9\%$ of the DQN return and $102.7\%$ of the PPO return, logistic regression keeps $90.5\%$ and $107.0\%$, and linear regression collapses to essentially zero. With 18 discrete actions, mapping a single continuous output onto that many action indices simply does not work. The A2C group is again single-action and uninformative.

\paragraph{LunarLander.}
Returns spread out more here. Under PPO the logistic regression retains $99.1\%$ and the decision tree retains $81.8\%$, while linear regression collapses to $-258.1\%$ once its per-step errors accumulate over an episode and the lander crashes. The weaker agents show the same failure: under A2C the tree retains $55.0\%$, logistic regression $44.9\%$, and linear regression $-598.8\%$; under DQN the tree retains $45.6\%$, logistic regression overshoots at $198.4\%$, and linear regression drops to $-1471.0\%$. On DQN, logistic regression clearly outperforms the tree, so we do not claim the tree is the stronger surrogate in this environment. Because LunarLander's features are physical quantities such as velocity, angle, and leg contact rather than CNN embeddings, the surrogate equations can be read directly in those terms, which helps offset the semantic opacity of the Atari feature extractor.

\paragraph{Summary.}
Across all three environments, decision trees and logistic regression both recover most of the RL return, while linear regression is consistently the weakest surrogate and breaks down as the action space grows. Neither the tree nor logistic regression dominates the other; logistic regression is strongest on LunarLander. Fidelity and return also do not always move together. On LunarLander DQN, the tree agrees with the policy on $78.0\%$ of states versus $73.0\%$ for logistic regression, yet its return is far lower ($45.6\%$ versus $198.4\%$ of the RL return). We therefore treat return as the primary task-level measure and fidelity as a secondary diagnostic.

\section{Limitations}
Our work has limitations that may need to be addressed in the future work. For details, please refer to Appendix~\ref{Appendix: Details of Limitations}.

\section{Conclusion}
In this paper, we propose \textsf{ReShap}, a framework designed to generate interpretable policies for deep RL algorithms. Experimental results demonstrate that our method achieves performance comparable to deep RL algorithms in high-dimensional and multi-action environments, while preserving human comprehensibility and trust.

Potential future work includes: (1) refining the feature extraction process to assign semantic meaning to each state feature, (2) extending the framework to LLMs' environments, particularly in the context of RL fine-tuning, and (3) conducting human-subject studies with diverse interpretable models to systematically compare their performance on comprehension and trust, and (4) deploying the compact surrogates under privacy-preserving secure inference~\cite{song2026oblivigate,wang2025privgnn}, where their low model complexity can substantially reduce cryptographic overhead compared with deep policies.

\bibliographystyle{named}
\bibliography{mybib}

\clearpage
\appendix
\raggedbottom  %
\setcounter{topnumber}{3}
\setcounter{dbltopnumber}{3}
\setcounter{totalnumber}{5}
\renewcommand{\topfraction}{0.92}
\renewcommand{\bottomfraction}{0.8}
\renewcommand{\textfraction}{0.07}
\renewcommand{\floatpagefraction}{0.6}
\renewcommand{\dbltopfraction}{0.92}
\renewcommand{\dblfloatpagefraction}{0.6}

\setlength{\floatsep}{6pt plus 1pt minus 2pt}
\setlength{\textfloatsep}{7pt plus 2pt minus 2pt}
\setlength{\intextsep}{6pt plus 1pt minus 2pt}
\setlength{\dblfloatsep}{6pt plus 1pt minus 2pt}
\setlength{\dbltextfloatsep}{7pt plus 2pt minus 2pt}
\setlength{\abovecaptionskip}{3pt}
\setlength{\belowcaptionskip}{0pt}
\setlength{\abovedisplayskip}{6pt plus 1pt minus 2pt}
\setlength{\belowdisplayskip}{6pt plus 1pt minus 2pt}
\setlength{\abovedisplayshortskip}{3pt plus 1pt minus 1pt}
\setlength{\belowdisplayshortskip}{4pt plus 1pt minus 1pt}

\makeatletter
\setlength{\@fptop}{0pt}
\setlength{\@fpsep}{6pt plus 1pt}
\setlength{\@fpbot}{0pt plus 1fil}
\setlength{\@dblfptop}{0pt}
\setlength{\@dblfpsep}{6pt plus 1pt}
\setlength{\@dblfpbot}{0pt plus 1fil}
\makeatother

\newcolumntype{L}[1]{>{\raggedright\arraybackslash}p{#1}}

\begin{table*}[t]
\section{Comparison of Representative XRL Methods}
\label{Appendix: Comparison between SILVER and SILVER with RL-Guided Labeling}
\centering
\renewcommand{\arraystretch}{1.5} %
\small %

\begin{tabularx}{\textwidth}{L{2.8cm} L{3.5cm} L{2cm} X L{2.2cm} L{2.2cm}}
\toprule
\textbf{Schemes} & \textbf{Target Environments} & \textbf{Action Space} & \textbf{Decision Boundary Modeling} & \textbf{Action Labeling} & \textbf{Policy-explanation Alignment} \\
\midrule

{Causal Lens}~\cite{madumal2020explainable} & 
Low-dimensional control and box2d (e.g., Mountain Car, Lunar Lander) & 
Discrete (low-cardinality) & 
None (Structural causal model only) & 
Not applicable & 
Not applicable \\

{SVERL}~\cite{beechey2023explaining} & 
Low-dimensional discrete states (e.g., Taxi, Minesweeper) & 
Discrete (low-cardinality) & 
None (Shapley attribution only) & 
Not applicable & 
Not applicable \\

{DISTILL2EXPLAIN}
~\cite{gokhale2024distill2explain} & 
Low-dimensional control (e.g., Simulation of home energy management) & 
Discrete (low-cardinality) & 
None (Policy distillation only) & 
Not applicable & 
Behaviorally aligned \\

{SILVER}~\cite{li2025explainability} & 
Low-dimensional control (e.g., CartPole, Mountain Car) & 
Discrete (binary) & 
Geometric distance in Shapley space & 
Implicit cluster-based & 
Potentially misaligned \\

\midrule
\rowcolor{gray!12} 
\textbf{\textsf{ReShap}} (Ours) &
\textbf{High-dimensional visual} (e.g., MsPacman, RoadRunner) &
\textbf{Discrete (high-cardinality)} &
\textbf{Full-distribution policy distillation} &
\textbf{Direct policy actions $\pi(s)$} &
\textbf{Behaviorally consistent} \\
\bottomrule
\end{tabularx}
\caption{Comparison of representative XRL methods}
\label{tab:appendix_comparison_fixed}

\vspace{1.2em}
\begin{minipage}{\textwidth}
\setlength{\parindent}{0pt}
\setlength{\parskip}{0.5em}
\noindent\textbf{Policy--explanation alignment.}
This column indicates whether the explanations or surrogate models produced by a given method faithfully reflect the action-selection behavior of RL policy. Specifically, alignment is assessed at the decision level, i.e., whether the actions prescribed by the explanation coincide with those taken by the RL policy in the same states.

{Causal Lens} offers causal-structural explanations of agent behavior without constructing an explicit interpretable policy; thus, policy--explanation alignment is not applicable.

For {SILVER}, action labels are inferred implicitly through geometric clustering in the Shapley space, which may not correspond exactly to the actions selected by the RL policy, leading to potential misalignment.

{DISTILL2EXPLAIN} constructs intrinsically interpretable policies (differentiable decision tree), where the policy itself serves as the explanation, resulting in behavioral alignment at the policy level.

{SVERL} focuses on post-hoc Shapley-based attribution of feature contributions and does not produce an executable policy or decision rule; therefore, policy--explanation alignment is not applicable.

In contrast, \textsf{ReShap} distills the surrogate from the full set of states the RL policy visits, with every action label taken directly from the RL policy, ensuring that the resulting interpretable surrogate remains behaviorally consistent with the RL policy.

Here, \emph{behaviorally aligned} means that the policy is interpretable by design, so the explanation naturally matches the policy’s behavior. In contrast, \emph{behaviorally consistent} means that the explanation is constructed to follow the RL policy, with action labels taken from the RL policy and consistency verified empirically.
\end{minipage}
\end{table*}
\FloatBarrier

\clearpage
\begin{figure*}[t]
    \section{Gameplay Clips and Feature Extractor}
    \label{Appendix: Game Clips and Feature Extractor}

    This appendix includes the gameplay clips used in the human study and the CNN feature extractor for Atari observations. Figure~\ref{fig:Gameplay_Clips} shows representative PPO clips, and Figure~\ref{fig: Features Extractor} shows the CNN that maps stacked RGB frames to a $16$-dimensional \textsf{ReShap} feature vector.

    \centering
    \subfloat[Gameplay clips of PPO in MsPacmanNoFrameskip-v4%
    \label{fig:MsPacman_gameclips}]{
        \includegraphics[width=0.44\linewidth]{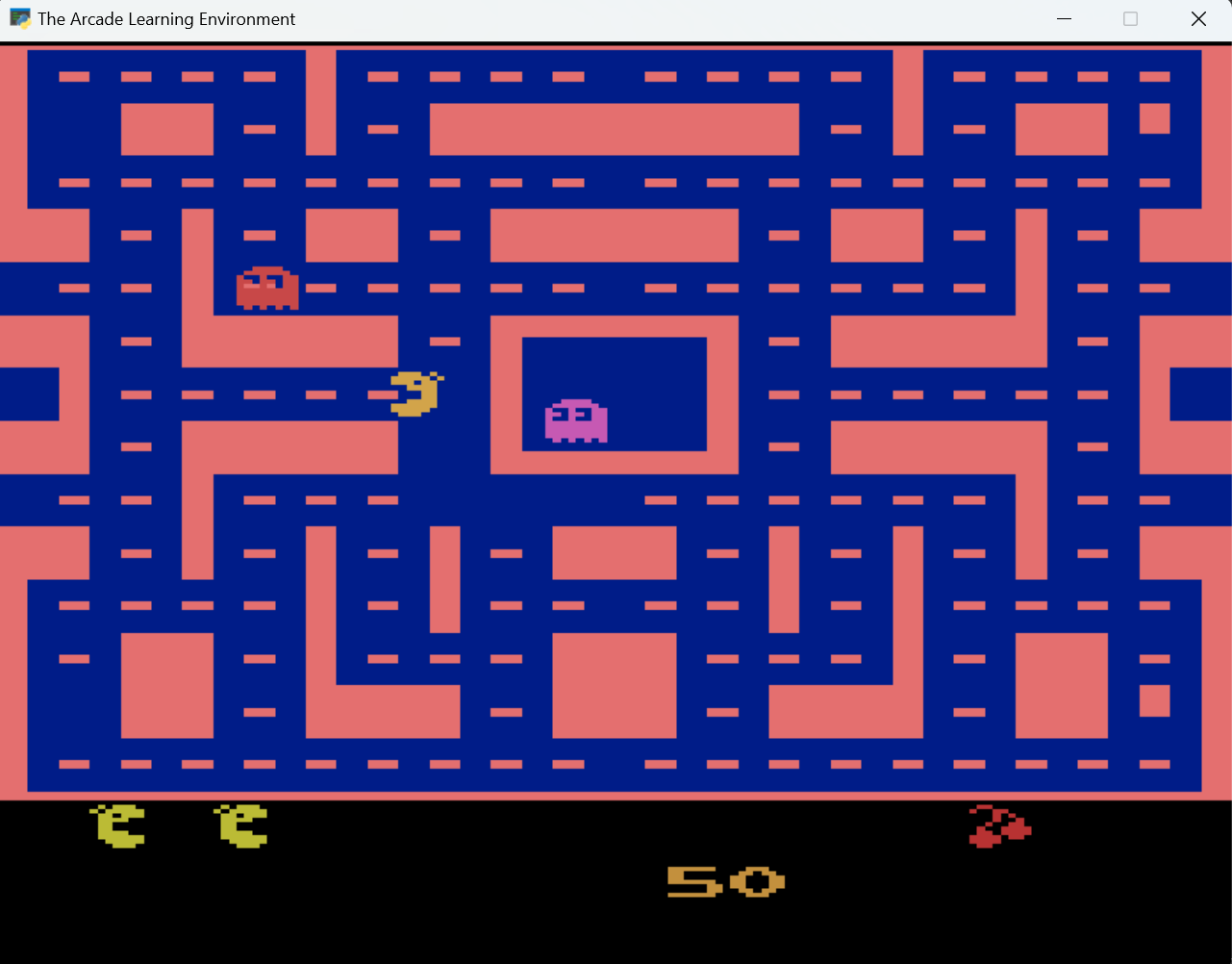}
    }
    \hfill
    \subfloat[Gameplay clips of PPO in RoadRunnerNoFrameskip-v4%
    \label{fig:RoadRunner_gameclips}]{
        \includegraphics[width=0.44\linewidth]{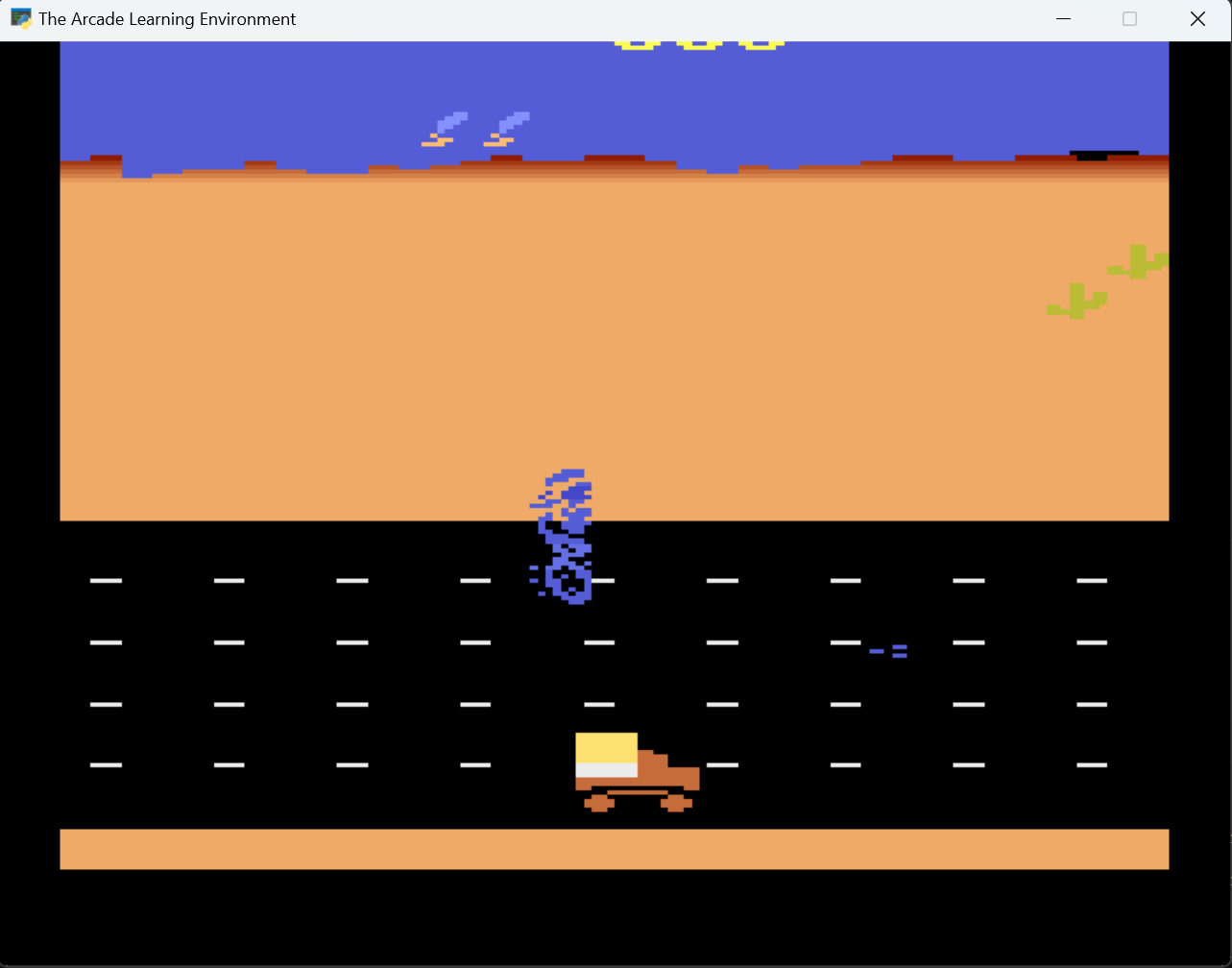}
    }
    \caption{Gameplay clips of PPO agents in Atari environments}
    \label{fig:Gameplay_Clips}
\end{figure*}

\begin{figure*}[t]
    \vspace{4mm}
    \centering
    \includegraphics[width=1\linewidth]{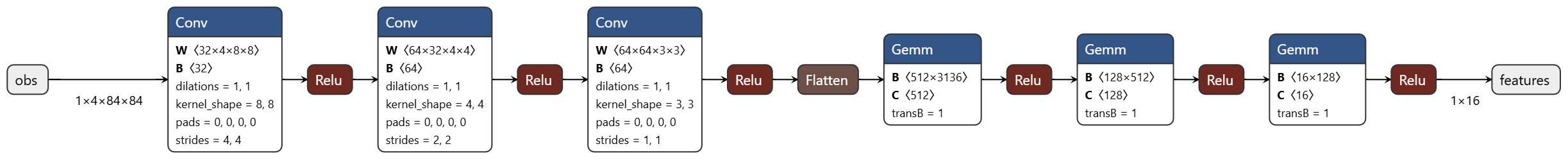}
    \caption{CNN-based feature extractor}
    \label{fig: Features Extractor}
\end{figure*}
\FloatBarrier

\begin{figure*}[t]
    \section{Feature Maps ($F$ = $Feature$, $Act$ = $Action$)}
    \label{Appendix: Feature Maps}
    This appendix visualizes the $16$-dimensional CNN features extracted by \textsf{ReShap} in both Atari environments. For each environment and RL algorithm, heatmaps report feature values for ten rollout states, annotated with the taken action ($s,a$). In the PPO panels, the first three columns are the worked-example states $S_1$ (decision tree), $S_2$ (linear regression), and $S_3$ (logistic regression) used in the corresponding walkthroughs, and the remaining seven columns are evenly spaced along the rollout; in the A2C and DQN panels all ten states are sampled along the rollout. These features (not raw pixels) are the inputs to the surrogate models.

    \centering
    \subfloat[MsPacman PPO]{\includegraphics[width=0.52\linewidth]{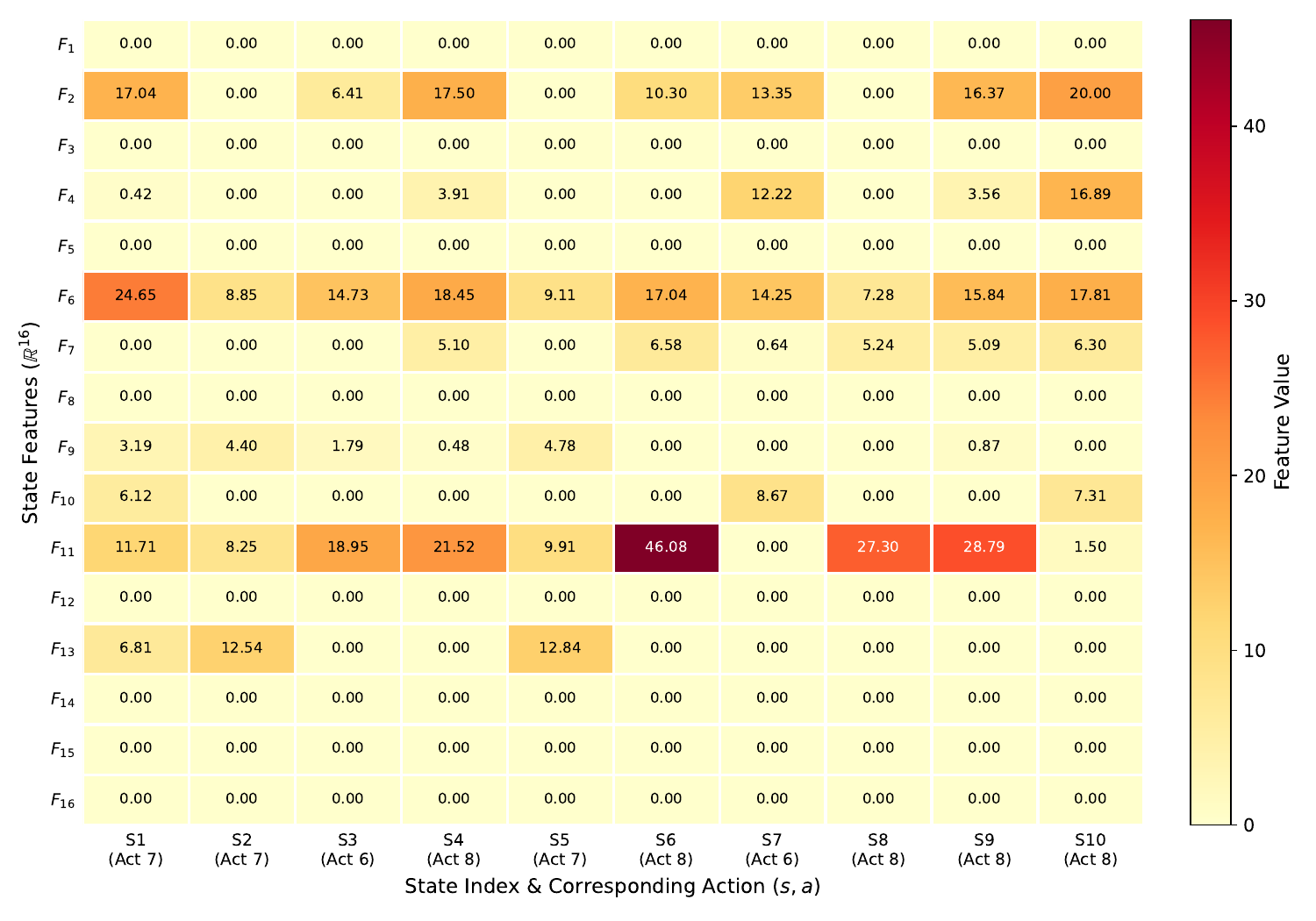}\label{fig:MsPacman_PPO_feature_map}}
    \hfill
    \subfloat[MsPacman A2C]{\includegraphics[width=0.52\linewidth]{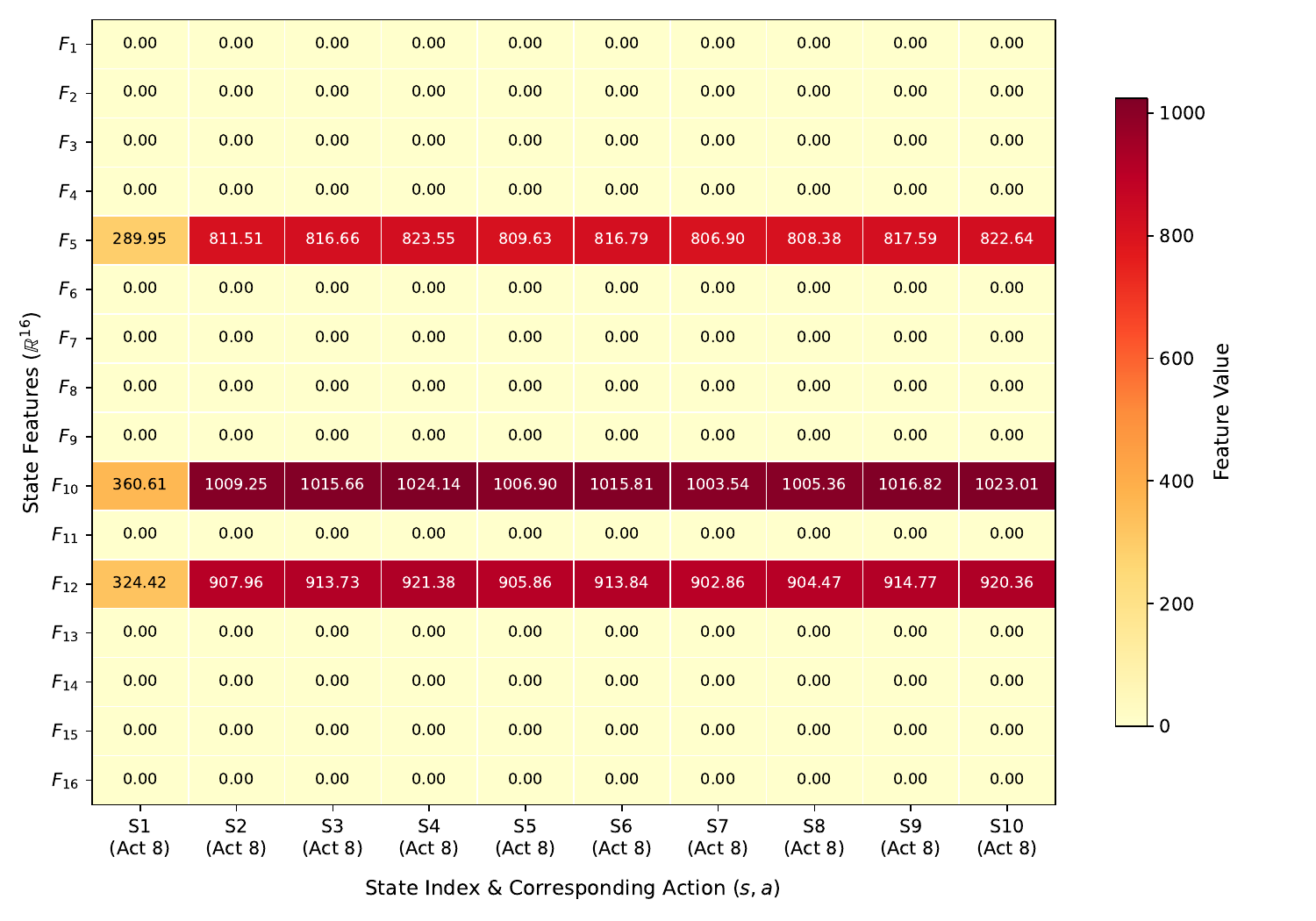}\label{fig:MsPacman_A2C_feature_map}}
    \par\smallskip
    \subfloat[MsPacman DQN]{\includegraphics[width=0.52\linewidth]{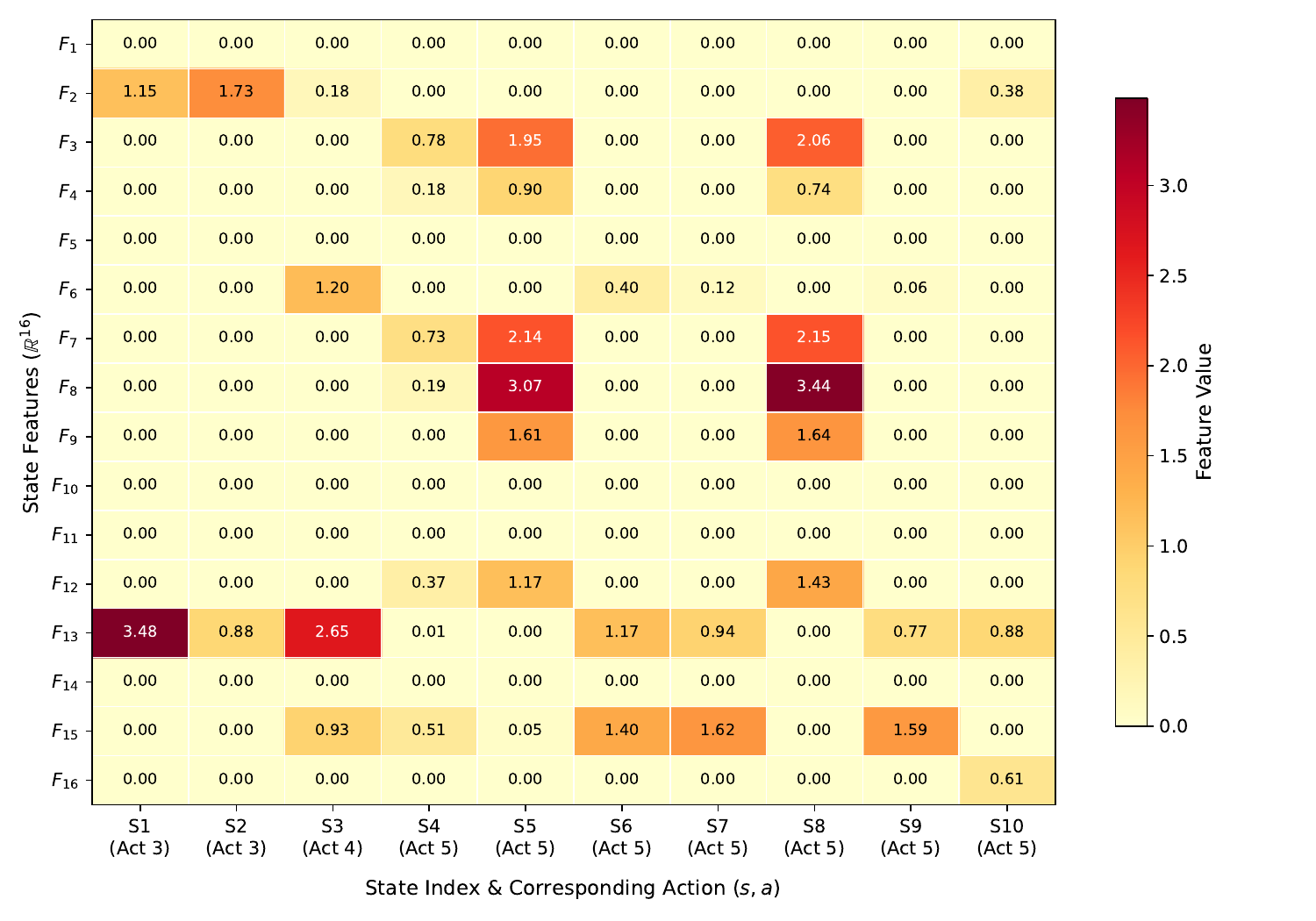}\label{fig:MsPacman_DQN_feature_map}}
    \caption{MsPacman feature maps: the $16$ CNN features for ten states sampled along each RL policy's rollout, annotated with the policy's action.}
    \label{fig:MsPacman_feature_maps}
\end{figure*}

\begin{figure*}[t]
    \centering
    \subfloat[RoadRunner PPO]{\includegraphics[width=0.52\linewidth]{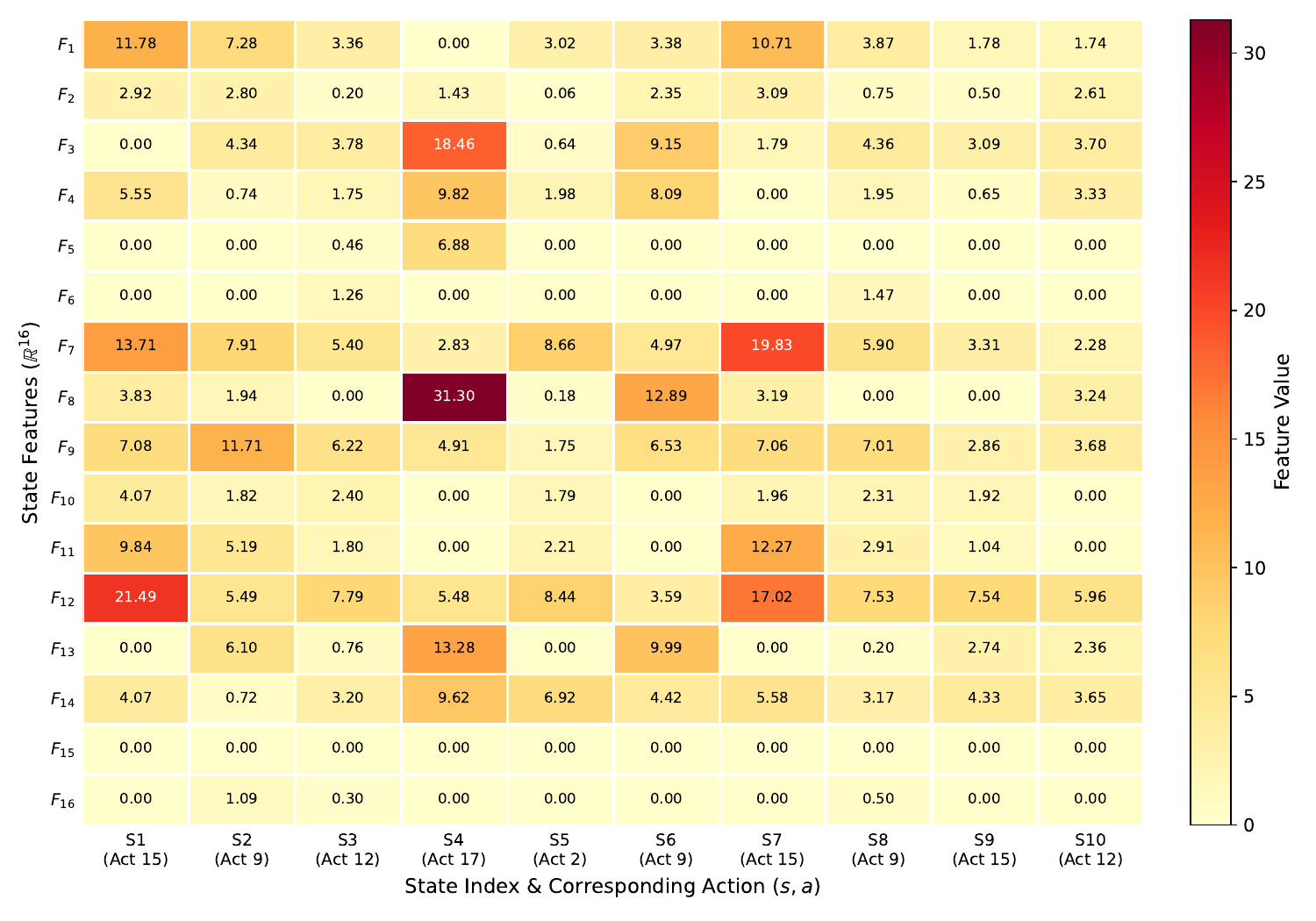}\label{fig:RoadRunner_PPO_feature_map}}
    \hfill
    \subfloat[RoadRunner A2C]{\includegraphics[width=0.52\linewidth]{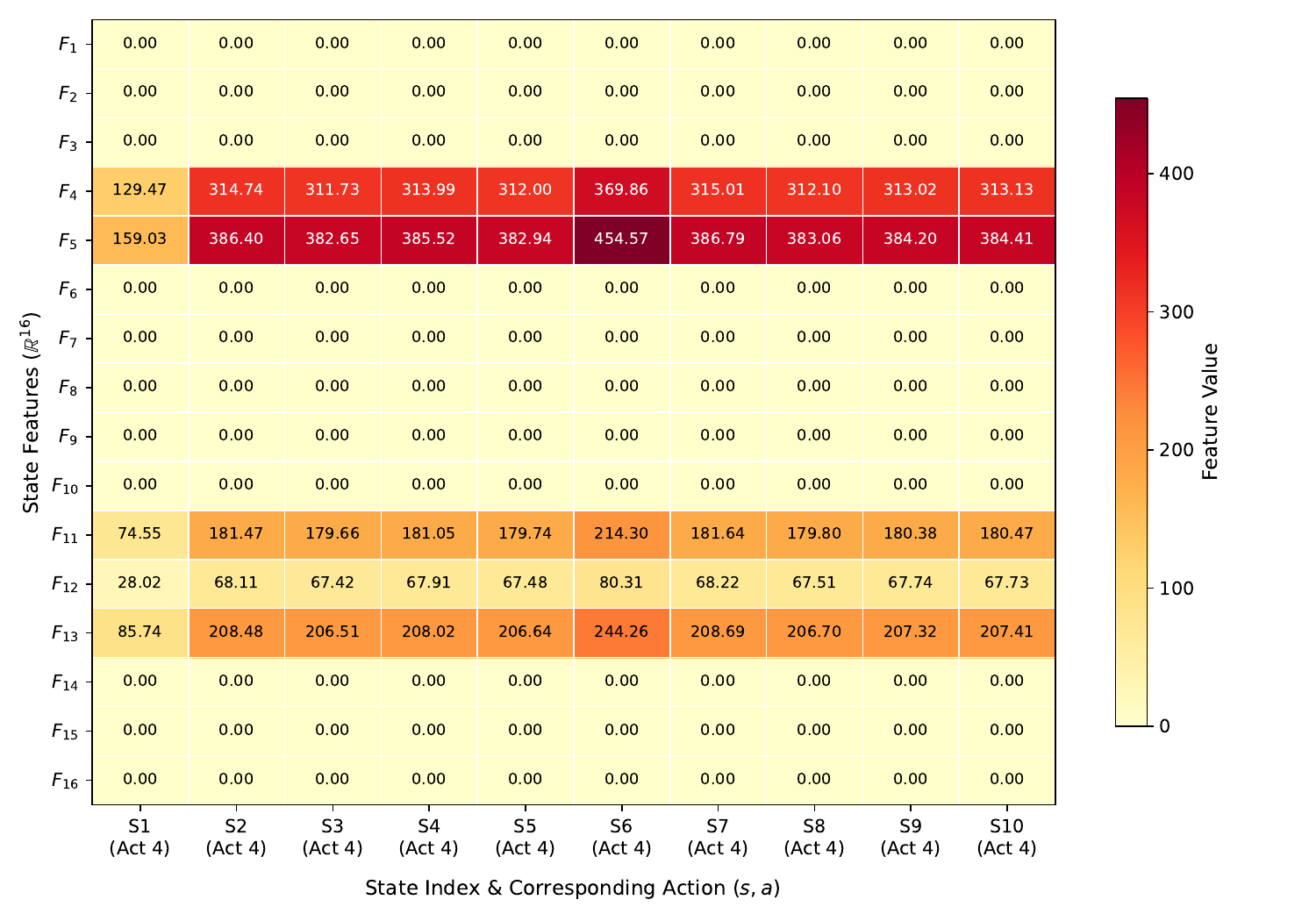}\label{fig:RoadRunner_A2C_feature_map}}
    \par\smallskip
    \subfloat[RoadRunner DQN]{\includegraphics[width=0.52\linewidth]{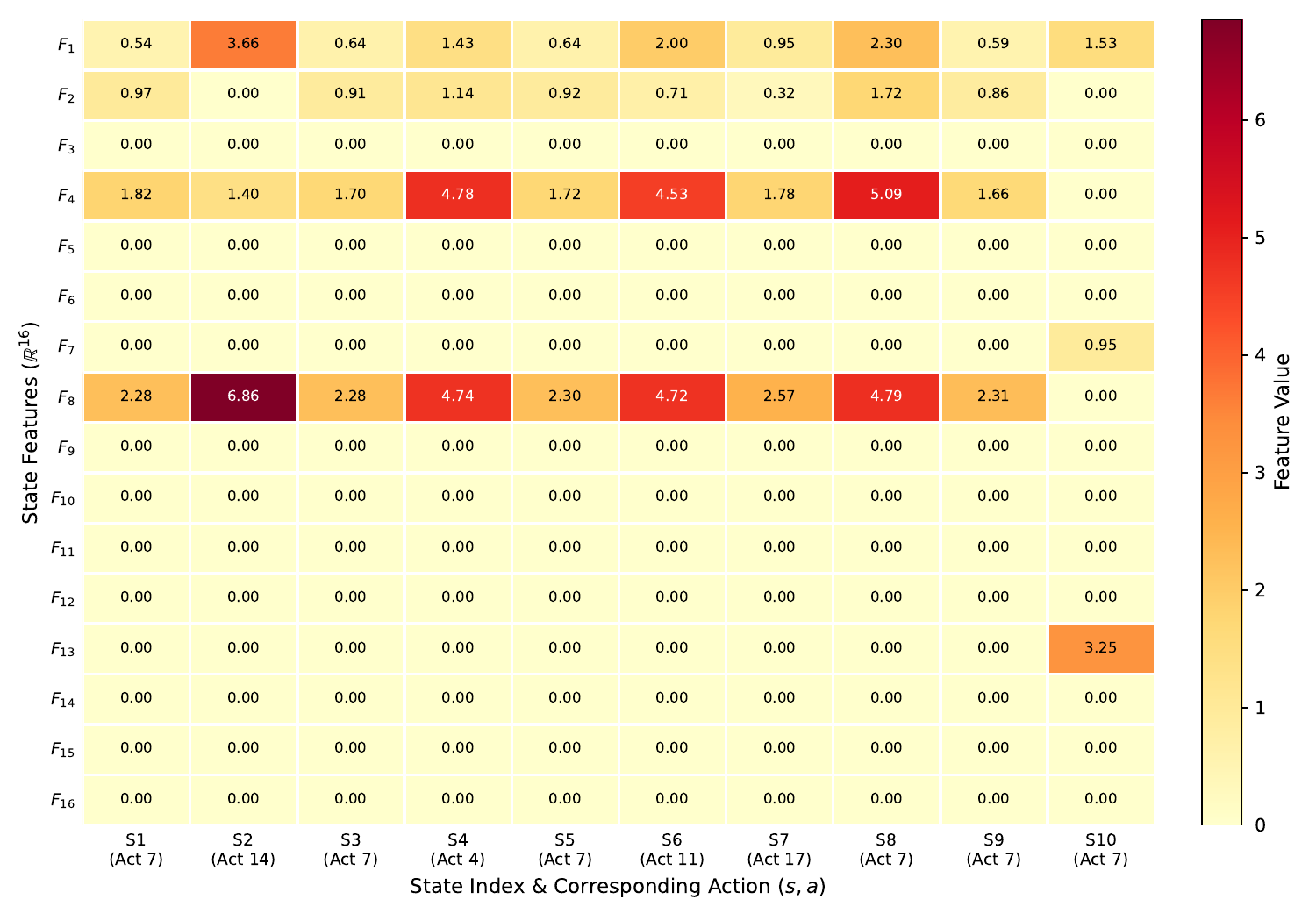}\label{fig:RoadRunner_DQN_feature_map}}
    \caption{RoadRunner feature maps: the $16$ CNN features for ten states sampled along each RL policy's rollout, annotated with the policy's action.}
    \label{fig:RoadRunner_feature_maps}
\end{figure*}

\FloatBarrier

\begin{figure*}[t]
    \section{Baseline Experiment}
    \label{Appendix: Baseline Experiment}
    We train A2C, DQN, and PPO agents in the CartPole environment and collect 200 episodes per agent for SHAP analysis. Linear regression interpretable policies are then constructed using {SILVER} and \textsf{ReShap}. We evaluate each framework under 100 episodes. Figure~\ref{fig: SILVER vs ReShap} compares the performance of the two frameworks.

    \centering
    \includegraphics[width=0.44\linewidth]{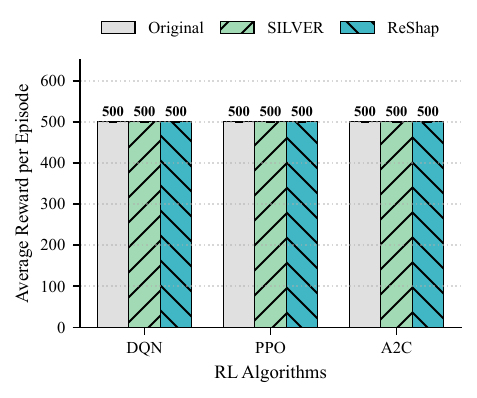}
    \caption{Performance comparison: {SILVER} vs \textsf{ReShap}}
    \label{fig: SILVER vs ReShap}
\end{figure*}

\begin{table*}[t]
\centering
\renewcommand{\arraystretch}{1.0}
\begin{tabular}{|c|l|}
\hline
\textbf{Algorithm} & \textbf{Decision Boundary} \\ \hline
{DQN} & $
\begin{aligned}
f(x, \dot{x}, \theta, \dot{\theta}) &= -0.5x - 0.687\dot{x} - 1.09\theta - \dot{\theta} - 0.018
\end{aligned}$ \\ \hline
{PPO} & $
\begin{aligned}
f(x, \dot{x}, \theta, \dot{\theta}) &= -0.193x - 0.523\dot{x} - \theta - \dot{\theta} + 0.0014
\end{aligned}$ \\ \hline
{A2C} & $
\begin{aligned}
f(x, \dot{x}, \theta, \dot{\theta}) &= -0.4875x - 0.9811\dot{x} - 1.09\theta - \dot{\theta}
\end{aligned}$ \\ \hline
\end{tabular}
\caption{CartPole: interpretable policy boundary for {SILVER}~\protect\cite{li2025explainability}}
\label{tab:SILVER Linear Regression}
\vspace{5mm}
\renewcommand{\arraystretch}{1.0}
\begin{tabular}{|c|l|}
\hline
\textbf{Algorithm} & \textbf{Decision Boundary} \\ \hline
{DQN} & $
\begin{aligned}
f(x, \dot{x}, \theta, \dot{\theta}) &= 0.594 + 1.48x + 1.26\dot{x} + 3.97\theta + 2.62\dot{\theta}
\end{aligned}$ \\ \hline
{PPO} & $
\begin{aligned}
f(x, \dot{x}, \theta, \dot{\theta}) &= 0.508 + 0.892x + 5.57\dot{x} + 38.7\theta + 5.94\dot{\theta}
\end{aligned}$ \\ \hline
{A2C} & $
\begin{aligned}
f(x, \dot{x}, \theta, \dot{\theta}) &= -0.498 + 0.841x + 3.58\dot{x} + 6.71\theta + 4.48\dot{\theta}
\end{aligned}$ \\ \hline
\end{tabular}
\caption{CartPole: interpretable policy boundary for \textsf{ReShap}}
\label{tab:ReShap Linear Regression}
\par\smallskip
\begin{minipage}{\textwidth}
The results indicate that \textsf{ReShap} exhibits performance comparable to {SILVER} on both actor--critic algorithms (PPO, A2C) and value-based algorithms (DQN).
\end{minipage}
\end{table*}
\FloatBarrier

\section{Experiment on RoadRunner}
\label{Appendix: Experiment on RoadRunner}

The RoadRunner environment is a fast-paced Atari game where the agent controls the roadrunner to evade the pursuing coyote across desert terrain. The observation space consists of high-dimensional RGB frames depicting the roadrunner, the coyote, bird seed, and obstacles. The action space includes 18 discrete controls for movement, jumping, ducking, and firing. The agent gains rewards for collecting bird seed and avoiding capture, and penalties for collisions or being caught. The objective is to maximize survival while efficiently collecting rewards.

\begin{figure*}[t]
    \centering
    \subfloat[RoadRunner DQN - SHAP value]{%
        \includegraphics[width=0.32\linewidth]{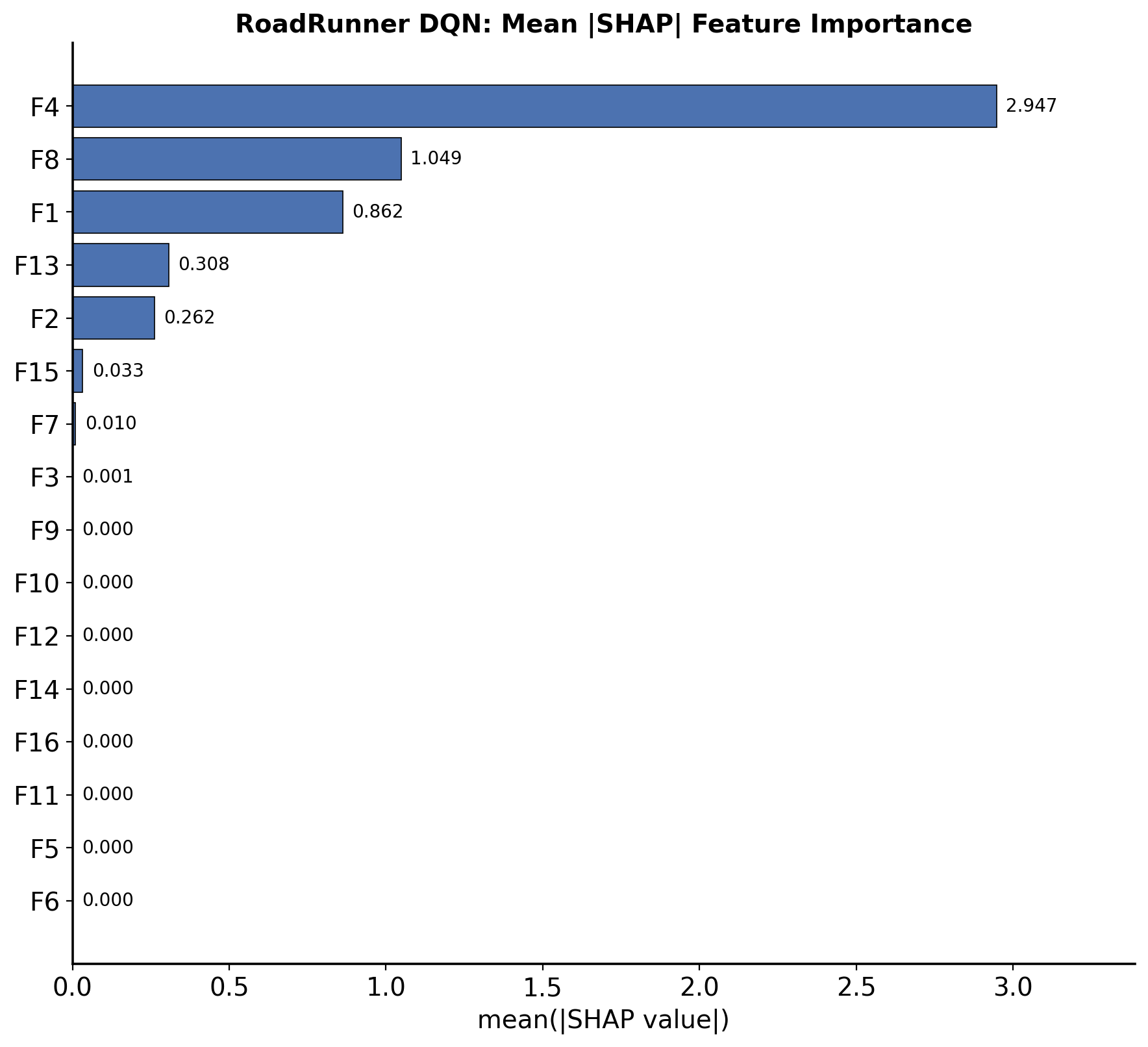}%
        \label{fig:RoadRunner_DQN_SHAP}%
    }
    \hfill
    \subfloat[RoadRunner A2C - SHAP value]{%
        \includegraphics[width=0.32\linewidth]{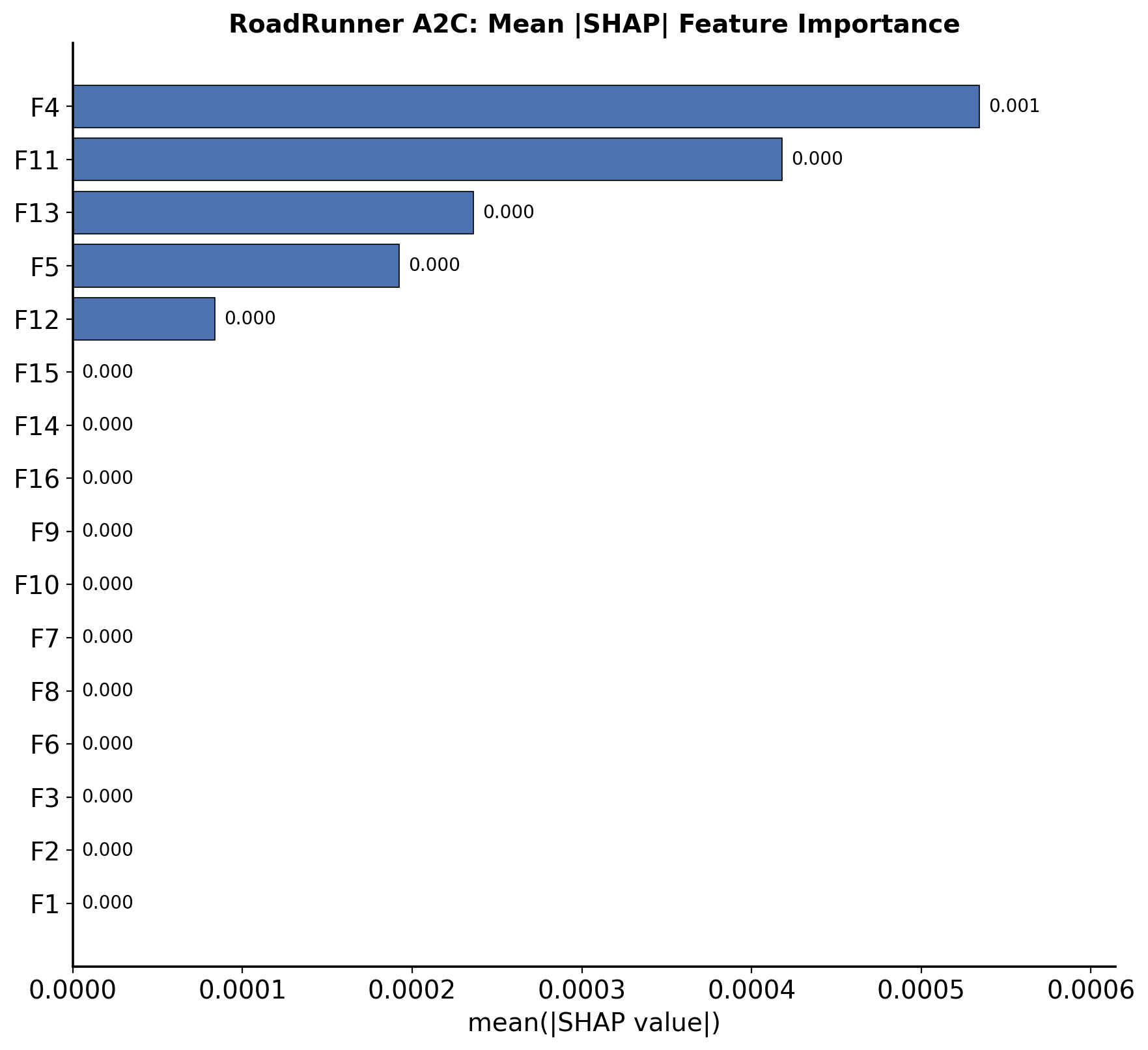}%
        \label{fig:RoadRunner_A2C_SHAP}%
    }
    \hfill
    \subfloat[RoadRunner PPO - SHAP value]{%
        \includegraphics[width=0.32\linewidth]{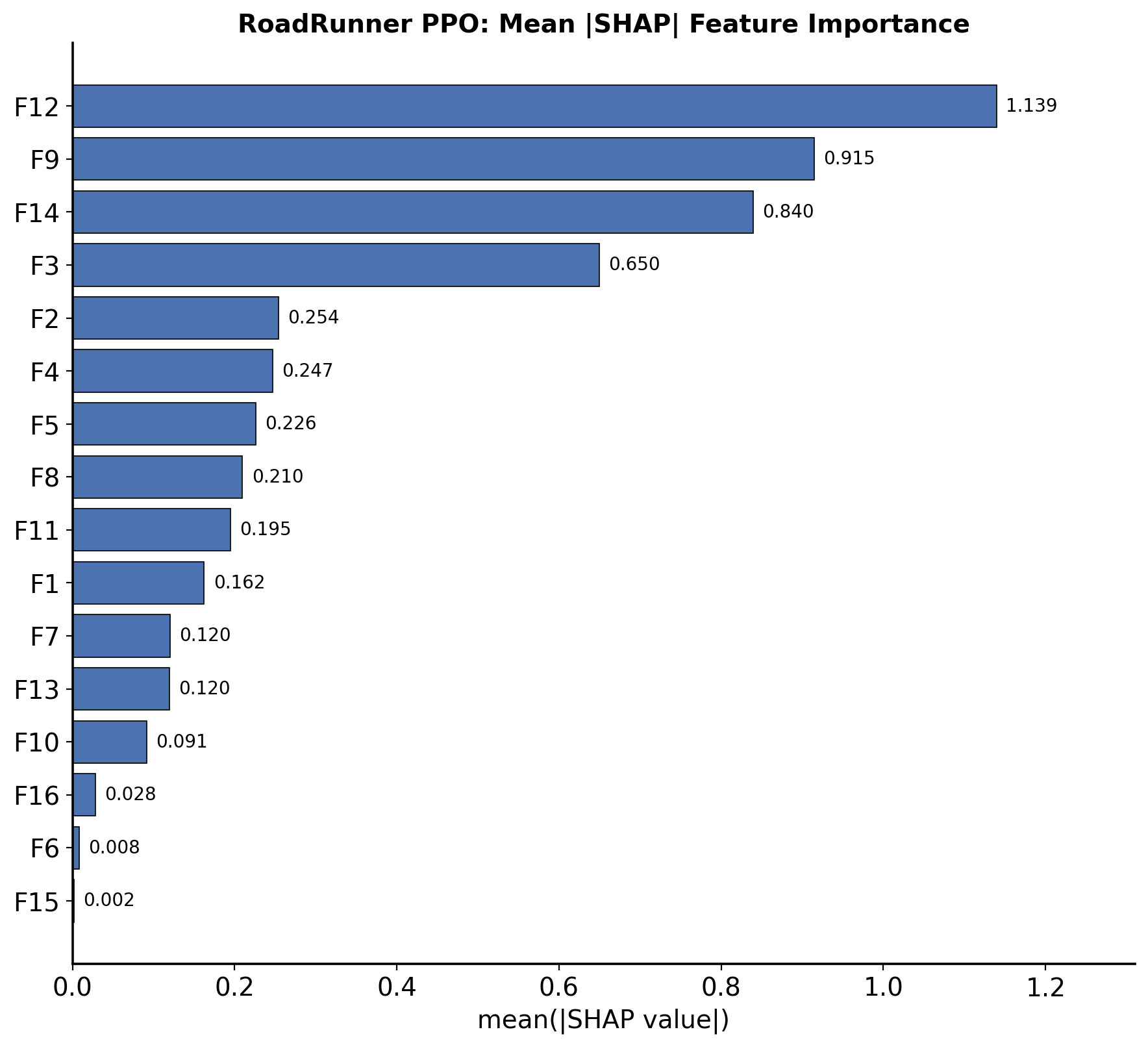}%
        \label{fig:RoadRunner_PPO_SHAP}%
    }
    \caption{Mean absolute SHAP value of each feature for RoadRunner under different algorithms (DQN, A2C, PPO). The A2C rollout is single-action, so its SHAP attributions are negligible (note the axis scale).}
    \label{fig:RoadRunner_SHAP_comparison}
\end{figure*}

The contributions of state features to RL policy’s decisions are visualized in Figure~\ref{fig:RoadRunner_SHAP_comparison}. We then roll out each RL policy, record all visited states together with the policy's chosen actions, and distill the interpretable surrogates on the full set of these $(\text{state}, \pi(\text{state}))$ pairs.

\begin{figure*}[p]
    \centering
    \includegraphics[width=1.0\textwidth]{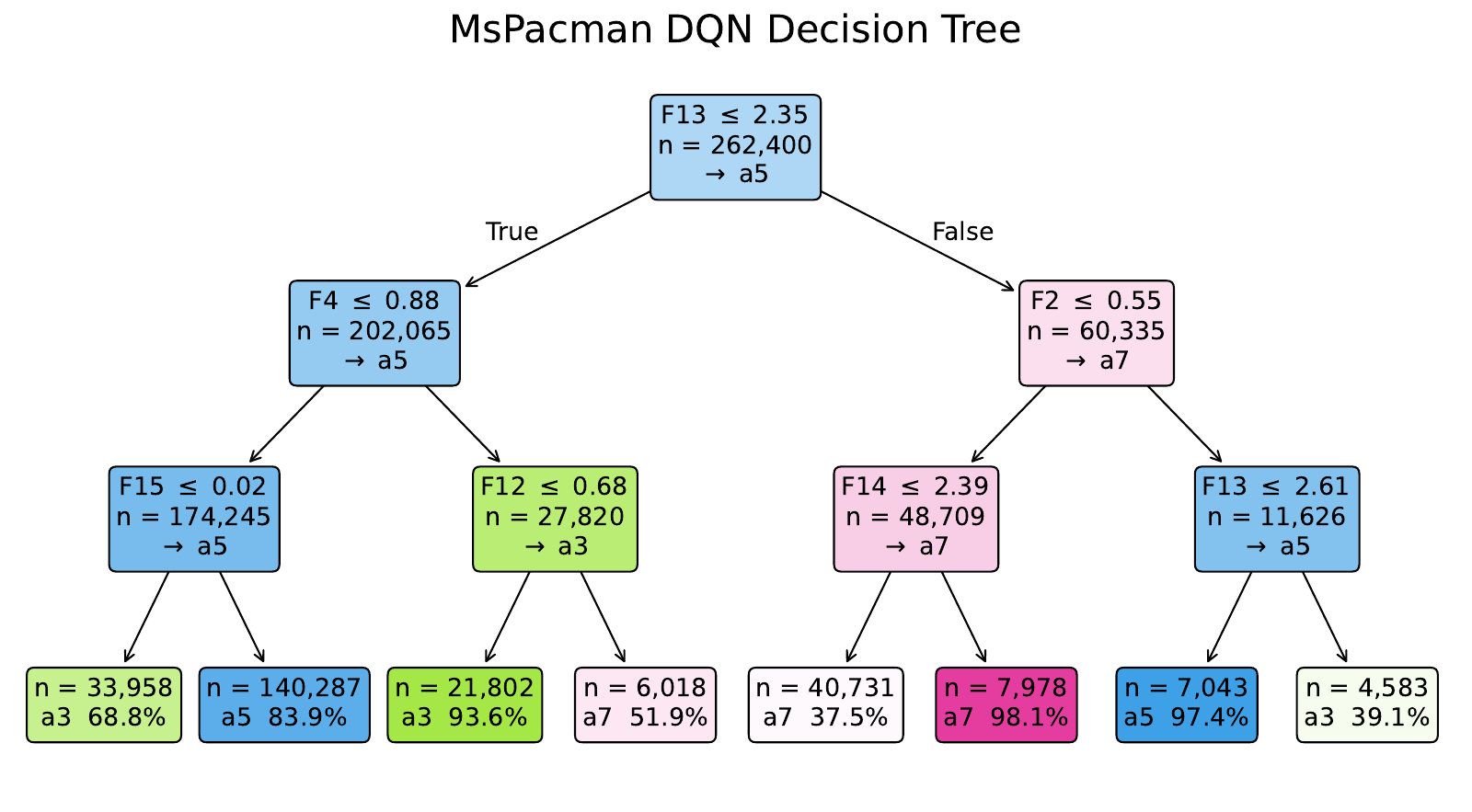}
    \caption{MsPacman DQN decision tree}
    \label{fig:MsPacman_dqn_decision_tree}

    \vspace{8mm}
    \includegraphics[width=1.0\textwidth]{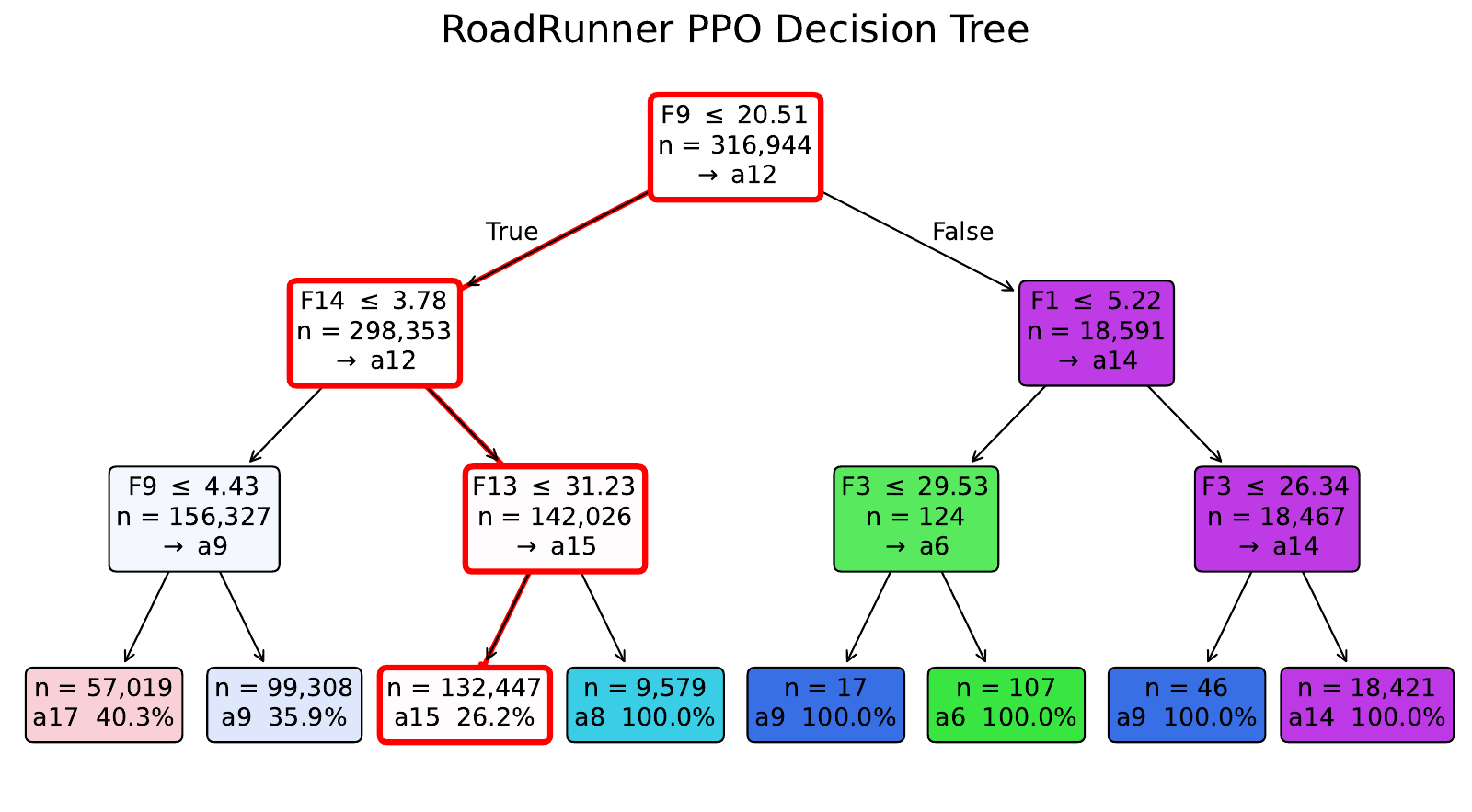}
    \caption{RoadRunner PPO decision tree}
    \label{fig:RoadRunner_ppo_decision_tree}
\end{figure*}

\begin{figure*}[p]
    \centering
    \includegraphics[width=1.0\textwidth]{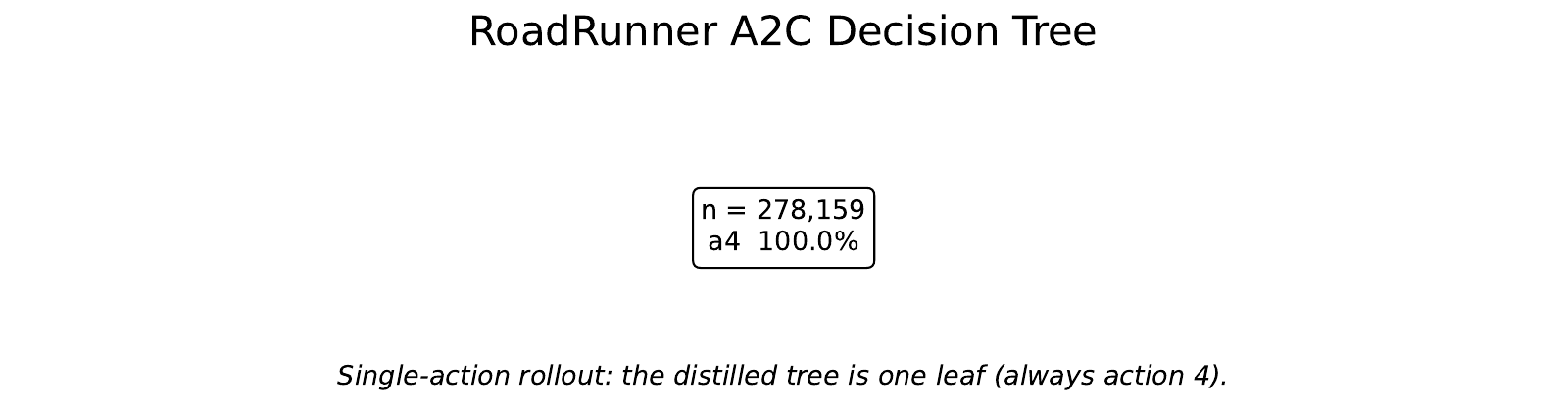}
    \caption{RoadRunner A2C decision tree}
    \label{fig:RoadRunner_a2c_decision_tree}

    \vspace{8mm}
    \includegraphics[width=1.0\textwidth]{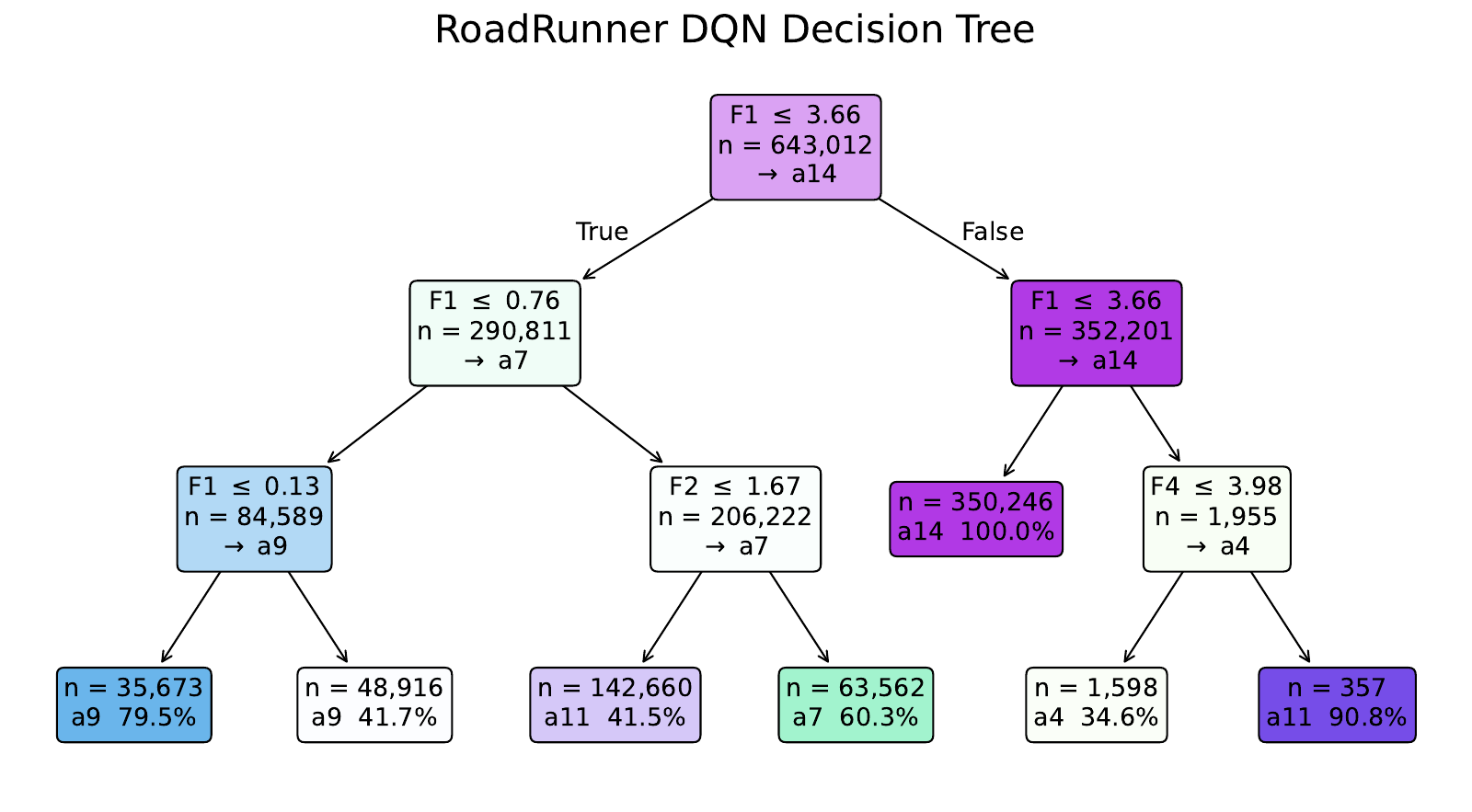}
    \caption{RoadRunner DQN decision tree}
    \label{fig:RoadRunner_dqn_decision_tree}
\end{figure*}

\paragraph{Decision Tree.}
Figure~\ref{fig:RoadRunner_ppo_decision_tree}, \ref{fig:RoadRunner_a2c_decision_tree}, and \ref{fig:RoadRunner_dqn_decision_tree} present decision trees for PPO, A2C, and DQN, respectively.

Taking PPO as an example, consider an example state $S_1$ in Figure~\ref{fig:RoadRunner_PPO_feature_map}. Starting from the root node of Figure~\ref{fig:RoadRunner_ppo_decision_tree}, the decision process unfolds as follows, with the traced path highlighted in red:

\begin{enumerate}[leftmargin=*, topsep=0pt, itemsep=1pt, parsep=0pt]
  \item At the root node, check whether $F_{9} \leq 20.51$.\\
        Since $7.08 \leq 20.51$, the process moves to the \textbf{left child}.
  \item Next, check whether $F_{14} \leq 3.78$.\\
        Since $4.07 > 3.78$, the process moves to the \textbf{right child}.
  \item Then, check whether $F_{13} \leq 31.23$.\\
        Since $0.00 \leq 31.23$, the process moves to the \textbf{left child}, reaching a leaf:
        $\boxed{\text{class} = a_{15}}$, which corresponds to  $\boxed{\text{action} = 15}$.
\end{enumerate}

\paragraph{Linear Regression.}
Table~\ref{tab:RoadRunner Linear Regression} illustrates the interpretable policy boundaries derived from linear regression for PPO, A2C, and DQN. Here, $x_{1}$ through $x_{16}$ represent the 16 state features used in the analysis.

Taking PPO as an example, consider an example state $S_2$ in Figure~\ref{fig:RoadRunner_PPO_feature_map}. Substituting its feature values into the PPO decision boundary in Table~\ref{tab:RoadRunner Linear Regression}, we obtain:
\[
\begin{aligned}
f(x) &= 10.88 - (5.08\times 10^{-1})\,7.28 + (6.38\times 10^{-1})\,2.80 \\
&\quad + (3.46\times 10^{-1})\,4.34 + (1.39\times 10^{-1})\,0.74 \\
&\quad - (6.66\times 10^{-2})\,7.91 - (2.59\times 10^{-1})\,1.94 \\
&\quad - (4.71\times 10^{-1})\,11.71 - (4.14\times 10^{-1})\,1.82 \\
&\quad + (2.61\times 10^{-1})\,5.19 + (4.84\times 10^{-1})\,5.49 \\
&\quad + (1.98\times 10^{-1})\,6.10 - (5.87\times 10^{-1})\,0.72 \\
&\quad + (4.15\times 10^{-1})\,1.09 \approx 8.52.
\end{aligned}
\]
Rounding to the nearest integer and clipping to the valid action range yields $\boxed{f(x) = 9}$, which corresponds to $\boxed{\text{action} = 9}$.

\paragraph{Logistic Regression.}
Table~\ref{tab:RoadRunne_PPO_LogReg} illustrates the interpretable policy boundaries derived from logistic regression for PPO. Taking an example state $S_3$ in Figure~\ref{fig:RoadRunner_PPO_feature_map} and substituting its features into the per-action logit functions yields, for the dominant actions,
\[
\begin{aligned}
f_{4}(x) &\approx 9.11, \quad f_{9}(x) \approx 8.38, \\
f_{12}(x) &\approx 9.85, \quad f_{17}(x) \approx 8.47,
\end{aligned}
\]
with the remaining logits smaller. Applying the softmax $P(a{=}i\mid x)=e^{f_i(x)}/\sum_{j} e^{f_j(x)}$ assigns the highest probability to action $12$ ($P\approx0.51$, versus $0.24$ for action $4$), so the model predicts $\boxed{\text{action} = 12}$. The procedure is identical to that demonstrated for the MsPacman environment.

\FloatBarrier

\clearpage
\section{Experiment on LunarLander} \label{Appendix: Experiment on LunarLander}
Beyond the Atari domains, we further evaluate \textsf{ReShap} on LunarLander, a Box2D continuous-control environment. Unlike the CNN-extracted Atari features, LunarLander exposes an $8$-dimensional \emph{physically meaningful} state vector: the lander's horizontal and vertical position $(x, y)$, its linear velocities $(v_x, v_y)$, its orientation $\theta$ (\textit{angle}) and angular velocity $\omega$ (\textit{angular\_vel}), and two Boolean leg-contact indicators $(c_L, c_R)$ for the left and right landing legs. The action space contains four discrete engine commands: $0$ = do nothing, $1$ = fire left engine, $2$ = fire main engine, and $3$ = fire right engine. The goal is to land the module softly between the two flags on the pad.

\begin{figure*}[t]
    \centering
    \subfloat[LunarLander DQN -- SHAP value\label{fig:LunarLander_DQN_SHAP}]{
        \includegraphics[width=0.32\textwidth]{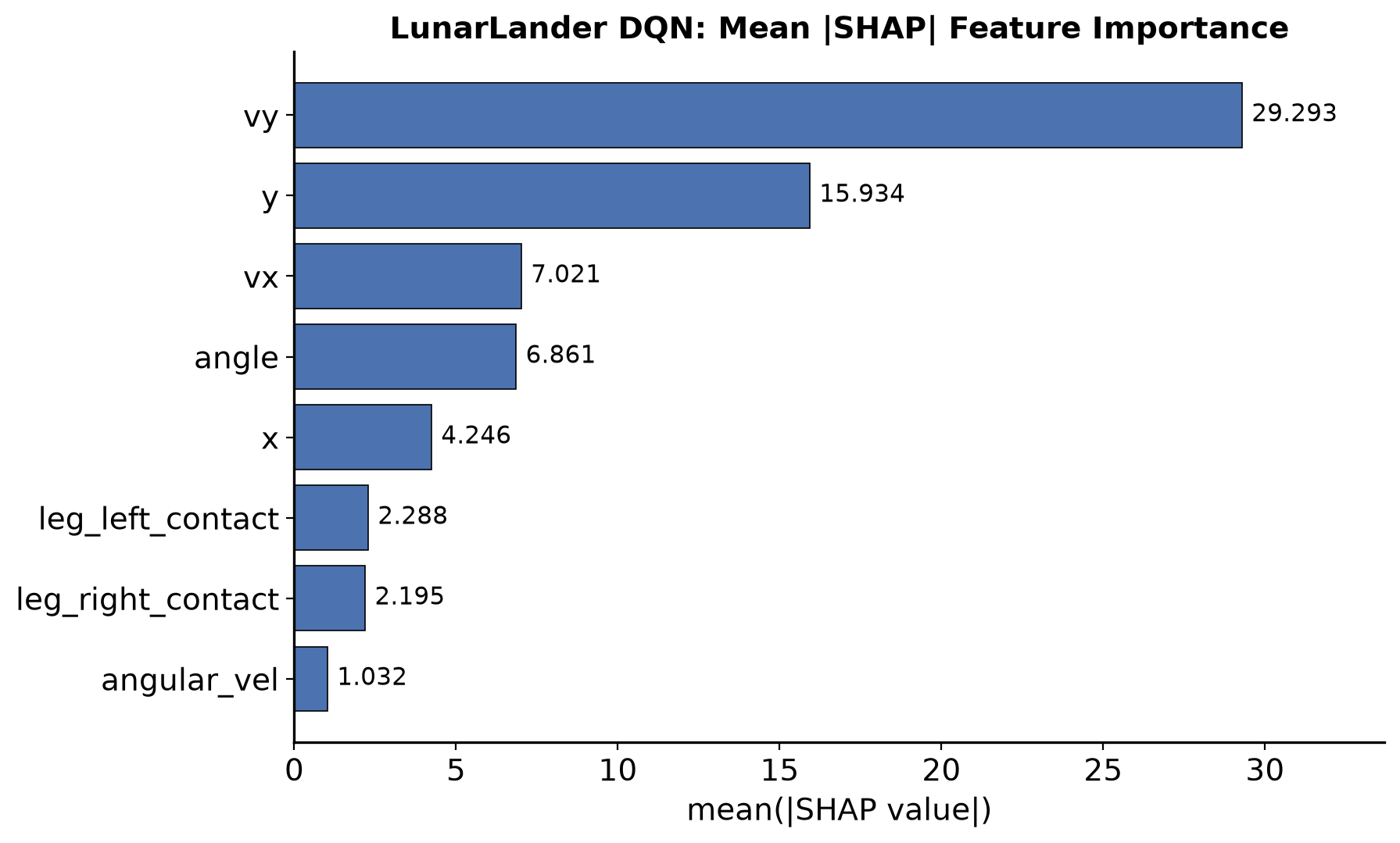}
    }
    \hfill
    \subfloat[LunarLander A2C -- SHAP value\label{fig:LunarLander_A2C_SHAP}]{
        \includegraphics[width=0.32\textwidth]{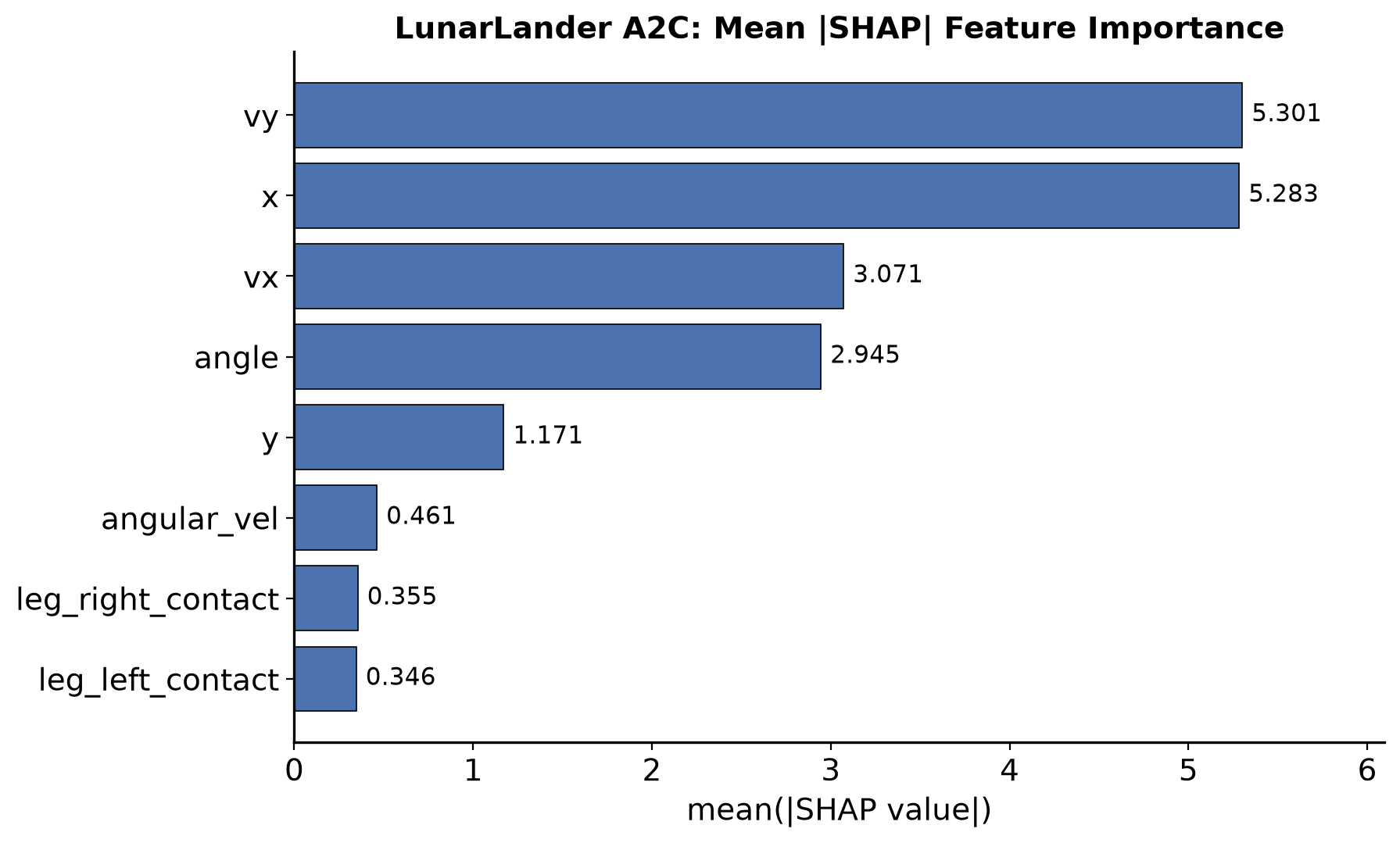}
    }
    \hfill
    \subfloat[LunarLander PPO -- SHAP value\label{fig:LunarLander_PPO_SHAP}]{
        \includegraphics[width=0.32\textwidth]{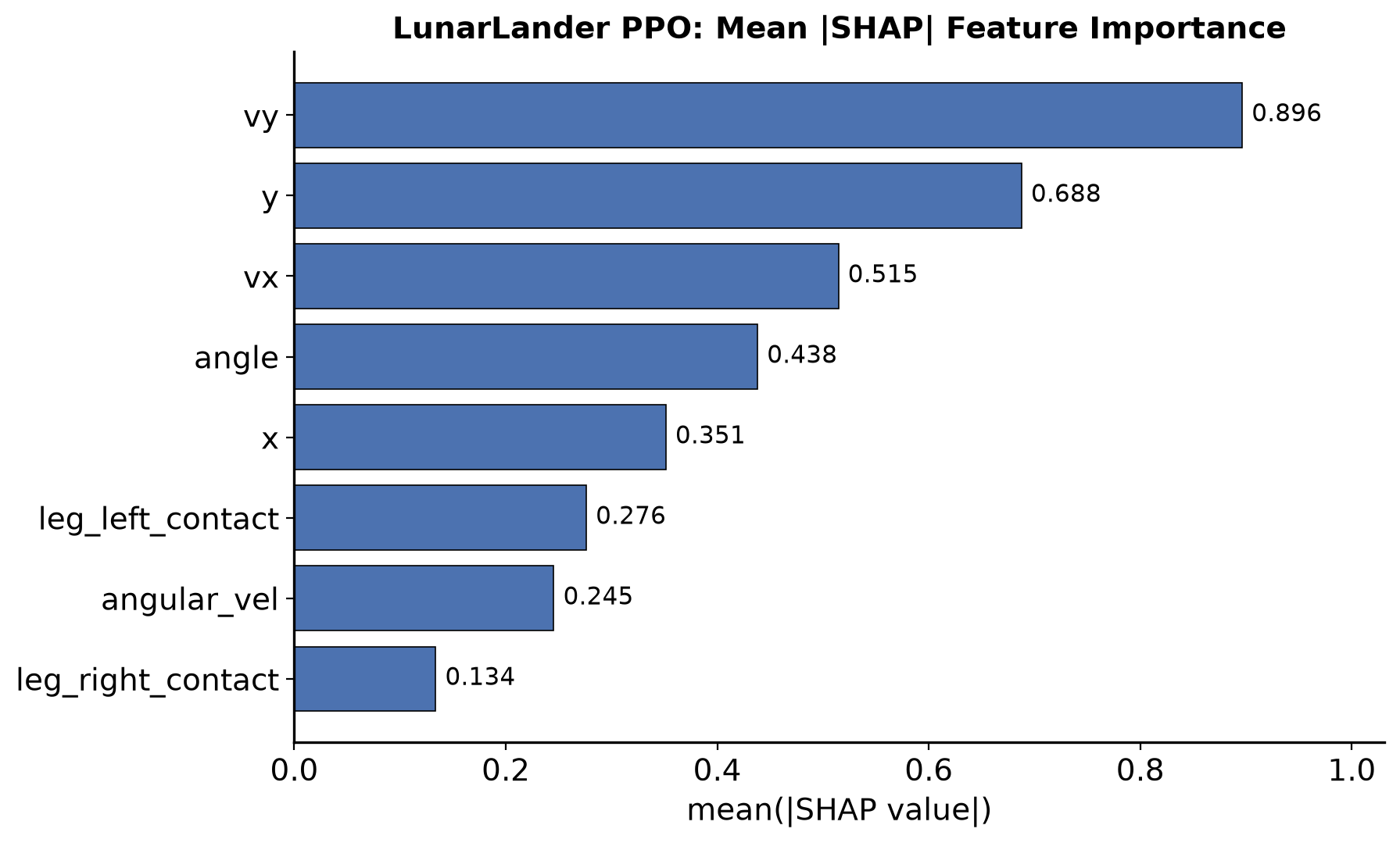}
    }
    \caption{SHAP value comparison of LunarLander under different algorithms (DQN, A2C, PPO)}
    \label{fig:LunarLander_SHAP_comparison}
\end{figure*}

We sample states from episodes collected by each deep RL algorithm and compute the Shapley value of every physical feature, yielding an $8$-dimensional Shapley vector per state. Figure~\ref{fig:LunarLander_SHAP_comparison} presents these attributions for DQN, A2C, and PPO. A key advantage of this domain is that the state features carry explicit physical meaning, so the resulting surrogate formulas are directly human-readable, in contrast to the opaque $16$-dimensional CNN features of the Atari environments. Across all three algorithms, the vertical velocity $v_y$ is the single most influential feature, consistent with the intuition that regulating descent speed dominates the landing decision; the next most important features are the altitude $y$ (PPO, DQN) and the horizontal position $x$ (A2C).

We then roll out each RL policy, record the visited physical states together with the policy's chosen actions, and distill the interpretable surrogates on the full set of these $(\text{state}, \pi(\text{state}))$ pairs. Because the state features are physical, the resulting surrogates are directly human-readable. In the worked examples below we walk through a few representative example states $S_1$, $S_2$, and $S_3$ to illustrate how each surrogate maps features to actions.

\begin{figure*}[p]
    \centering
    \includegraphics[width=1.0\textwidth]{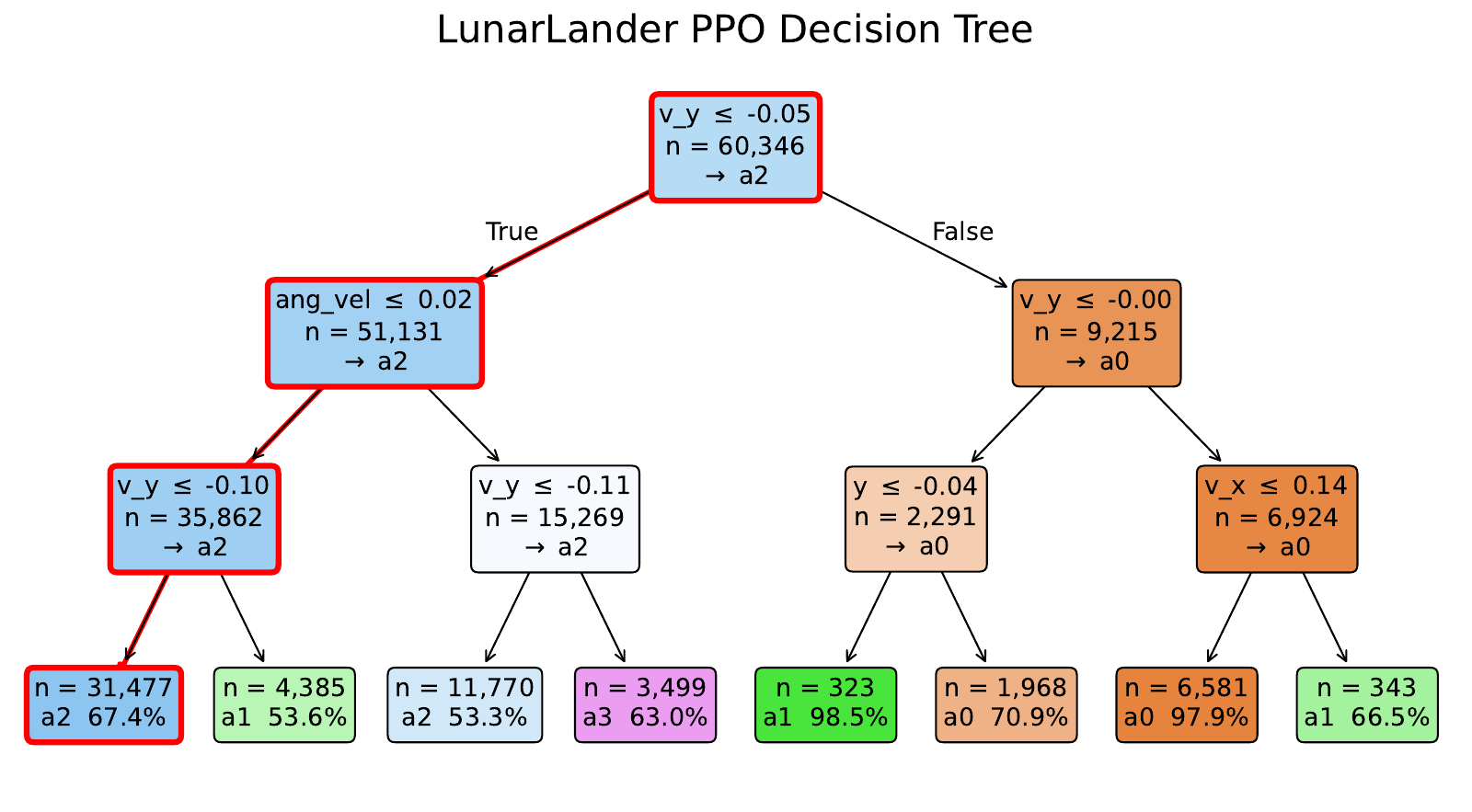}
    \caption{LunarLander PPO decision tree}
    \label{fig:LunarLander_ppo_decision_tree}

    \vspace{8mm}
    \includegraphics[width=1.0\textwidth]{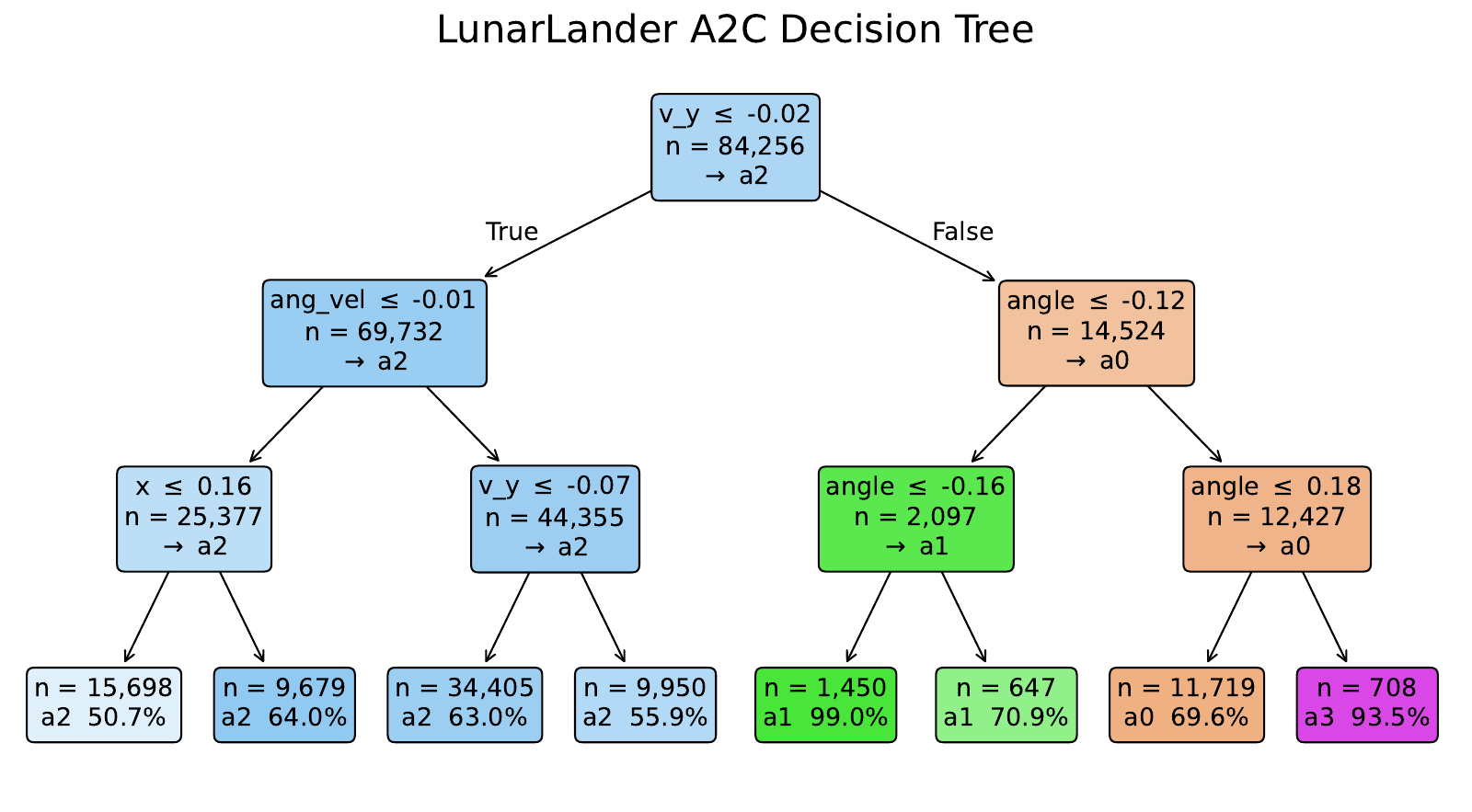}
    \caption{LunarLander A2C decision tree}
    \label{fig:LunarLander_a2c_decision_tree}
\end{figure*}

\begin{figure*}[p]
    \centering
    \includegraphics[width=1.0\textwidth]{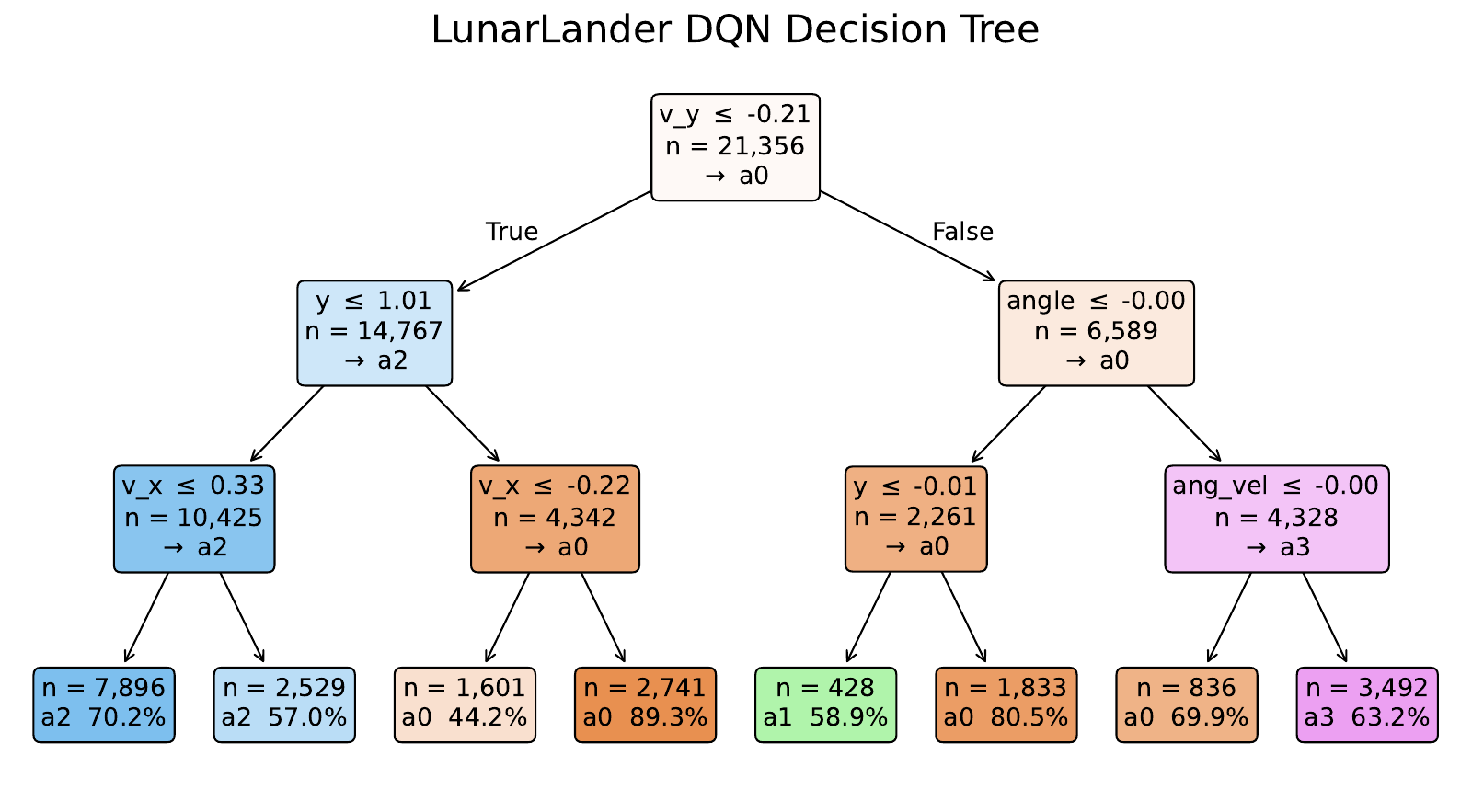}
    \caption{LunarLander DQN decision tree}
    \label{fig:LunarLander_dqn_decision_tree}
\end{figure*}

\paragraph{Decision Tree.}
Figures~\ref{fig:LunarLander_ppo_decision_tree}, \ref{fig:LunarLander_a2c_decision_tree}, and \ref{fig:LunarLander_dqn_decision_tree} present the decision trees for PPO, A2C, and DQN, respectively. Because the split variables are physical quantities such as $v_y$ and $\omega$, each internal node reads as an intuitive threshold on the lander's motion. For legibility the trees are shown pruned to depth three; the surrogate evaluated for the reported results is the full tree.

Taking PPO as an example, consider an example state $S_1$ with $x=0.017$, $y=1.374$, $v_x=0.557$, $v_y=-0.580$, $\theta=-0.014$, $\omega=-0.050$, and $c_L=c_R=0$. Starting from the root node of Figure~\ref{fig:LunarLander_ppo_decision_tree}, the decision process unfolds as follows, with the traced path highlighted in red:

\begin{enumerate}[leftmargin=*, topsep=0pt, itemsep=1pt, parsep=0pt]
  \item At the root node, check whether $v_y \leq -0.05$.\\
        Since $-0.580 < -0.05$, the process moves to the \textbf{left child}.
  \item Next, check whether $\omega \leq 0.02$.\\
        Since $-0.050 < 0.02$, the process moves to the \textbf{left child}.
  \item Then, check whether $v_y \leq -0.1$.\\
        Since $-0.580 < -0.1$, the process moves to the \textbf{left child}, reaching a leaf:
        $\boxed{\text{class} = 2}$, which corresponds to $\boxed{\text{action} = 2}$ (fire main engine).
\end{enumerate}

The descent speed $v_y$ dominates the path, reflecting the intuition that regulating vertical velocity governs the landing decision.

\paragraph{Linear Regression.}
Table~\ref{tab:LunarLander Linear Regression} reports the interpretable policy boundaries derived from linear regression for PPO, A2C, and DQN; here the regressors are the named physical features rather than anonymous indices. Taking PPO as an example, consider an example state $S_2$ with $x=-0.155$, $y=0.455$, $v_x=0.021$, $v_y=-0.170$, $\theta=-0.032$, $\omega=0.013$, and $c_L=c_R=0$. 

Substituting into the PPO boundary yields:

\[
\begin{aligned}
f(x) &= 1.81 - 0.0252 \times (-0.155) - 0.575 \times 0.455 \\
&\quad - 0.676 \times 0.021 - 2.69 \times (-0.170) \\
&\quad + 1.12 \times (-0.032) + 4.57 \times 0.013 \\
&\approx 2.02.
\end{aligned}
\]

Rounding to the nearest integer yields $\boxed{f(x)=2}$, i.e., $\boxed{\text{action}=2}$. The dominant positive term comes from $v_y$ ($+0.46$), confirming that the descent speed drives the choice to fire the main engine. We caution, however, that regressing discrete action \emph{indices} with a single continuous output is ill-posed: although it recovers individual example states, its episode-level return collapses (see Discussion).

\paragraph{Logistic Regression.}
Table~\ref{tab:LunarLander_PPO_LogReg}, Table~\ref{tab:LunarLander_A2C_LogReg}, and Table~\ref{tab:LunarLander_DQN_LogReg} report the per-action logit functions $f_0,\dots,f_3$ for PPO, A2C, and DQN. Taking PPO as an example, consider an example state $S_3$ with $x=0.160$, $y=0.630$, $v_x=-0.184$, $v_y=-0.205$, $\theta=-0.005$, $\omega=-0.042$, and $c_L=c_R=0$. Substituting into Table~\ref{tab:LunarLander_PPO_LogReg} gives the logits
\[
\begin{aligned}
f_0(x)&\approx-1.93, \quad f_1(x)\approx0.82, \\
f_2(x)&\approx0.50, \quad f_3(x)\approx0.62.
\end{aligned}
\]

Applying the softmax $P(a=i\mid x)=e^{f_i(x)}/\sum_j e^{f_j(x)}$ yields probabilities $(0.03,\,0.38,\,0.28,\,0.31)$, so the predicted action is $\hat{a}=\arg\max_i P(a=i\mid x)=\boxed{1}$ (fire left engine).

\clearpage
\begin{figure*}[t]
    \section{Full Fidelity Results}
    \label{Appendix: Fidelity}
    This appendix reports the action-agreement fidelity of the three interpretable surrogates for every environment and RL algorithm, complementing the return comparison of Figure~\ref{fig:Performance_Comparison} in the main text.

    \centering
    \subfloat[LunarLander\label{fig:newfid_LunarLander}]{
        \includegraphics[width=0.32\textwidth]{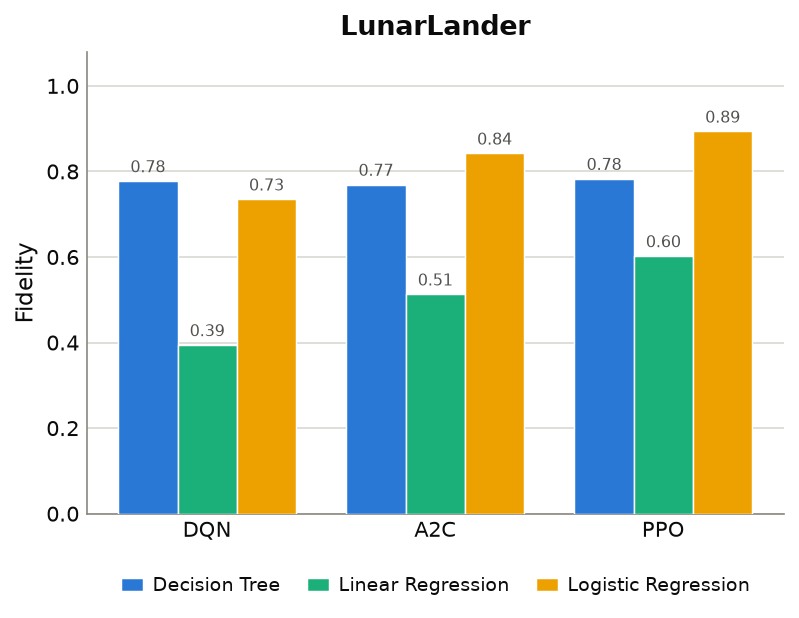}
    }
    \hfill
    \subfloat[MsPacman\label{fig:newfid_MsPacman}]{
        \includegraphics[width=0.32\textwidth]{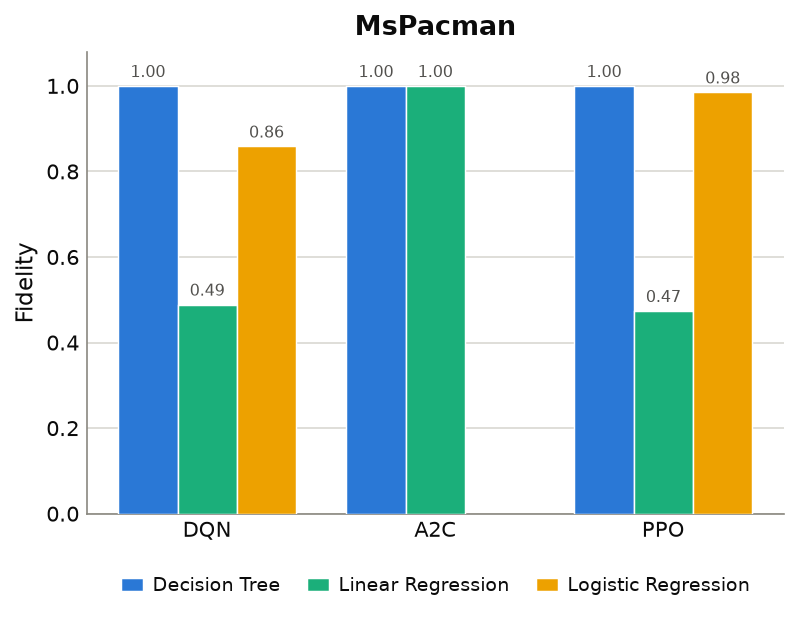}
    }
    \hfill
    \subfloat[RoadRunner\label{fig:newfid_RoadRunner}]{
        \includegraphics[width=0.32\textwidth]{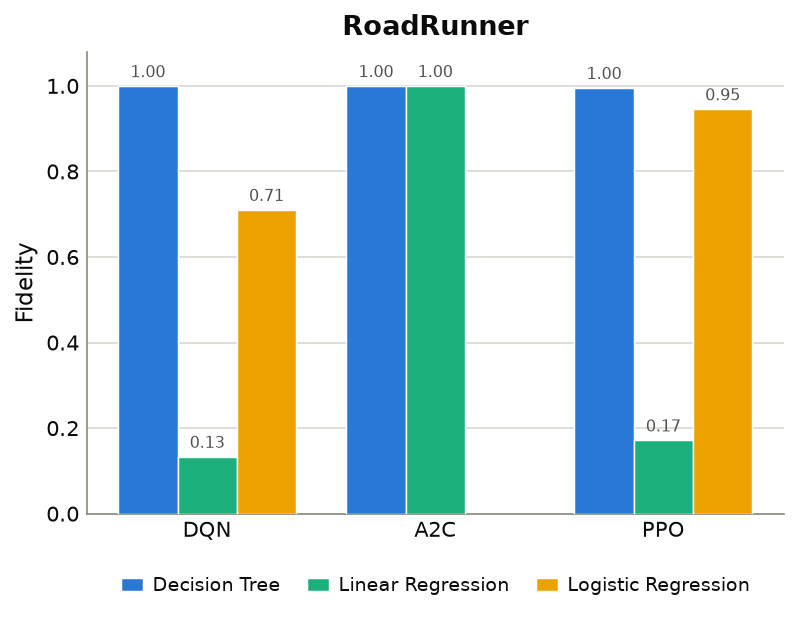}
    }
    \caption{Fidelity (action agreement with the RL policy) of the three interpretable surrogates (decision tree, linear regression, logistic regression) under full-distribution distillation, for each RL algorithm (DQN, A2C, PPO) and environment. Bars are grouped by algorithm; within each group one bar per surrogate. The A2C logistic-regression bar is omitted in MsPacman and RoadRunner because the collected rollouts contain fewer than two distinct actions.}
    \label{fig:new_fidelity_matrix}
\end{figure*}
\FloatBarrier

\clearpage
\begin{figure*}[t]
    \section{Results of Human Study}
    \label{Appendix: Results of Human Study}

    Based on the decision tree shown in Figure~\ref{fig:MsPacman_ppo_decision_tree}, we designed four questions. The corresponding feature values for each question are listed below:
    \begin{itemize}[leftmargin=*, topsep=0pt, itemsep=1pt, parsep=0pt]
        \item \textbf{Question 1}: Feature 2 = 6909.615, Feature 4 = 2007.798, Feature 12 = 15111.683.
        \item \textbf{Question 2}: Feature 2 = 6187.522, Feature 4 = 765.015, Feature 12 = 6892.513.
        \item \textbf{Question 3}: Feature 2 = 4254.228, Feature 4 = 1093.875, Feature 12 = 24553.357.
        \item \textbf{Question 4}: Feature 2 = 8804.404, Feature 4 = 0.000, Feature 12 = 21856.025.
    \end{itemize}

    \centering
    \subfloat[Education level\label{Education level}]{
        \includegraphics[width=0.44\textwidth]{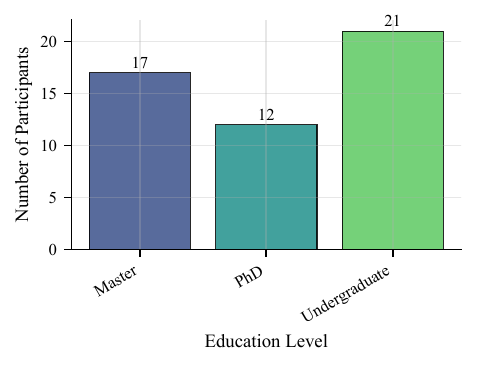}
    }
    \subfloat[Experience in RL\label{RL background}]{
        \includegraphics[width=0.44\textwidth]{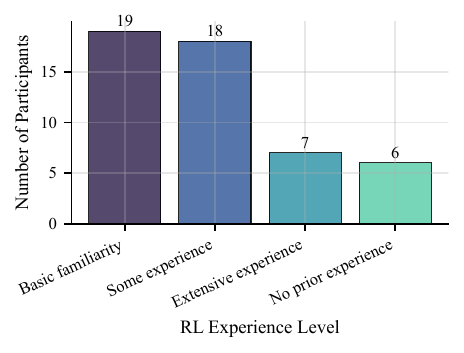}
    }
    \caption{Participants' education level and experience in RL}
    \label{Education level and RL background}
\end{figure*}

\begin{figure*}[t]
\centering
    \subfloat[Distribution of Q1 responses\label{Q1 responses}]{
        \includegraphics[width=0.34\textwidth]{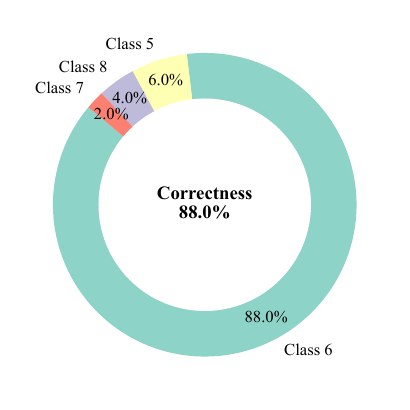}
    }
    \subfloat[Distribution of Q2 responses\label{Q2 responses}]{
        \includegraphics[width=0.34\textwidth]{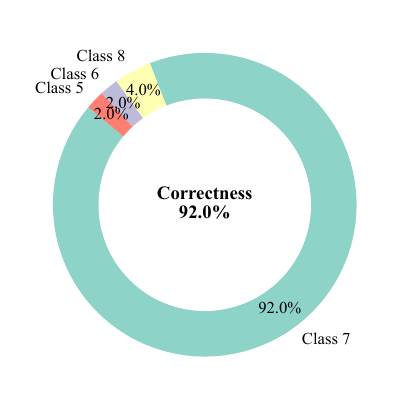}
    }

    \vspace{1em} %
    
    \subfloat[Distribution of Q3 responses\label{Q3 responses}]{
        \includegraphics[width=0.34\textwidth]{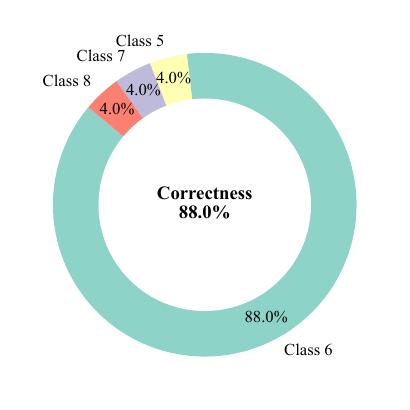}
    }
    \subfloat[Distribution of Q4 responses\label{Q4 responses}]{
        \includegraphics[width=0.34\textwidth]{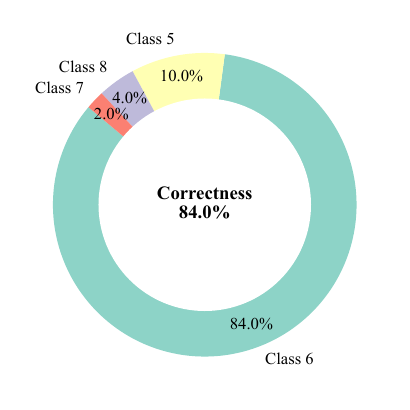}
    }
    \caption{Distribution of responses from Q1 to Q4}
    \label{Distribution of responses from Q1 to Q4}
\end{figure*}

\begin{figure*}[t]
    \centering
    \includegraphics[width=0.46\textwidth]{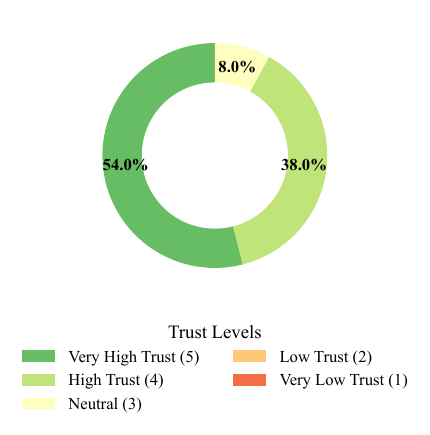}
    \caption{Participants' trust ratings}
    \label{trust_score_distribution}
    \vspace{4mm}
\end{figure*}

\begin{figure*}[t]
\centering
    \subfloat[Distribution of Q1 time cost\label{Q1 time cost}]{
        \includegraphics[width=0.34\textwidth]{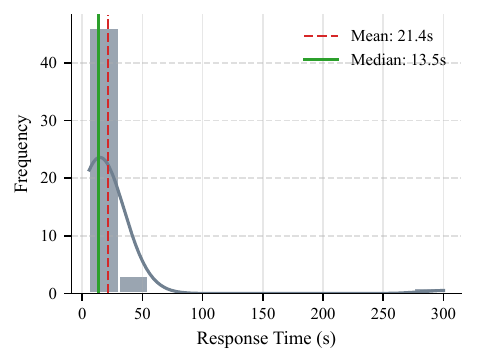}
    }
    \subfloat[Distribution of Q2 time cost\label{Q2 time cost}]{
        \includegraphics[width=0.34\textwidth]{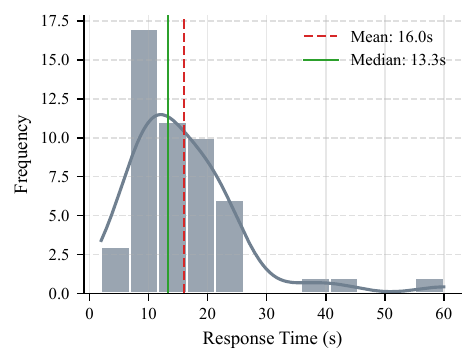}
    }

    \vspace{1em} %
    
    \subfloat[Distribution of Q3 time cost\label{Q3 time cost}]{
        \includegraphics[width=0.34\textwidth]{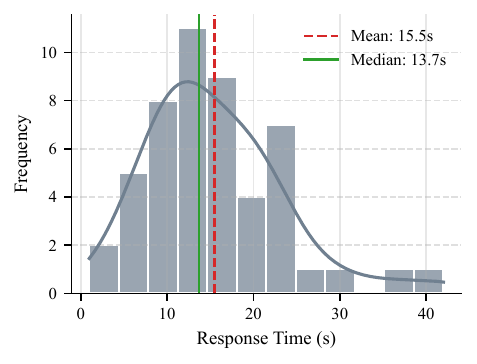}
    }
    \subfloat[Distribution of Q4 time cost\label{Q4 time cost}]{
        \includegraphics[width=0.34\textwidth]{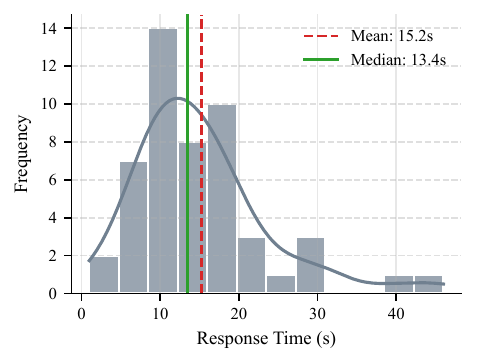}
    }
    \caption{Distribution of time cost from Q1 to Q4}
    \label{Distribution of time cost from Q1 to Q4}
\end{figure*}

\FloatBarrier
\begin{figure*}[t]
\section{Details of Limitations} \label{Appendix: Details of Limitations}

\paragraph{Loss of Semantic Meaning in Feature Extraction.}
The extracted state features lack explicit semantic meanings. Unlike classic control environments (e.g., CartPole, MountainCar) with well-defined variables, our state vectors lack such semantic clarity. We plan to integrate Vision Language Models (VLMs)~\cite{chen2024vision} to derive semantically grounded state features from complex, high-dimensional observations. 

\paragraph{Limited to Virtual Environments.} 
Although the framework has been evaluated in complex virtual domains, these environments do not fully represent real-world decision-making scenarios. With the growing interest in explainable RL fine-tuning for LLMs, extending the framework to transformer-based deep RL algorithms (e.g., GRPO, DAPO, GSPO) represents a promising future direction. In such contexts, user prompts could be modeled as observations, while the LLMs' output tokens serve as actions.

\paragraph{Restricted Scope of Human-subject Study.}
The human-subject experiment is conducted on a single environment with one interpretable policy. Future work could involve broader evaluations across multiple environments and diverse interpretable policy forms to better assess generalizability and human comprehension.
\end{figure*}
\FloatBarrier

\begin{figure*}[p]
    \section{Decision Trees} \label{Appendix: Decision Trees}

    The trees are grouped two to a page for legibility. In every tree an internal node reports its split condition, the number of training samples routed to it ($n$), and the majority class ($\rightarrow a_i$); a leaf reports its sample count, its majority class, and the percentage of those samples belonging to that class (its purity). Classes are written $a_i$, where $i$ is the discrete action index. Node colour encodes the majority class and its saturation the node purity, following the default \textsf{scikit-learn} palette. The split features are the $16$ CNN-extracted features $F_1,\dots,F_{16}$ for the Atari environments and the eight physical state variables for LunarLander, so the LunarLander splits read directly as thresholds on position, velocity, angle, and leg contact. For the MsPacman PPO, RoadRunner PPO, and LunarLander PPO trees, the decision path of the corresponding worked-example state is highlighted in red. For readability the trees are pruned to depth three; the surrogate evaluated for the reported fidelity and return is the full, unpruned tree. When a deterministic rollout is single-action (as for A2C in both Atari environments) the distilled tree collapses to a single leaf, shown with an explanatory note.

    \centering
    \par\vspace{3mm}
    \includegraphics[width=0.96\textwidth]{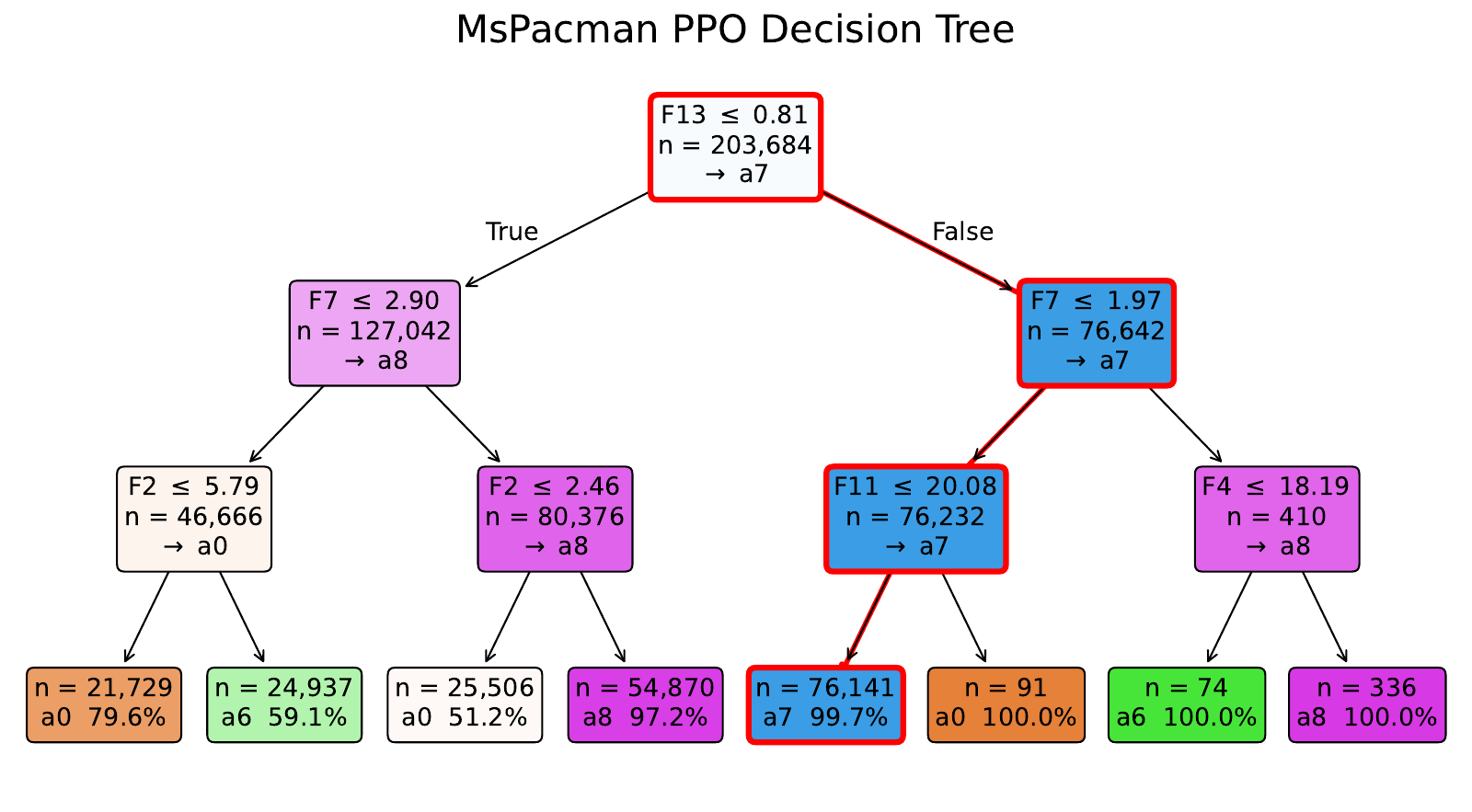}
    \caption{MsPacman PPO decision tree}
    \label{fig:MsPacman_ppo_decision_tree}

    \vspace{8mm}
    \includegraphics[width=0.96\textwidth]{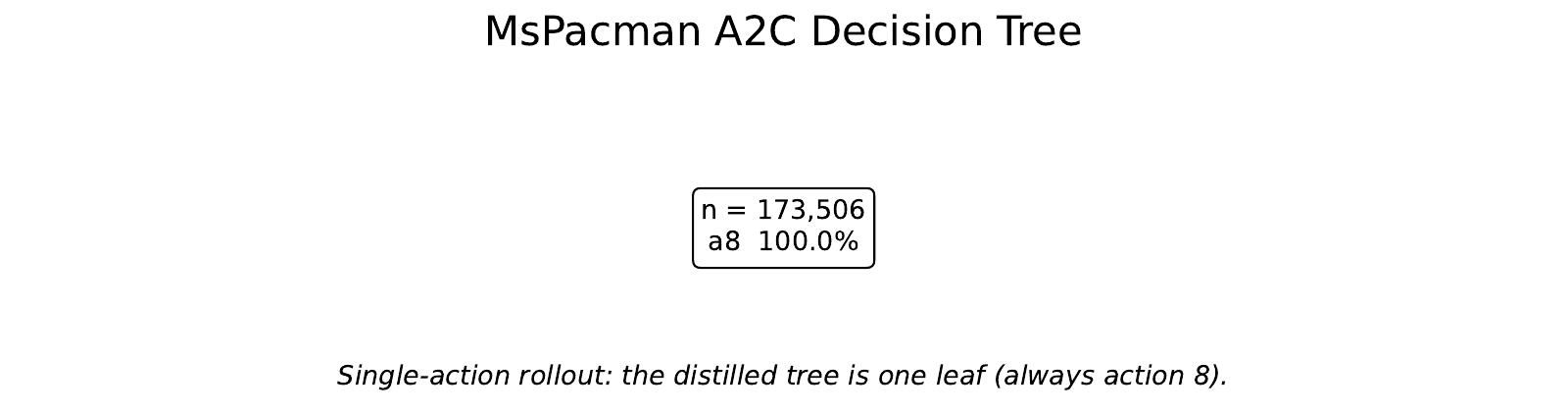}
    \caption{MsPacman A2C decision tree}
    \label{fig:MsPacman_a2c_decision_tree}
\end{figure*}

\clearpage
\begin{figure*}[t]
\begin{minipage}[t]{0.48\textwidth}
\vspace{0pt}
\section{Decision Tree' Prompt} \label{Appendix: Decision Tree' Prompt}

\begin{promptbox}{XRL Decision Tree Prompt Template}
\footnotesize
\textbf{Role:} You are an Explainable AI (XAI) Assistant. Your goal is to provide a clear, logical explanation for a Reinforcement Learning agent's decision by tracing its path through a Decision Tree.

\vspace{2mm}
\textbf{1. Input Parameters:}
\begin{itemize}
    \item \textbf{Domain:} [Enter Domain, e.g., Ms. Pac-Man]
    \item \textbf{Current State (Features):} [Enter State Values, e.g., F13 = 6.81, F7 = 0.00, F11 = 11.71]
    \item \textbf{Decision Tree Structure:} [Describe logic or refer to Figure~\ref{fig:MsPacman_ppo_decision_tree}]
\end{itemize}

\textbf{2. Task Instructions:}
\begin{itemize}
    \item \textbf{Trace the Path:} Start at the root node. For every decision node, compare the \textbf{Current State} values against the threshold. Explicitly state if the condition is True or False.
    \item \textbf{Identify the Leaf:} Determine the final \textbf{Class} (Action) reached. Include the confidence metrics (e.g., sample size and value distribution).
    \item \textbf{Contextualize:} Translate the numerical logic into a domain-specific reason.
\end{itemize}

\textbf{3. Output Format:}
\begin{itemize}
    \item \textbf{[Logic Path]:} A step-by-step trace of the comparison gates.
    \item \textbf{[Decision]:} The final action/class selected by the agent.
    \item \textbf{[Human-Readable Explanation]:} A 2--3 sentence narrative summarizing why the agent made this choice.
\end{itemize}
\end{promptbox}
\end{minipage}
\hfill
\begin{minipage}[t]{0.48\textwidth}
\vspace{0pt}

\section{Natural Language Explanation Synthesis for Decision Tree Policy in MsPacMan PPO} \label{Appendix: Natural Language Explanation Synthesis for Decision Tree}

\begin{itemize}
    \item \textbf{[Logic Path]:}
    \begin{enumerate}
        \item \textbf{Root Node:} $F_{13} (6.81) \leq 0.81$ is \textbf{False}, so the trace moves to the right child.
        \item \textbf{Internal Node:} $F_{7} (0.00) \leq 1.97$ is \textbf{True}.
        \item \textbf{Leaf Decision:} $F_{11} (11.71) \leq 20.08$ is \textbf{True}.
    \end{enumerate}

    \item \textbf{[Decision]:}
    The agent selected \textbf{Class $a_7$}. This leaf node represents $76{,}141$ samples with a value distribution of $[0, 71, 75899, 171]$ over classes $(a_0, a_6, a_7, a_8)$, indicating $\approx 99.7\%$ classification confidence.

    \item \textbf{[Human-Readable Explanation]:}
    The agent executed Action 7 because feature $F_{13}$ exceeded its root threshold while $F_{7}$ and $F_{11}$ fell below theirs. Given the high purity of the final leaf node, the agent identifies this specific state configuration as a high-certainty scenario for this action within the MsPacman domain.
\end{itemize}
\end{minipage}
\end{figure*}

\clearpage
\begingroup
\setlength{\dblfloatsep}{8mm}
\begin{table*}[t]
\section{Linear Regression}
\label{Appendix: Linear Regression}
\centering
\renewcommand{\arraystretch}{1.0}
\begin{tabular}{|c|l|}
\hline
\textbf{Algorithm} & \textbf{Decision Boundary} \\ \hline
PPO & $
\begin{aligned}
f(x) &= 5.68 + 7.64\times 10^{-1}\,x_{1} + 5.57\times 10^{-1}\,x_{2} - 1.70\times 10^{-1}\,x_{4} - 4.19\times 10^{-1}\,x_{6} \\
     &\quad + 1.05\times 10^{-1}\,x_{7} + 1.90\times 10^{-1}\,x_{9} + 5.91\times 10^{-2}\,x_{10} + 6.51\times 10^{-2}\,x_{11} \\
     &\quad - 8.92\times 10^{-1}\,x_{12} + 2.72\times 10^{-1}\,x_{13} - 6.18\times 10^{-1}\,x_{15} + 2.66\times 10^{-1}\,x_{16}
\end{aligned}$ \\ \hline
A2C & $f(x) = 8.00$ \quad {\footnotesize (constant; the A2C rollout is single-action)} \\ \hline
DQN & $
\begin{aligned}
f(x) &= 4.79 - 4.36\times 10^{-1}\,x_{1} - 3.29\times 10^{-1}\,x_{2} - 1.43\times 10^{-1}\,x_{3} - 8.48\times 10^{-1}\,x_{4} \\
     &\quad + 1.31\times 10^{0}\,x_{6} - 6.67\times 10^{-1}\,x_{7} - 6.73\times 10^{-1}\,x_{8} + 1.47\times 10^{0}\,x_{9} \\
     &\quad - 3.73\times 10^{-1}\,x_{11} + 1.88\times 10^{0}\,x_{12} - 4.43\times 10^{-1}\,x_{13} + 2.19\times 10^{0}\,x_{14} \\
     &\quad + 2.85\times 10^{-1}\,x_{15} - 1.05\times 10^{0}\,x_{16}
\end{aligned}$ \\ \hline
\end{tabular}
\caption{MsPacman: interpretable policy boundary for linear regression}
\label{tab:MsPacman Linear Regression}
\end{table*}

\begin{table*}[t]
\centering
\renewcommand{\arraystretch}{1.0}
\begin{tabular}{|c|l|}
\hline
\textbf{Algorithm} & \textbf{Decision Boundary} \\ \hline
PPO & $
\begin{aligned}
f(x) &= 10.88 - 5.08\times 10^{-1}\,x_{1} + 6.38\times 10^{-1}\,x_{2} + 3.46\times 10^{-1}\,x_{3} + 1.39\times 10^{-1}\,x_{4} \\
     &\quad + 6.90\times 10^{-3}\,x_{5} + 4.73\times 10^{-1}\,x_{6} - 6.66\times 10^{-2}\,x_{7} - 2.59\times 10^{-1}\,x_{8} \\
     &\quad - 4.71\times 10^{-1}\,x_{9} - 4.14\times 10^{-1}\,x_{10} + 2.61\times 10^{-1}\,x_{11} + 4.84\times 10^{-1}\,x_{12} \\
     &\quad + 1.98\times 10^{-1}\,x_{13} - 5.87\times 10^{-1}\,x_{14} - 6.19\times 10^{-2}\,x_{15} + 4.15\times 10^{-1}\,x_{16}
\end{aligned}$ \\ \hline
A2C & $f(x) = 4.00$ \quad {\footnotesize (constant; the A2C rollout is single-action)} \\ \hline
DQN & $
\begin{aligned}
f(x) &= 6.81 + 4.05\times 10^{-1}\,x_{1} - 2.50\times 10^{-1}\,x_{2} + 1.26\times 10^{1}\,x_{3} - 7.79\times 10^{-1}\,x_{4} \\
     &\quad - 7.55\times 10^{-1}\,x_{7} + 9.87\times 10^{-1}\,x_{8} - 1.12\times 10^{1}\,x_{9} + 3.53\times 10^{-1}\,x_{13} \\
     &\quad - 1.20\times 10^{-1}\,x_{15}
\end{aligned}$ \\ \hline
\end{tabular}
\caption{RoadRunner: interpretable policy boundary for linear regression}
\label{tab:RoadRunner Linear Regression}
\end{table*}

\begin{table*}[t]
\centering
\renewcommand{\arraystretch}{1.2}
\begin{tabular}{|c|l|}
\hline
\textbf{Algorithm} & \textbf{Decision Boundary} \\ \hline
PPO & $
\begin{aligned}
f(x) &= 1.81 - 0.0252\,x - 0.575\,y - 0.676\,v_x - 2.69\,v_y \\
&\quad + 1.12\,\theta + 4.57\,\omega - 0.536\,c_L - 1.12\,c_R
\end{aligned}$ \\ \hline
A2C & $
\begin{aligned}
f(x) &= 1.28 + 0.0504\,x + 0.0434\,y - 0.912\,v_x - 3.29\,v_y \\
&\quad + 1.57\,\theta + 1.27\,\omega + 0.163\,c_L - 0.671\,c_R
\end{aligned}$ \\ \hline
DQN & $
\begin{aligned}
f(x) &= 2.18 - 0.0538\,x - 1.19\,y - 0.938\,v_x - 0.0293\,v_y \\
&\quad - 0.0808\,\theta - 0.178\,\omega - 0.957\,c_L - 0.247\,c_R
\end{aligned}$ \\ \hline
\end{tabular}
\caption{LunarLander: interpretable policy boundary for linear regression. Regressors are the physical state variables: position $(x,y)$, velocity $(v_x,v_y)$, angle $\theta$, angular velocity $\omega$, and leg-contact flags $(c_L,c_R)$.}
\label{tab:LunarLander Linear Regression}
\end{table*}

\clearpage
\endgroup
\begin{figure*}[t]
\begin{minipage}[t]{0.48\textwidth}
\vspace{0pt}
\section{Linear Regression' Prompt} \label{Appendix: Linear Regression' Prompt}

\begin{promptbox}{XRL Linear Regression Prompt Template}
\footnotesize
\textbf{Role:} You are an Explainable AI (XAI) Assistant. Your goal is to explain a Reinforcement Learning agent's discrete action by tracing the calculation of a Linear Regression decision boundary.

\vspace{2mm}
\textbf{1. Input Parameters:}
\begin{itemize}
    \item \textbf{Domain:} [e.g., Ms. Pac-Man]
    \item \textbf{Linear Equation:} [e.g., $f(x) = 5.68 + 7.64{\times}10^{-1}\,x_1 + \cdots$]
    \item \textbf{Current State Features:} [e.g., $x_6=8.85, x_9=4.40, x_{11}=8.25, x_{13}=12.54$]
\end{itemize}

\textbf{2. Task Instructions:}
\begin{itemize}
    \item \textbf{Computation:} Show the substitution of feature values into the equation to find the continuous value $f(x)$.
    \item \textbf{Discretization:} Round $f(x)$ to the nearest integer and clip it to the valid action range to identify the final discrete \textbf{action index}.
    \item \textbf{Impact Analysis:} Identify which features (weighted terms) most significantly pushed the value toward the chosen action or prevented it from falling into a different action index.
\end{itemize}

\textbf{3. Output Format:}
\begin{itemize}
    \item \textbf{[Mathematical Trace]:} Step-by-step calculation leading to the continuous result (e.g., $\approx 6.76$).
    \item \textbf{[Action Recovery]:} Show the rounding step: $\mathrm{round}(f(x)) = \text{Action Index}$.
    \item \textbf{[Human-Readable Explanation]:} A brief summary (2--3 sentences) explaining how the current features determined the action, highlighting the most influential state variables.
\end{itemize}
\end{promptbox}
\end{minipage}
\hfill
\begin{minipage}[t]{0.48\textwidth}
\vspace{0pt}

\section{Natural Language Explanation Synthesis for Linear Regression Policy in MsPacMan PPO} \label{Appendix: Natural Language Explanation Synthesis for Linear Regression}

\begin{itemize}
    \item \textbf{[Mathematical Trace]:} 
    The continuous decision value $f(x)$ is calculated by substituting the current state features into the regression equation:
    \begin{align*}
        f(x) &= 5.68 - (4.19 \times 10^{-1} \cdot 8.85) \\
             &\quad + (1.90 \times 10^{-1} \cdot 4.40) + (6.51 \times 10^{-2} \cdot 8.25) \\
             &\quad + (2.72 \times 10^{-1} \cdot 12.54) \\
        f(x) &\approx 5.68 - 3.705 + 0.835 + 0.537 + 3.415 \\
        f(x) &\approx 6.76
    \end{align*}

    \item \textbf{[Action Recovery]:}
    The continuous result is discretized by rounding to the nearest integer (and clipping to the valid action range) to determine the specific action index:
    \[ \text{Action Index} = \mathrm{round}(6.76) = \mathbf{7} \]

    \item \textbf{[Human-Readable Explanation]:}
    The agent selected \textbf{Action 7} due to a specific balance between the state variables. Starting from the intercept ($5.68$), the dominant positive contributions came from $F_{13}$ ($+3.42$) and $F_{9}$ ($+0.84$), while $F_{6}$ exerted the strongest negative pressure ($-3.71$); together these push the continuous value to $\approx 6.76$, which rounds to Action 7.
\end{itemize}
\end{minipage}
\end{figure*}

\clearpage
\begin{table*}[t]
\section{Logistic Regression}
\label{Appendix: Logistic Regression}
\centering
\begin{tabularx}{\textwidth}{c X}
\hline
Action $i$ & $f_i(x)$ \\
\hline
0 & $3.34 - 1.75\times 10^{-4}x_{1} - 3.20\times 10^{0}x_{2} - 2.14\times 10^{0}x_{4} + 6.40\times 10^{-1}x_{6} + 4.59\times 10^{-1}x_{7} + 2.47\times 10^{-1}x_{9} - 8.89\times 10^{-2}x_{10} + 5.90\times 10^{-1}x_{11} - 5.92\times 10^{-4}x_{12} - 4.26\times 10^{0}x_{13} - 3.47\times 10^{-3}x_{15} - 1.69\times 10^{-3}x_{16}$ \\

6 & $0.96 - 1.35\times 10^{0}x_{1} + 3.52\times 10^{0}x_{2} + 3.65\times 10^{0}x_{4} - 2.73\times 10^{0}x_{6} + 3.09\times 10^{0}x_{7} + 2.57\times 10^{0}x_{9} - 2.38\times 10^{0}x_{10} + 8.93\times 10^{-1}x_{11} - 3.32\times 10^{0}x_{12} - 5.43\times 10^{0}x_{13} - 6.44\times 10^{-1}x_{15} + 6.89\times 10^{0}x_{16}$ \\

7 & $-4.72 + 3.74\times 10^{0}x_{1} - 4.79\times 10^{0}x_{2} - 3.93\times 10^{0}x_{4} + 5.13\times 10^{0}x_{6} - 8.32\times 10^{0}x_{7} - 2.25\times 10^{0}x_{9} + 4.57\times 10^{0}x_{10} - 2.54\times 10^{0}x_{11} + 3.81\times 10^{0}x_{12} + 1.36\times 10^{1}x_{13} + 2.55\times 10^{0}x_{15} - 1.46\times 10^{1}x_{16}$ \\

8 & $0.42 - 2.38\times 10^{0}x_{1} + 4.47\times 10^{0}x_{2} + 2.42\times 10^{0}x_{4} - 3.04\times 10^{0}x_{6} + 4.77\times 10^{0}x_{7} - 5.63\times 10^{-1}x_{9} - 2.10\times 10^{0}x_{10} + 1.06\times 10^{0}x_{11} - 4.90\times 10^{-1}x_{12} - 3.89\times 10^{0}x_{13} - 1.90\times 10^{0}x_{15} + 7.70\times 10^{0}x_{16}$ \\
\hline
\end{tabularx}
\caption{MsPacman PPO: interpretable policy boundary for logistic regression}
\label{tab:MsPacman_PPO_LogReg}
\end{table*}

\begin{table*}[t]
\centering
\begin{tabularx}{\textwidth}{X}
\hline
The MsPacman A2C policy is single-action in the held-out rollout (it always selects action $8$), so only one class is present. Logistic regression requires at least two distinct classes and is therefore undefined for this configuration. \\
\hline
\end{tabularx}
\caption{MsPacman A2C: logistic regression is undefined; the rollout is single-action (always action $8$).}
\label{tab:MsPacman_A2C_LogReg}
\end{table*}

\begin{table*}[htbp]
\centering
\begin{tabularx}{\textwidth}{c X}
\hline
Action $i$ & $f_i(x)$ \\
\hline
1 & $-6.24 - 2.84\times 10^{-3}x_{1} - 5.12\times 10^{-1}x_{2} - 4.79\times 10^{0}x_{3} + 1.45\times 10^{1}x_{4} - 2.30\times 10^{1}x_{6} + 8.50\times 10^{0}x_{7} - 1.32\times 10^{0}x_{8} - 7.43\times 10^{0}x_{9} - 6.60\times 10^{0}x_{11} - 9.37\times 10^{0}x_{12} + 9.69\times 10^{-1}x_{13} - 1.69\times 10^{-2}x_{14} + 7.85\times 10^{0}x_{15} - 8.84\times 10^{-2}x_{16}$ \\

3 & $-0.37 + 1.71\times 10^{0}x_{1} + 2.67\times 10^{0}x_{2} + 7.97\times 10^{0}x_{3} - 7.02\times 10^{0}x_{4} + 7.05\times 10^{0}x_{6} + 1.07\times 10^{1}x_{7} - 1.53\times 10^{1}x_{8} + 1.34\times 10^{1}x_{9} + 1.33\times 10^{1}x_{11} - 1.76\times 10^{1}x_{12} + 2.22\times 10^{0}x_{13} - 7.86\times 10^{0}x_{14} - 3.79\times 10^{0}x_{15} + 6.95\times 10^{0}x_{16}$ \\

4 & $-5.71 - 1.10\times 10^{0}x_{1} + 1.99\times 10^{0}x_{2} + 7.48\times 10^{0}x_{3} - 1.30\times 10^{0}x_{4} + 9.24\times 10^{0}x_{6} + 2.64\times 10^{0}x_{7} - 1.72\times 10^{0}x_{8} - 4.19\times 10^{0}x_{9} - 6.59\times 10^{0}x_{11} + 2.93\times 10^{0}x_{12} + 2.58\times 10^{0}x_{13} + 5.67\times 10^{0}x_{14} + 4.58\times 10^{-1}x_{15} - 7.37\times 10^{0}x_{16}$ \\

5 & $5.71 - 3.10\times 10^{0}x_{1} - 1.19\times 10^{0}x_{2} - 3.58\times 10^{0}x_{3} - 9.69\times 10^{-1}x_{4} + 1.45\times 10^{0}x_{6} + 2.17\times 10^{0}x_{7} + 1.43\times 10^{0}x_{8} - 9.01\times 10^{0}x_{9} + 9.06\times 10^{0}x_{11} + 2.10\times 10^{1}x_{12} + 1.53\times 10^{0}x_{13} + 3.87\times 10^{0}x_{14} + 1.23\times 10^{-1}x_{15} - 1.55\times 10^{0}x_{16}$ \\

6 & $6.98 - 2.80\times 10^{0}x_{1} - 2.53\times 10^{0}x_{2} + 1.05\times 10^{1}x_{3} - 1.63\times 10^{1}x_{4} - 4.93\times 10^{0}x_{6} - 2.00\times 10^{0}x_{7} - 3.42\times 10^{0}x_{8} + 4.68\times 10^{0}x_{9} - 1.00\times 10^{1}x_{11} - 9.11\times 10^{0}x_{12} - 7.74\times 10^{0}x_{13} - 1.47\times 10^{0}x_{14} - 5.48\times 10^{0}x_{15} - 2.26\times 10^{0}x_{16}$ \\

7 & $-0.37 + 5.29\times 10^{0}x_{1} - 4.24\times 10^{-1}x_{2} - 1.76\times 10^{1}x_{3} + 1.11\times 10^{1}x_{4} + 1.01\times 10^{1}x_{6} - 2.20\times 10^{1}x_{7} + 2.04\times 10^{1}x_{8} + 2.55\times 10^{0}x_{9} + 8.42\times 10^{-1}x_{11} + 1.21\times 10^{1}x_{12} + 4.45\times 10^{-1}x_{13} - 1.85\times 10^{-1}x_{14} + 8.36\times 10^{-1}x_{15} + 4.33\times 10^{0}x_{16}$ \\
\hline
\end{tabularx}
\caption{MsPacman DQN: interpretable policy boundary for logistic regression}
\label{tab:MsPacman_DQN_LogReg}
\end{table*}

\begin{table*}[htbp]
\centering
\begin{tabularx}{\textwidth}{c X}
\hline
Action $i$ & $f_i(x)$ \\
\hline
2 & $1.46 + 3.21\times 10^{-1}x_{1} - 2.23\times 10^{0}x_{2} - 3.64\times 10^{0}x_{3} - 1.39\times 10^{0}x_{4} + 3.23\times 10^{-1}x_{5} + 1.29\times 10^{0}x_{6} + 4.60\times 10^{-1}x_{7} + 7.28\times 10^{-1}x_{8} - 1.82\times 10^{0}x_{9} + 1.41\times 10^{0}x_{10} - 1.37\times 10^{-1}x_{11} - 6.14\times 10^{-1}x_{12} - 3.62\times 10^{-2}x_{13} + 7.33\times 10^{0}x_{14} + 3.36\times 10^{-1}x_{15} + 5.36\times 10^{-1}x_{16}$ \\

4 & $-1.11 + 4.25\times 10^{-1}x_{1} - 6.47\times 10^{-1}x_{2} - 8.72\times 10^{-1}x_{3} + 1.77\times 10^{-1}x_{4} - 6.32\times 10^{-1}x_{5} - 5.72\times 10^{-1}x_{6} - 2.04\times 10^{-1}x_{7} + 1.22\times 10^{0}x_{8} + 3.71\times 10^{0}x_{9} - 9.32\times 10^{-1}x_{10} + 3.12\times 10^{-1}x_{11} - 4.23\times 10^{-1}x_{12} + 2.56\times 10^{-3}x_{13} - 1.19\times 10^{0}x_{14} + 2.47\times 10^{-1}x_{15} - 8.40\times 10^{-1}x_{16}$ \\

5 & $0.34 + 6.35\times 10^{-1}x_{1} + 1.90\times 10^{0}x_{2} + 4.35\times 10^{0}x_{3} + 5.28\times 10^{-2}x_{4} - 6.20\times 10^{-1}x_{5} - 8.39\times 10^{-1}x_{6} + 5.40\times 10^{-1}x_{7} - 2.29\times 10^{-1}x_{8} + 5.34\times 10^{-1}x_{9} + 5.31\times 10^{-1}x_{10} + 1.01\times 10^{0}x_{11} - 1.07\times 10^{0}x_{12} + 7.69\times 10^{-1}x_{13} - 5.88\times 10^{0}x_{14} - 2.04\times 10^{0}x_{15} + 7.79\times 10^{-1}x_{16}$ \\

6 & $-0.07 - 1.06\times 10^{0}x_{1} - 1.09\times 10^{-1}x_{2} + 7.66\times 10^{-1}x_{3} + 2.15\times 10^{0}x_{4} - 3.32\times 10^{-2}x_{5} - 8.83\times 10^{-3}x_{6} - 4.67\times 10^{-1}x_{7} - 2.28\times 10^{0}x_{8} - 2.71\times 10^{0}x_{9} - 8.40\times 10^{-2}x_{10} - 6.64\times 10^{-2}x_{11} - 1.42\times 10^{0}x_{12} - 1.21\times 10^{-1}x_{13} - 1.74\times 10^{0}x_{14} - 1.37\times 10^{-3}x_{15} - 5.16\times 10^{-3}x_{16}$ \\

7 & $-0.74 + 6.55\times 10^{-1}x_{1} - 2.39\times 10^{0}x_{2} - 3.94\times 10^{0}x_{3} - 2.90\times 10^{-1}x_{4} + 4.26\times 10^{-1}x_{5} + 2.18\times 10^{-1}x_{6} + 2.12\times 10^{-2}x_{7} + 7.59\times 10^{-1}x_{8} + 2.92\times 10^{0}x_{9} + 1.41\times 10^{-1}x_{10} + 4.77\times 10^{-1}x_{11} - 2.30\times 10^{0}x_{12} - 2.68\times 10^{-2}x_{13} + 4.88\times 10^{0}x_{14} + 5.42\times 10^{-3}x_{15} + 5.90\times 10^{-1}x_{16}$ \\

8 & $-0.29 - 1.36\times 10^{-1}x_{1} - 8.04\times 10^{-2}x_{2} - 7.25\times 10^{-1}x_{3} - 1.91\times 10^{-1}x_{4} + 1.53\times 10^{0}x_{5} - 8.89\times 10^{-3}x_{6} - 2.63\times 10^{-1}x_{7} - 1.84\times 10^{0}x_{8} - 5.94\times 10^{-1}x_{9} - 1.08\times 10^{-1}x_{10} - 6.78\times 10^{-2}x_{11} - 4.03\times 10^{-1}x_{12} + 4.90\times 10^{-1}x_{13} - 3.36\times 10^{-1}x_{14} - 1.32\times 10^{-1}x_{15} - 4.92\times 10^{-3}x_{16}$ \\

9 & $0.96 + 4.04\times 10^{-1}x_{1} + 7.69\times 10^{-1}x_{2} + 1.56\times 10^{0}x_{3} + 6.04\times 10^{-1}x_{4} - 5.09\times 10^{-1}x_{5} - 1.21\times 10^{0}x_{6} + 3.04\times 10^{-1}x_{7} + 7.66\times 10^{-1}x_{8} + 3.56\times 10^{0}x_{9} - 2.42\times 10^{-1}x_{10} + 1.07\times 10^{0}x_{11} - 1.65\times 10^{0}x_{12} + 3.16\times 10^{-1}x_{13} - 3.75\times 10^{0}x_{14} - 7.01\times 10^{-1}x_{15} + 7.40\times 10^{-1}x_{16}$ \\

12 & $0.39 - 9.98\times 10^{-1}x_{1} + 8.61\times 10^{-1}x_{2} - 2.75\times 10^{-3}x_{3} - 1.74\times 10^{-1}x_{4} - 6.20\times 10^{-2}x_{5} + 3.66\times 10^{-1}x_{6} - 1.54\times 10^{-1}x_{7} + 7.18\times 10^{-1}x_{8} - 1.48\times 10^{0}x_{9} + 4.50\times 10^{-1}x_{10} - 7.87\times 10^{-1}x_{11} + 2.89\times 10^{0}x_{12} - 3.48\times 10^{-1}x_{13} + 3.41\times 10^{-1}x_{14} + 9.82\times 10^{-1}x_{15} - 1.67\times 10^{0}x_{16}$ \\

14 & $-1.23 + 1.12\times 10^{0}x_{1} - 5.37\times 10^{-1}x_{2} + 1.75\times 10^{0}x_{3} + 1.86\times 10^{-2}x_{4} - 1.99\times 10^{-1}x_{5} - 2.87\times 10^{-1}x_{6} - 8.80\times 10^{-2}x_{7} - 7.84\times 10^{-1}x_{8} - 4.17\times 10^{-1}x_{9} - 1.25\times 10^{0}x_{10} - 9.16\times 10^{-1}x_{11} + 1.05\times 10^{-1}x_{12} - 1.12\times 10^{0}x_{13} + 1.30\times 10^{-1}x_{14} - 4.25\times 10^{-3}x_{15} - 6.06\times 10^{-2}x_{16}$ \\

15 & $-0.54 - 6.91\times 10^{-1}x_{1} + 6.65\times 10^{-2}x_{2} - 2.58\times 10^{0}x_{3} - 6.76\times 10^{-1}x_{4} - 3.67\times 10^{-1}x_{5} + 7.71\times 10^{-1}x_{6} - 1.05\times 10^{-1}x_{7} + 7.40\times 10^{-1}x_{8} - 2.43\times 10^{0}x_{9} - 1.54\times 10^{-1}x_{10} - 9.79\times 10^{-1}x_{11} + 2.76\times 10^{0}x_{12} - 2.97\times 10^{-1}x_{13} + 4.25\times 10^{0}x_{14} + 1.85\times 10^{0}x_{15} - 1.25\times 10^{0}x_{16}$ \\

17 & $0.84 - 6.76\times 10^{-1}x_{1} + 2.39\times 10^{0}x_{2} + 3.34\times 10^{0}x_{3} - 2.74\times 10^{-1}x_{4} + 1.43\times 10^{-1}x_{5} + 2.82\times 10^{-1}x_{6} - 4.51\times 10^{-2}x_{7} + 1.89\times 10^{-1}x_{8} - 1.28\times 10^{0}x_{9} + 2.36\times 10^{-1}x_{10} + 9.13\times 10^{-2}x_{11} + 2.12\times 10^{0}x_{12} + 3.67\times 10^{-1}x_{13} - 4.03\times 10^{0}x_{14} - 5.49\times 10^{-1}x_{15} + 1.19\times 10^{0}x_{16}$ \\
\hline
\end{tabularx}
\caption{RoadRunner PPO: interpretable policy boundary for logistic regression}
\label{tab:RoadRunne_PPO_LogReg}
\end{table*}

\begin{table*}[htbp]
\centering
\begin{tabularx}{\textwidth}{X}
\hline
The RoadRunner A2C policy is single-action in the held-out rollout (it always selects action $4$), so only one class is present. Logistic regression requires at least two distinct classes and is therefore undefined for this configuration. \\
\hline
\end{tabularx}
\caption{RoadRunner A2C: logistic regression is undefined; the rollout is single-action (always action $4$)}
\label{tab:RoadRunne_A2C_LogReg}
\end{table*}

\begin{table*}[htbp]
\centering
\begin{tabularx}{\textwidth}{c X}
\hline
Action $i$ & $f_i(x)$ \\
\hline
0 & $8.30 - 1.00\times 10^{0}x_{1} - 5.01\times 10^{-2}x_{2} - 6.39\times 10^{0}x_{3} - 5.02\times 10^{-2}x_{4} - 6.78\times 10^{-1}x_{7} - 2.71\times 10^{0}x_{8} - 2.67\times 10^{0}x_{13} + 5.60\times 10^{0}x_{15}$ \\

2 & $3.41 - 1.29\times 10^{0}x_{1} - 5.54\times 10^{-1}x_{2} + 4.16\times 10^{-2}x_{4} + 2.10\times 10^{0}x_{7} - 4.70\times 10^{-1}x_{8} - 1.32\times 10^{-2}x_{13} - 1.24\times 10^{-3}x_{15}$ \\

3 & $-21.15 - 4.34\times 10^{0}x_{1} - 1.56\times 10^{1}x_{2} + 4.43\times 10^{-1}x_{4} - 2.51\times 10^{0}x_{7} + 1.05\times 10^{1}x_{8} - 1.52\times 10^{-2}x_{13} - 2.31\times 10^{-3}x_{15}$ \\

4 & $18.63 - 2.16\times 10^{0}x_{1} + 1.41\times 10^{1}x_{2} - 5.48\times 10^{0}x_{3} - 1.64\times 10^{-1}x_{4} - 4.28\times 10^{-1}x_{7} - 4.00\times 10^{0}x_{8} + 1.51\times 10^{0}x_{13} + 3.73\times 10^{0}x_{15}$ \\

5 & $-46.56 - 2.02\times 10^{1}x_{1} - 2.14\times 10^{1}x_{2} + 3.84\times 10^{0}x_{4} - 4.35\times 10^{-1}x_{7} + 1.78\times 10^{1}x_{8} - 1.11\times 10^{-2}x_{13} - 2.38\times 10^{-3}x_{15}$ \\

7 & $27.66 - 3.71\times 10^{0}x_{1} + 1.90\times 10^{1}x_{2} + 2.82\times 10^{0}x_{3} + 8.55\times 10^{-1}x_{4} - 6.67\times 10^{0}x_{7} - 7.80\times 10^{0}x_{8} + 6.62\times 10^{-1}x_{13} + 3.64\times 10^{0}x_{15}$ \\

8 & $-24.34 + 1.04\times 10^{1}x_{1} - 7.13\times 10^{-2}x_{2} - 8.97\times 10^{-3}x_{3} - 4.48\times 10^{-1}x_{4} + 1.79\times 10^{1}x_{7} - 2.18\times 10^{-1}x_{8} + 1.57\times 10^{0}x_{13} - 3.22\times 10^{0}x_{15}$ \\

9 & $-4.29 - 1.37\times 10^{1}x_{1} + 1.01\times 10^{1}x_{2} + 1.86\times 10^{0}x_{3} - 7.28\times 10^{-1}x_{4} - 2.44\times 10^{0}x_{7} + 5.98\times 10^{0}x_{8} + 1.28\times 10^{1}x_{13} + 8.91\times 10^{0}x_{15}$ \\

10 & $5.94 - 4.41\times 10^{-3}x_{1} - 4.05\times 10^{0}x_{2} - 3.67\times 10^{0}x_{3} + 8.58\times 10^{0}x_{4} - 1.19\times 10^{1}x_{7} - 7.06\times 10^{0}x_{8} - 1.21\times 10^{-2}x_{9} - 1.28\times 10^{-2}x_{13} - 5.96\times 10^{0}x_{15}$ \\

11 & $6.46 + 1.34\times 10^{1}x_{1} - 1.11\times 10^{1}x_{2} - 1.41\times 10^{0}x_{3} + 1.58\times 10^{0}x_{4} - 6.26\times 10^{0}x_{7} - 3.00\times 10^{0}x_{8} - 1.92\times 10^{-3}x_{9} - 4.88\times 10^{-1}x_{13} - 1.05\times 10^{1}x_{15}$ \\

14 & $-30.22 + 8.89\times 10^{0}x_{1} - 1.99\times 10^{0}x_{2} - 1.20\times 10^{1}x_{4} - 3.59\times 10^{0}x_{7} + 8.83\times 10^{0}x_{8} + 1.15\times 10^{1}x_{13} - 3.24\times 10^{0}x_{15}$ \\

15 & $24.43 + 4.67\times 10^{-1}x_{1} - 5.95\times 10^{-1}x_{2} - 2.39\times 10^{-2}x_{3} + 2.56\times 10^{0}x_{4} - 7.16\times 10^{0}x_{7} - 9.78\times 10^{0}x_{8} - 1.50\times 10^{0}x_{13} + 6.54\times 10^{-1}x_{15}$ \\

16 & $11.32 + 8.23\times 10^{0}x_{1} + 3.60\times 10^{0}x_{2} + 6.12\times 10^{0}x_{3} - 1.41\times 10^{0}x_{4} + 1.31\times 10^{1}x_{7} - 4.34\times 10^{0}x_{8} + 2.74\times 10^{-2}x_{9} - 1.15\times 10^{1}x_{13} - 8.99\times 10^{-1}x_{15}$ \\

17 & $20.39 + 4.95\times 10^{0}x_{1} + 8.64\times 10^{0}x_{2} + 6.18\times 10^{0}x_{3} - 3.12\times 10^{0}x_{4} + 8.99\times 10^{0}x_{7} - 3.70\times 10^{0}x_{8} - 1.33\times 10^{-2}x_{9} - 1.19\times 10^{1}x_{13} + 1.33\times 10^{0}x_{15}$ \\
\hline
\end{tabularx}
\caption{RoadRunner DQN: interpretable policy boundary for logistic regression}
\label{tab:RoadRunner_DQN_LogReg}
\end{table*}

\begin{table*}[htbp]
\centering
\begin{tabularx}{\textwidth}{c X}
\hline
Action $i$ & $f_i(x)$ \\
\hline
0 & $0.137 + 3.54\,x + 6.66\,y - 3.27\,v_x + 37.1\,v_y - 4.07\,\theta - 3.52\,\omega + 6.20\,c_L + 4.52\,c_R$ \\
1 & $0.318 + 11.5\,x + 5.51\,y + 18.3\,v_x + 14.7\,v_y - 41.9\,\theta - 32.9\,\omega - 0.354\,c_L - 2.54\,c_R$ \\
2 & $-3.41 - 4.62\,x - 11.5\,y + 5.91\,v_x - 61.9\,v_y + 3.86\,\theta - 7.75\,\omega - 1.32\,c_L - 0.984\,c_R$ \\
3 & $2.95 - 10.4\,x - 0.643\,y - 20.9\,v_x + 10.1\,v_y + 42.1\,\theta + 44.2\,\omega - 4.53\,c_L - 0.994\,c_R$ \\
\hline
\end{tabularx}
\caption{LunarLander PPO: interpretable policy boundary for logistic regression}
\label{tab:LunarLander_PPO_LogReg}
\end{table*}

\begin{table*}[htbp]
\centering
\begin{tabularx}{\textwidth}{c X}
\hline
Action $i$ & $f_i(x)$ \\
\hline
0 & $4.95 - 4.23\,x - 1.31\,y + 9.25\,v_x + 47.5\,v_y + 14.8\,\theta + 2.87\,\omega + 0.432\,c_L + 3.11\,c_R$ \\
1 & $0.0420 - 5.63\,x + 4.25\,y + 16.5\,v_x + 11.6\,v_y - 66.7\,\theta - 27.9\,\omega - 0.334\,c_L - 0.985\,c_R$ \\
2 & $-4.74 + 11.1\,x - 3.73\,y - 8.10\,v_x - 66.1\,v_y - 6.90\,\theta + 4.13\,\omega - 2.42\,c_L - 2.39\,c_R$ \\
3 & $-0.245 - 1.24\,x + 0.796\,y - 17.6\,v_x + 7.03\,v_y + 58.8\,\theta + 20.9\,\omega + 2.32\,c_L + 0.262\,c_R$ \\
\hline
\end{tabularx}
\caption{LunarLander A2C: interpretable policy boundary for logistic regression}
\label{tab:LunarLander_A2C_LogReg}
\end{table*}

\begin{table*}[htbp]
\centering
\begin{tabularx}{\textwidth}{c X}
\hline
Action $i$ & $f_i(x)$ \\
\hline
0 & $-0.774 + 0.798\,x + 2.71\,y + 1.28\,v_x + 0.263\,v_y + 1.63\,\theta + 1.35\,\omega + 0.990\,c_L + 1.25\,c_R$ \\
1 & $-0.178 + 3.15\,x + 0.443\,y + 1.70\,v_x + 0.924\,v_y - 5.19\,\theta - 1.67\,\omega - 0.338\,c_L + 0.00963\,c_R$ \\
2 & $-0.00264 - 3.30\,x - 3.55\,y - 0.417\,v_x - 5.08\,v_y - 1.36\,\theta - 2.79\,\omega + 0.439\,c_L - 1.81\,c_R$ \\
3 & $0.954 - 0.649\,x + 0.395\,y - 2.57\,v_x + 3.89\,v_y + 4.92\,\theta + 3.10\,\omega - 1.09\,c_L + 0.549\,c_R$ \\
\hline
\end{tabularx}
\caption{LunarLander DQN: interpretable policy boundary for logistic regression}
\label{tab:LunarLander_DQN_LogReg}
\end{table*}

\clearpage
\begin{figure*}[t]
\begin{minipage}[t]{0.48\textwidth}
\vspace{0pt}
\section{Logistic Regression' Prompt} \label{Appendix: Logistic Regression' Prompt}

\begin{promptbox}{XRL Logistic Regression Prompt Template}
\footnotesize
\textbf{Role:} You are an Explainable AI (XAI) Assistant. Explain an RL agent's decision from the probability distribution of a Logistic Regression policy.

\textbf{1. Input Parameters:}
\begin{itemize}
    \item \textbf{Domain:} [e.g., Ms. Pac-Man]
    \item \textbf{Logit Equations:} [User Input, e.g., $f_0(x) = \dots, f_6(x) = \dots$]
    \item \textbf{Current State Features:} [e.g., $x_1, x_2, \dots, x_{16}$]
\end{itemize}

\textbf{2. Task Instructions:}
\begin{itemize}
    \item \textbf{Logit Calculation:} Compute the logit for each action $i$ from the given equations.
    \item \textbf{Probability Transformation:} Apply softmax, $P(a{=}i|x) = e^{f_i(x)}/\sum_j e^{f_j(x)}$, to normalize the logits.
    \item \textbf{Selection Logic:} Identify the \textit{highest-probability} action ($\arg\max$) and explain which features drove its dominant logit.
\end{itemize}

\textbf{3. Output Format:}
\begin{itemize}
    \item \textbf{[Logit Scores]:} The raw logit per action (e.g., $f_6(x)\approx4.89$).
    \item \textbf{[Probability Distribution]:} The softmax probability per action.
    \item \textbf{[Human-Readable Explanation]:} A 2--3 sentence summary of the winning action.
\end{itemize}
\end{promptbox}
\end{minipage}
\hfill
\begin{minipage}[t]{0.48\textwidth}
\vspace{0pt}

\section{Natural Language Explanation Synthesis for Logistic Regression Policy in MsPacMan PPO} \label{Appendix: Natural Language Explanation Synthesis for Logistic Regression}

\begin{itemize}
    \item \textbf{[Logit Scores]}
    \begin{itemize}
        \item $f_0(x) \approx 3.84$
        \item $f_6(x) \approx 4.89$
        \item $f_7(x) \approx -12.02$
        \item $f_8(x) \approx 3.29$
    \end{itemize}

    \item \textbf{[Probability Distribution]}
    \begin{itemize}
        \item The Softmax function $P(a = i \mid x) = \frac{e^{f_i(x)}}{\sum_{j} e^{f_j(x)}}$ converts logits to the following probabilities:
        \item \begin{tabular}{lrr}
            \toprule
            \textbf{Action ($i$)} & \textbf{Logit ($f_i$)} & \textbf{Probability} \\
            \midrule
            Action 0 & $3.84$ & $\approx 23\%$ \\
            \textbf{Action 6} & $\mathbf{4.89}$ & $\mathbf{\approx 65\%}$ \\
            Action 7 & $-12.02$ & $\approx 0\%$ \\
            Action 8 & $3.29$ & $\approx 13\%$ \\
            \bottomrule
            \end{tabular}
    \end{itemize}

    \item \textbf{[Human-Readable Explanation]}
    \begin{itemize}
        \item The agent selects \textbf{Action 6} because its logit ($4.89$) is the highest, modestly ahead of Action 0 ($3.84$) and Action 8 ($3.29$), while Action 7 is strongly suppressed ($-12.02$).
        \item The softmax turns this lead into a probability of $\approx 0.65$ for Action 6, versus $\approx 0.23$ and $\approx 0.13$ for Actions 0 and 8.
        \item The choice is therefore confident but not deterministic: the neighbouring actions retain a non-trivial share of probability, reflecting a genuinely competitive state.
    \end{itemize}
\end{itemize}
\end{minipage}
\end{figure*}

\clearpage

\end{document}